\documentclass[10pt,journal,compsoc]{IEEEtran}

\ifCLASSOPTIONcompsoc
\usepackage[nocompress]{cite}
\else
\usepackage{cite}
\fi

\ifCLASSINFOpdf
\else
\fi
\usepackage{cite}
\usepackage{graphicx}
\usepackage{textcomp}
\usepackage[T1]{fontenc}
\usepackage{booktabs} 
\usepackage[ruled,linesnumbered]{algorithm2e}
\usepackage{booktabs}
\usepackage[british]{babel}
\usepackage{paralist}
\usepackage{multirow}
\usepackage{amsmath}
\usepackage{amssymb}
\usepackage[marginal]{footmisc}
\usepackage{subfigure}
\usepackage{balance}
\usepackage{threeparttable}
\usepackage{hyperref}
\usepackage{color}
\usepackage[dvipsnames]{xcolor}
\usepackage{amsfonts}
\usepackage{colortbl}
\usepackage{rotating}
\usepackage{array}
\usepackage[framemethod=default]{mdframed}
\usepackage{showexpl}
\usepackage{comment}
\usepackage{float}
\usepackage{url}
\usepackage{mathalfa}
\makeatletter
\g@addto@macro{\UrlBreaks}{\UrlOrds}
\makeatother
\usepackage{framed}
\usepackage{alltt}
\usepackage{enumitem}

\SetCommentSty{mycommfont}
\usepackage{algpseudocode}
\usepackage{tikz}
\usetikzlibrary{trees}
\usepackage{breqn}
\usepackage{pgfplots}
\usepackage{bchart}
\usepackage{makecell}
\usepackage{caption} 
\newcommand{\papernum}{144\xspace}

\newcommand{\jie}[1]{\textcolor{black}{#1}}
\newcommand{\jienew}[1]{\textcolor{black}{#1}}

\newcommand{\mlt}[1]{\textcolor{black}{ML testing}}
\newtheorem{definition}{Definition}

\begin{document}
	%
	\title{Machine Learning Testing: \\Survey, Landscapes and Horizons}

	\author{Jie M. Zhang*, Mark Harman, Lei Ma, Yang Liu
		\IEEEcompsocitemizethanks{\IEEEcompsocthanksitem Jie M. Zhang and Mark Harman are with CREST, University College London, United Kingdom. Mark Harman is also with Facebook London. \protect\\
			E-mail: {jie.zhang, mark.harman}@ucl.ac.uk\protect\\
			* Jie M. Zhang is the corresponding author.}
		\IEEEcompsocitemizethanks{\IEEEcompsocthanksitem Lei Ma is with Kyushu University, Japan.\protect\\
			E-mail: {malei@ait.kyushu-u.ac.jp}}
			\IEEEcompsocitemizethanks{\IEEEcompsocthanksitem Yang Liu is with Nanyang Technological University, Singapore.\protect\\
			E-mail: {yangliu@ntu.edu.sg}}
		
	}

 	\markboth{}%
 	{Zhang \MakeLowercase{\textit{et al.}}: Machine Learning Testing: Survey, Landscapes and Horizons }
	
	\IEEEtitleabstractindextext{%
		\begin{abstract}
			
			This paper provides a comprehensive survey of techniques for testing machine learning systems; Machine Learning Testing (ML testing) research.
			It covers \papernum papers on testing properties (e.g., correctness, robustness, and fairness), 
			testing components (e.g., the data, learning program, and framework), 
		    testing workflow (e.g., test generation and test evaluation), and application scenarios (e.g., autonomous driving, machine translation).
			The paper also analyses trends concerning datasets, research trends, and research focus, concluding with research challenges and promising research directions in \mlt{}.
		\end{abstract}
		
		}

	\maketitle

	\IEEEdisplaynontitleabstractindextext

	%
	\IEEEpeerreviewmaketitle

	\IEEEraisesectionheading{\section{Introduction}
	\label{sec:introduction}}



The prevalent applications
of machine learning
arouse natural concerns about trustworthiness. 
Safety-critical applications such as 
self-driving systems \cite{pei2017deepxplore,chen2015deepdriving} 
and medical treatments \cite{litjens2017survey}, 
increase the importance of behaviour relating to correctness, robustness, privacy, efficiency and fairness.
Software testing refers to any activity that aims to detect the differences between existing and required behaviour~\cite{ammann2016introduction}.
With the recent rapid rise in interest and activity,
testing has been demonstrated to be an effective way to expose problems and potentially facilitate to improve the trustworthiness of machine learning systems.

For example, DeepXplore~\cite{pei2017deepxplore}, a differential white-box testing technique for deep learning, revealed thousands of incorrect corner
case behaviours in autonomous driving learning systems;
Themis~\cite{galhotra2017fairness}, a fairness testing technique for detecting causal discrimination, 
detected significant ML model discrimination towards gender, marital status, or race for as many as 77.2\% of the individuals in datasets to which it was applied.

In fact, some aspects of the testing problem for machine learning systems are shared with well-known solutions already widely studied in the software engineering literature.
Nevertheless, the statistical nature of machine learning systems and their ability to make autonomous decisions raise additional, and challenging, research questions for software testing \cite{Paculablog2011,Ramanathan2016}.


Machine learning testing poses challenges that arise from the fundamentally different nature and construction of machine learning systems, compared to traditional (relatively more deterministic and less statistically-orientated) software systems.
For instance, a machine learning system inherently follows a data-driven programming paradigm, where the decision logic is obtained via a training procedure from training data under the machine learning algorithm's architecture~\cite{Amershi2019se4ai}. 
The model's behaviour may evolve over time, in response to the frequent provision of new data~\cite{Amershi2019se4ai}. 
While this is also true of traditional software systems, the core underlying behaviour of a traditional system does not typically change in response to new data, in the way that a machine learning system can.

Testing machine learning also suffers from a particularly pernicious instance of the \emph{Oracle Problem} \cite{barr2015oracle}.
Machine learning systems are difficult to test because they are designed to provide an answer to a question for which no previous answer exists~\cite{murphy2007approach}.
As Davis and Weyuker said~\cite{Davis1981}, for these kinds of systems `There would be no need to write such programs, if the correct answer were known'. 
Much of the literature on testing machine learning systems seeks to find techniques that can tackle the Oracle problem, often drawing on traditional software testing approaches.

The behaviours of interest for machine learning systems are also typified by emergent properties, the effects of which can only be fully understood by considering the machine learning  system as a whole.
This makes testing harder, because it is less obvious how to break the system into smaller components that can be tested, as units, in isolation.
From a testing point of view, this emergent behaviour has a tendency to migrate testing challenges from the unit level to the integration and system level.
For example, low accuracy/precision of a machine learning model is typically a composite effect, arising from a combination of the behaviours of different components such as the training data, the learning program, and even the learning framework/library~\cite{Amershi2019se4ai}. 

Errors may propagate to become amplified or suppressed, inhibiting the tester's ability to decide where the fault lies.
These challenges also apply in more traditional software systems, where, for example, previous work has considered failed error propagation \cite{kaetal:analysis,voas:testability}
and the subtleties introduced by
fault masking \cite{clark:squeeziness,yjmh:hom-scam-best-paper}. 
However, these problems are far-reaching in machine learning systems, since they arise out of the nature of the machine learning approach and fundamentally affect all behaviours, rather than arising as a side effect of traditional data and control flow \cite{Amershi2019se4ai}.

\begin{figure}[h!]
	\centering
	\includegraphics[scale=0.48]{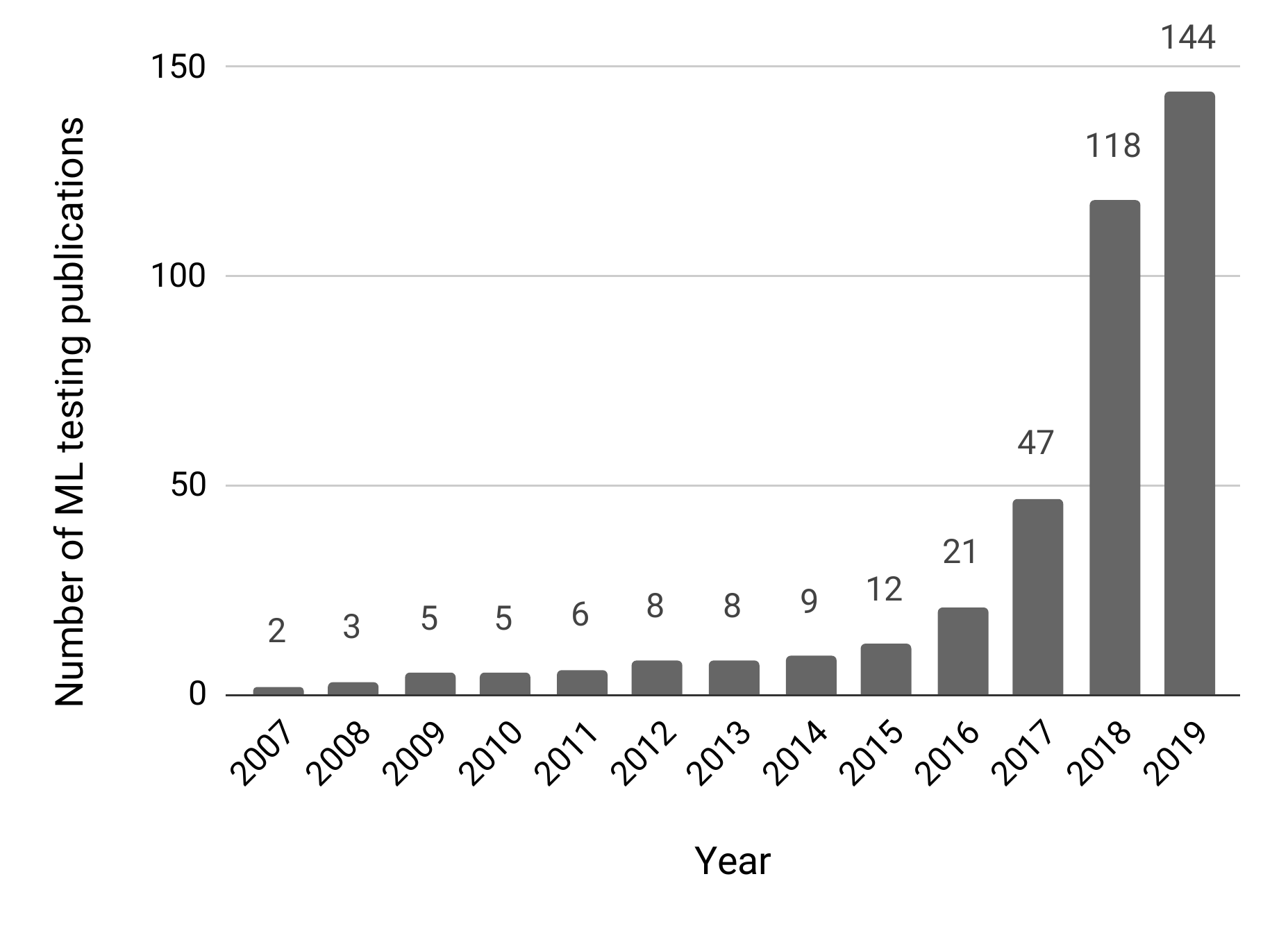}
	\caption{Machine Learning Testing Publications (accumulative) during 2007-2019}
	\label{fig:numberofpublications}
\end{figure}

For these reasons, machine learning systems are thus sometimes regarded as \emph{`non-testable'}  software.
Rising to these challenges, the literature has seen considerable progress and a notable upturn in interest and activity:
Figure~\ref{fig:numberofpublications} shows the cumulative number of publications on the topic of testing machine learning systems between 2007 and June 2019 \jie{(we introduce how we collected these papers in Section~\ref{sec:papercollection})}. 
From this figure, we can see that 85\% of papers have appeared since 2016, testifying to the emergence of new software testing domain of interest: machine learning testing.

In this paper, we use the term `\textbf{Machine Learning Testing}' (\textbf{ML testing}) to refer to
any activity aimed at detecting differences between existing and required behaviours of machine learning systems.
\mlt{} is different from testing approaches that use machine learning or those that are guided by machine learning, which should be referred to as `machine learning-based testing'.
This nomenclature accords with previous usages in the software engineering literature. 
For example, the literature uses the terms `state-based testing' \cite{turner1993state} 
and `search-based testing' \cite{harman2001search, harman2010theoretical} 
to refer to testing techniques that make use of concepts of state and search space, 
whereas we use the terms `GUI testing' \cite{memon2002gui}
and `unit testing' \cite{sen2005cute} 
to refer to test techniques that tackle challenges of testing GUIs (Graphical User Interfaces) and code units.




This paper seeks to provide a comprehensive survey of \mlt{}.
We draw together the aspects of previous work that specifically concern software testing, while simultaneously covering all types of approaches to machine learning that have hitherto been tackled using testing.
The literature is organised according to four different aspects: the testing properties (such as correctness, robustness, and fairness), machine learning components (such as the data, learning program, and framework), testing workflow (e.g., test generation, test execution, and test evaluation), and application scenarios (e.g., autonomous driving and machine translation).
Some papers address multiple aspects.
For such papers, we mention them in all the aspects correlated (in different sections).
This ensures that each aspect is complete. 

Additionally, we summarise research distribution (e.g., among testing different machine learning categories), trends, and datasets.
We also identify open problems and challenges for the emerging research community working at the intersection between techniques for software testing and problems in machine learning testing. 
To ensure that our survey is self-contained, we aimed to include sufficient material to fully orientate software engineering researchers who are interested in testing and curious about testing techniques for machine learning applications.
We also seek to provide machine learning researchers with a complete survey of software testing solutions for improving the trustworthiness of machine learning systems.

There has been previous work that discussed or surveyed aspects of the literature related to \mlt{}.
Hains et al.~\cite{Hains2018}, Ma et al.~\cite{ma2018secure}, and Huang et al.~\cite{huang2018safety}
surveyed secure deep learning, \jie{in which the focus was deep learning security with testing being one of the assurance techniques}.
Masuda et al.~\cite{masuda2018survey} outlined their collected papers on software quality for machine learning applications in a short paper. 
Ishikawa~\cite{ishikawa2018concepts} discussed the foundational concepts that might be used in any and all \mlt{} approaches.
Braiek and Khomh~\cite{Braiek2018} discussed defect detection in machine learning data and/or models in their review of 39 papers.
As far as we know, no previous work has provided a comprehensive survey particularly focused on machine learning testing.



In summary, the paper makes the following contributions:

1) \textbf{Definition}. The paper defines Machine Learning Testing (\mlt{}), 
overviewing the concepts, testing workflow, testing properties, and testing components related to machine learning testing. 

2) \textbf{Survey}. The paper provides a comprehensive survey of \papernum machine learning testing papers, across various publishing areas such as software engineering, artificial intelligence, systems and networking, and data mining.

3) \textbf{Analyses}. The paper analyses and reports data on the research distribution, datasets, and trends that characterise the machine learning testing literature. We observed a pronounced imbalance in the distribution of research efforts: among the \papernum papers we collected,
around 120 of them tackle supervised learning testing, three of them tackle unsupervised learning testing, and only one paper tests reinforcement learning.
Additionally, 
most of them (93) centre on correctness and robustness, 
but only a few papers test interpretability, privacy, or efficiency.

4) \textbf{Horizons}. The paper identifies challenges, open problems, and promising research directions for \mlt{}, with the aim of facilitating and stimulating further research.

\begin{figure}[t!]
	\centering
	\includegraphics[scale=0.55]{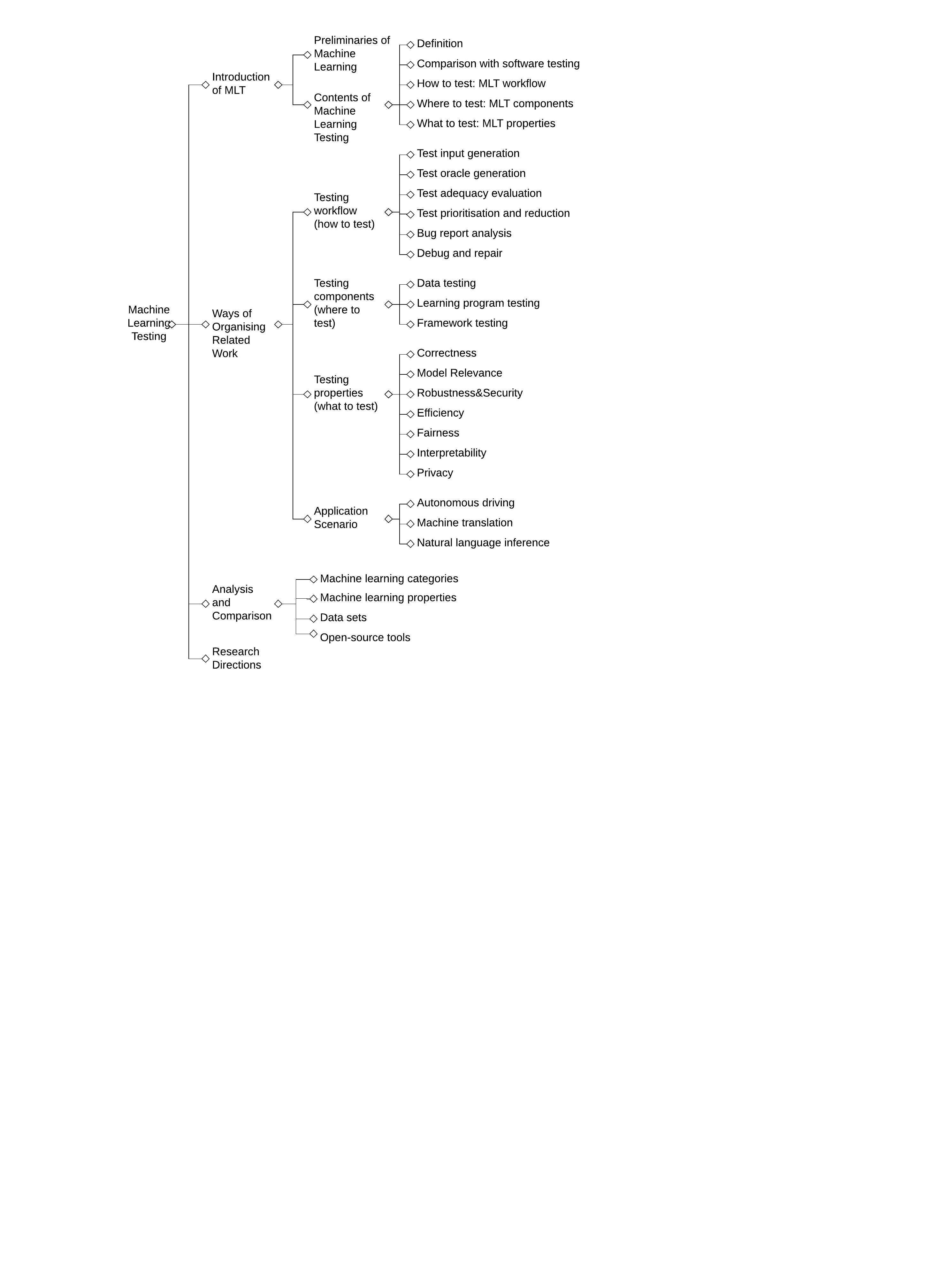}
	\caption{Tree structure of the contents in this paper}
	\label{fig:sectiontreegraph}
\end{figure}

Figure~\ref{fig:sectiontreegraph} depicts the paper structure. 
More details of the review schema can be found in Section~\ref{sec:schema}.



\section{Preliminaries of Machine Learning}
\label{sec:preliminaries}

This section reviews the fundamental terminology in machine learning so as to make the survey self-contained.


Machine Learning (ML) is a type of artificial intelligence technique that makes decisions or predictions from data~\cite{shawe2004kernel,mohri2012foundations}. 
A machine learning system is typically composed from following elements or terms.
\noindent \textbf{Dataset}: A set of instances for building or evaluating a machine learning model. 

At the top level, the data could be categorised as:


\begin{itemize}
\item\textbf{Training data}: the data used to `teach' (train) the algorithm to perform its task.
\item \textbf{Validation data}: the data used to tune the hyper-parameters of a learning algorithm.
\item \textbf{Test data}: the data used to validate machine learning model behaviour. 
\end{itemize}

\noindent \textbf{Learning program}: the code written by developers to build and validate the machine learning system.

\noindent \textbf{Framework}: the library, or platform being used when building a machine learning model, such as \emph{Pytorch}~\cite{paszke2017automatic}, \emph{TensorFlow}~\cite{tensorflow2015whitepaper}, \emph{Scikit-learn}~\cite{scikitlearn}, \emph{Keras}~\cite{chollet2015keras}, and \emph{Caffe}~\cite{jia2014caffe}.


In the remainder of this section, we give definitions for other ML terminology used throughout the paper.

\noindent \textbf{Instance}: a piece of data recording the information about an object. 

\noindent \textbf{Feature}: a measurable property or characteristic of a phenomenon being observed to describe the instances.

\noindent \textbf{Label}: value or category assigned to each data instance. 

\noindent \textbf{Test error}: the difference ratio between the real conditions and the predicted conditions.

\noindent \textbf{Generalisation error}: the expected difference ratio between the real conditions and the predicted conditions of any valid data.

\noindent \textbf{Model}: the learned machine learning artefact that encodes decision or prediction logic which is trained from the training data, the learning program, and frameworks. 


There are different types of machine learning.
From the perspective of training data characteristics, machine learning includes:

\noindent \textbf{Supervised learning}: a type of machine learning that learns from training data with labels as learning targets. 
It is the most widely used type of machine learning~\cite{jordan2015machine}.

\noindent \textbf{Unsupervised learning}: a learning methodology that learns from training data without labels and relies on understanding the data itself.

\noindent \textbf{Reinforcement learning}: a type of machine learning where the data are in the form of sequences of
actions, observations, and rewards, and the learner learns how to take actions to interact in a specific environment so as to maximise the specified rewards.

Let $\mathcal{X}=(x_1,...,x_m)$ be the set of unlabelled training data.
Let $\mathcal{Y}=(c(x_1),...,c(x_m))$ be the set of labels corresponding to each piece of training data $x_i$.
Let concept $C: \mathcal{X}\rightarrow \mathcal{Y}$ be the mapping from $\mathcal{X}$ to $\mathcal{Y}$ (the real pattern).
The task of supervised learning is to learn a mapping pattern, i.e., a model, $h$ based on $\mathcal{X}$ and $\mathcal{Y}$ so that the learned model $h$ is similar to its true concept $C$ with a very small generalisation error. 
The task of unsupervised learning is to learn patterns or clusters from the data without knowing the existence of labels $\mathcal{Y}$.

\jienew{Reinforcement learning guides and plans with the learner actively interacting with the environment to achieve a certain goal.
It is usually modelled as a Markov decision process.
Let $S$ be a set of states, $A$ be a series of actions.
Let $s$ and $s'$ be two states. 
Let $r_a(s,s')$ be the reward after transition from $s$ to $s'$ with action $a \in A$.
Reinforcement learning is to learn how to take actions in each step to maximise the target awards.  
}


Machine learning can be applied to the following typical tasks~\cite{mohri2012foundations}\footnote{\jienew{These tasks are defined based on the nature of the problems solved instead of specific application scenarios such as language modelling.}}:

1) \textbf{Classification}: to assign a category to each data instance; E.g., image classification, handwriting recognition.

2) \textbf{Regression}: to predict a value for each data instance; 
E.g., temperature/age/income prediction.


3) \textbf{Clustering}: to partition instances into homogeneous regions;
E.g., pattern recognition, market/image segmentation.

4) \textbf{Dimension reduction}: to reduce the training complexity;
E.g., dataset representation, data pre-processing.

5) \textbf{Control}: to control actions to maximise rewards;
E.g., game playing.

\begin{figure}[h!]
	\centering
	\includegraphics[scale=0.63]{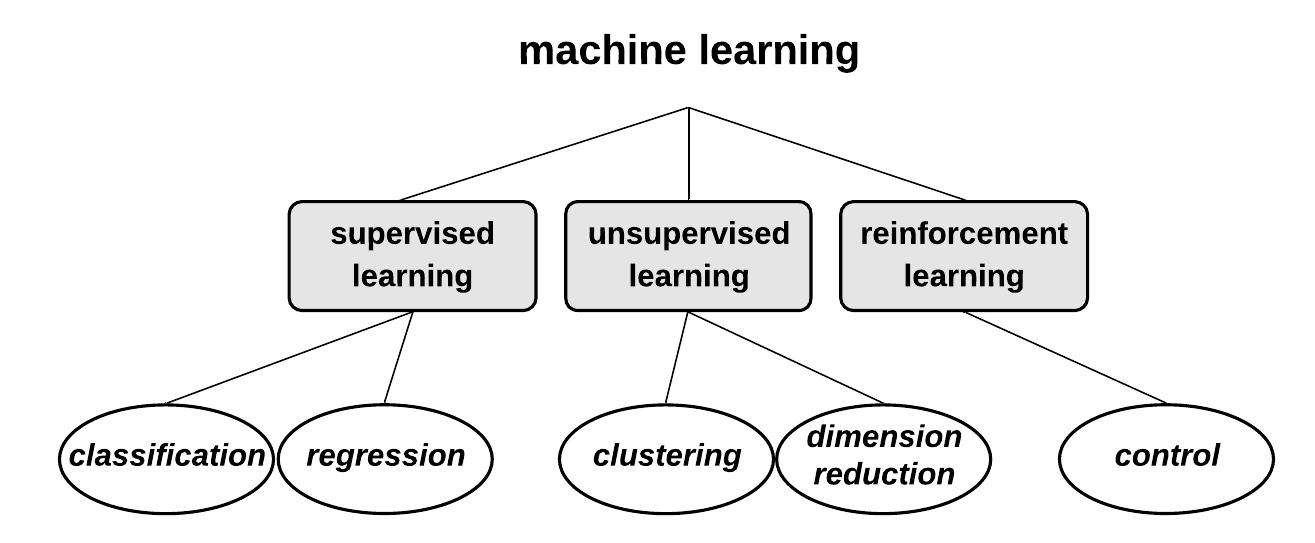}
	\caption{Machine learning categories and tasks}
	\label{fig:mltypeandtask}
\end{figure}

Figure~\ref{fig:mltypeandtask} shows the relationship between different categories of machine learning and the five machine learning tasks.
Among the five tasks, classification and regression belong to supervised learning;
Clustering and dimension reduction belong to unsupervised learning.
Reinforcement learning is widely adopted to control actions, such as to control AI-game players to maximise the rewards for a game agent.

In addition, machine learning can be classified into \textbf{classic machine learning} and \textbf{deep learning}.
Algorithms like Decision Tree~\cite{safavian1991survey}, SVM~\cite{shawe2004kernel}, linear regression~\cite{neter1996applied}, and Naive Bayes~\cite{mccallum1998comparison} all belong to classic machine learning.
Deep learning~\cite{lecun2015deep} applies Deep Neural Networks (DNNs) that uses multiple layers of nonlinear processing units for feature extraction and transformation. 
Typical deep learning algorithms often follow some widely used neural network structures like Convolutional Neural Networks (CNNs)~\cite{kim2014convolutional}
and Recurrent Neural Networks (RNNs)~\cite{graves2013speech}.
The scope of this paper involves both classic machine learning and deep learning.



\section{Machine Learning Testing}
\label{sec:mltesting}

This section gives a definition and analyses of \mlt{}.
It describes the testing workflow (\textbf{how} to test), testing properties (\textbf{what} to test), and testing components (\textbf{where} to test).

\subsection{Definition}
\label{sec:definition}


A software bug refers to an imperfection in a computer program that causes a discordance between the existing and the required conditions~\cite{5399061}.
In this paper, we refer the term `bug' to the differences between existing and required behaviours of an ML system\footnote{The existing related papers may use other terms like `defect' or `issue'. This paper uses `bug' as a representative of all such related terms considering that `bug' has a more general meaning~\cite{5399061}.}.

\begin{definition}[ML Bug]
An ML bug refers to any imperfection in a machine learning item that causes a discordance between the existing and the required conditions.
\end{definition}


We define \mlt{} as any activity aimed to detect ML bugs.

\begin{definition}[ML Testing]
Machine Learning Testing (ML testing) refers to any activity designed to reveal machine learning bugs.
\end{definition}

The definitions of machine learning bugs and \mlt{} indicate three aspects of machine learning: the required conditions, the machine learning items, and the testing activities.
A machine learning system may have different types of `required conditions', such as correctness, robustness, and privacy.
An ML bug may exist in the data, the learning program, or the framework.
The testing activities may include test input generation, test oracle identification, test adequacy evaluation, and bug triage.
In this survey, we refer to the above three aspects as testing properties, testing components, and testing workflow, respectively, according to which we collect and organise the related work.

Note that a test input in \mlt{} can be much more diverse in its form than that used in traditional software testing, because it is not only the code that may contain bugs, but also the data.
\jie{When we try to detect bugs in data, one may even use a training program as a test input to check some properties required for the data.}


\subsection{ML Testing Workflow}
\label{sec:mltestingworkflow}

ML testing workflow is about \textbf{how} to conduct \mlt{} with different testing activities. 
In this section, we first briefly introduce the role of ML testing when building ML models,
then present the key procedures and activities in ML testing.
We introduce more details of the current research related to each procedure in Section~\ref{sec:testingworkflow}.

\subsubsection{Role of Testing in ML Development} 

\begin{figure}[h]
	\centering
	\includegraphics[width=0.5\textwidth]{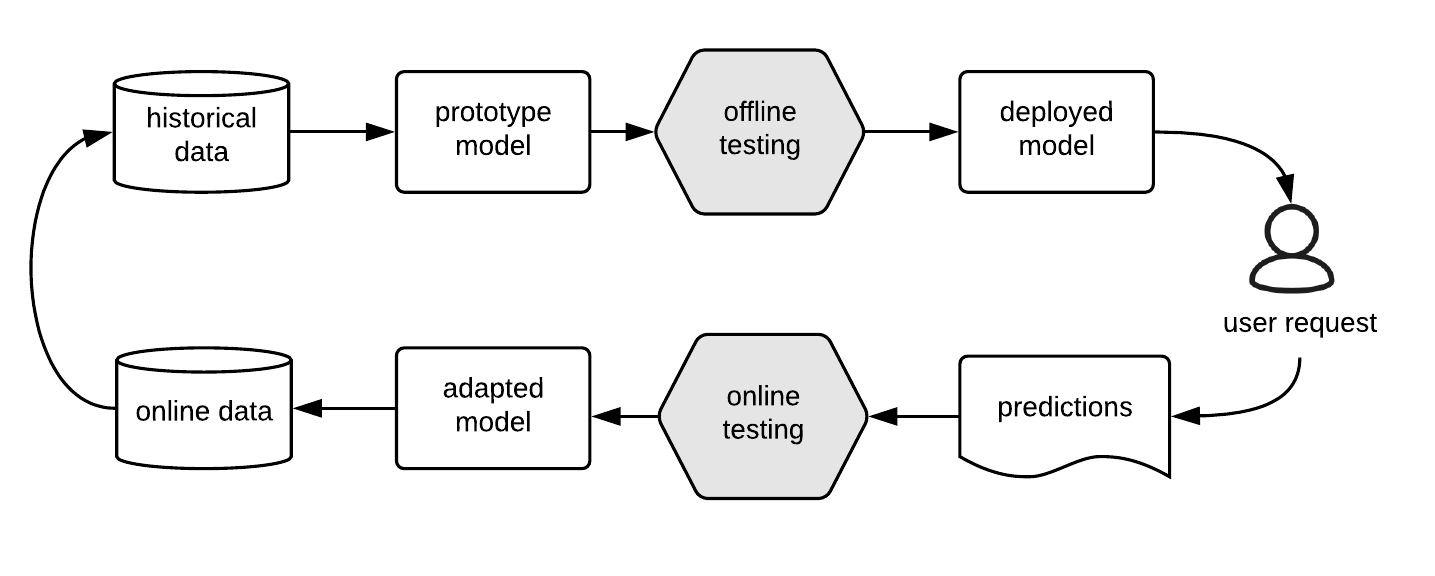}
	\caption{Role of ML testing in ML system development}
	\label{fig:mltestroles}
\end{figure}

Figure~\ref{fig:mltestroles} shows the life cycle of deploying a machine learning system with ML testing activities involved.
At the very beginning, a prototype model is generated based on historical data;
before deploying the model online, one needs to conduct offline testing, such as cross-validation, to make sure that the model meets the required conditions.
After deployment, the model makes predictions, yielding new data that can be analysed via online testing to evaluate how the model interacts with user behaviours.

There are several reasons that make online testing essential.
First, offline testing usually relies on test data, while test data usually fails to fully represent future data~\cite{werpachowski2019detecting};
Second, offline testing is not able to test some circumstances that may be problematic in real applied scenarios, such as data loss and call delays.
In addition, offline testing has no access to some business metrics such as open rate, reading time, and click-through rate.

In the following, we present an \mlt{} workflow adapted from classic software testing workflows.
Figure~\ref{fig:testworkflow} shows the workflow, including both offline testing and online testing.

\subsubsection{Offline Testing}

\begin{figure*}[h!]
	\centering
	\includegraphics[width=0.85\textwidth]{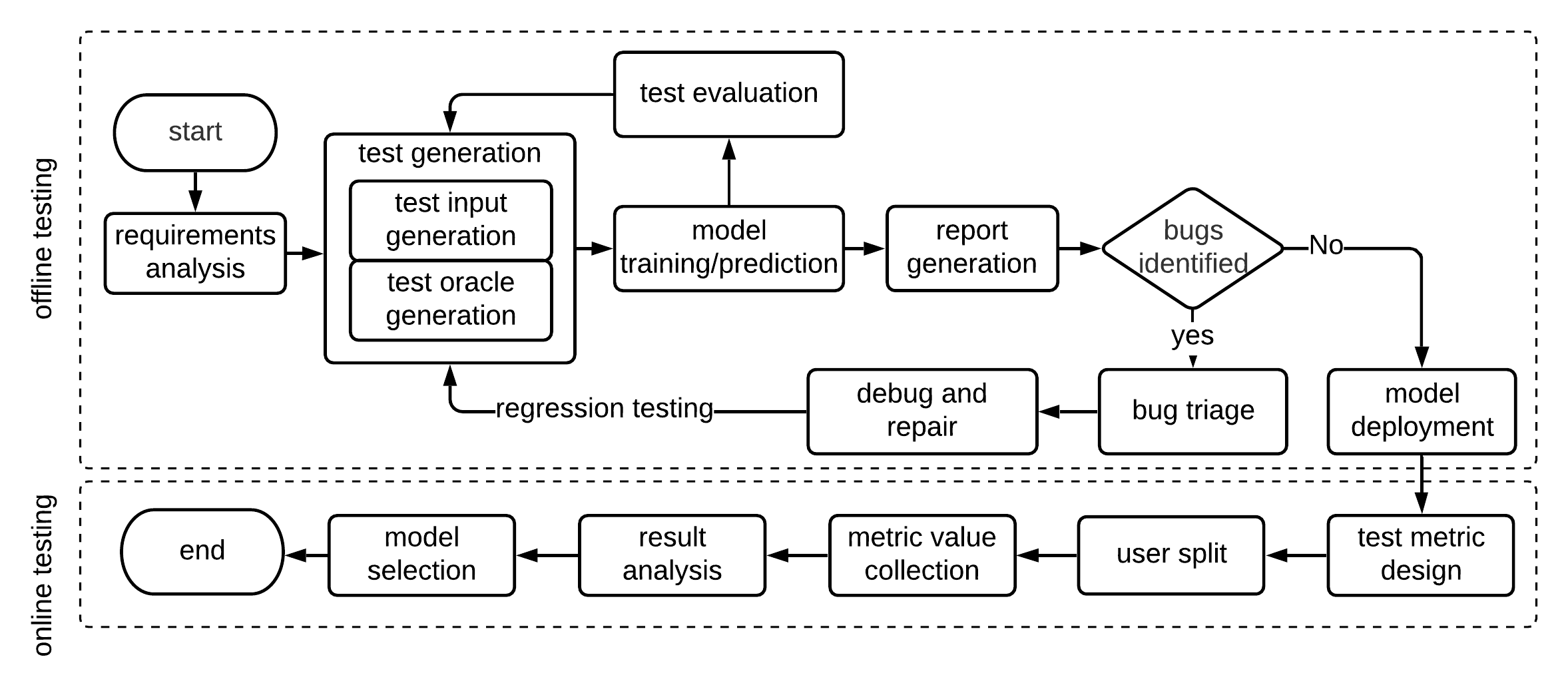}
	\caption{Idealised Workflow of ML testing}
	\label{fig:testworkflow}
\end{figure*}

The workflow of offline testing is shown by the top dotted rectangle of Figure~\ref{fig:testworkflow}.
At the very beginning, developers need to conduct requirement analysis to define the expectations of the users for the machine learning system under test. 
In requirement analysis, specifications of a machine learning system are analysed and the whole testing procedure is planned.
After that, test inputs are either sampled from the collected data or generated based on a specific purpose.
Test oracles are then identified or generated (see Section~\ref{sec:oracle} for more details of test oracles in machine learning). 
When the tests are ready, they need to be executed for developers to collect results.
The test execution process involves building a model with the tests (when the tests are training data) or running a built model against the tests (when the tests are test data), as well as checking whether the test oracles are violated. 
After the process of test execution, developers may use evaluation metrics to check the quality of tests, i.e., the ability of the tests to expose ML problems. 

The test execution results yield a bug report to help developers to duplicate, locate, and solve the bug.
Those identified bugs will be labelled with different severity and assigned for different developers.
Once the bug is debugged and repaired, regression testing is conducted to make sure the repair solves the reported problem and does not bring new problems. 
If no bugs are identified, the offline testing process ends, and the model is deployed.

\subsubsection{Online Testing}

Offline testing tests the model with historical data without in the real application environment.
It also lacks the data collection process of user behaviours.
Online testing complements the shortage of offline testing, and aims to detect bugs after the model is deployed online.

The workflow of online testing is shown by the bottom of Figure~\ref{fig:testworkflow}.
\jienew{There are different methods of conducting online testing for different purposes. For example, runtime monitoring keeps checking whether the running ML systems meet the requirements or violate some desired runtime properties. 
Another commonly used scenario is to monitor user responses, based on which to find out whether the new model is superior to the old model under certain application contexts.
A/B testing is one typical type of such online testing~\cite{cruz2018enabling}.
It splits customers to compare two versions of the system (e.g., web pages).
When performing A/B testing on ML systems, the sampled users will be split into two groups using the new and old ML models separately. 
}


MAB (Multi-Armed Bandit) is another online testing approach~\cite{kaufmann2016complexity}. 
It first conducts A/B testing for a short time and finds out the best model, then put more resources on the chosen model.


\subsection{ML Testing Components}
\label{sec:mltestingcomponents}

To build a machine learning model, an ML software developer usually needs to collect data, label the data, design learning program architecture, and implement the proposed architecture based on specific frameworks. 
The procedure of machine learning model development requires interaction with several components such as data, learning program, and learning framework, while each component may contain bugs.


Figure~\ref{fig:newmlcomponent} shows the basic procedure of building an ML model and the major components involved in the process.
Data are collected and pre-processed for use;
the learning program is the code for running to train the model;
the framework (e.g., Weka, scikit-learn, and TensorFlow) offers algorithms and other libraries for developers to choose from, when writing the learning program.

Thus, when conducting \mlt{}, developers may need to try to find bugs in every component including the data, the learning program, and the framework.
In particular, error propagation is a more serious problem in ML development because the components are more closely bonded with each other than traditional software~\cite{Amershi2019se4ai}, which indicates the importance of testing each of the ML components.
We introduce the bug detection in each ML component below:

\begin{figure}[h]
	\centering
	\includegraphics[scale=0.48]{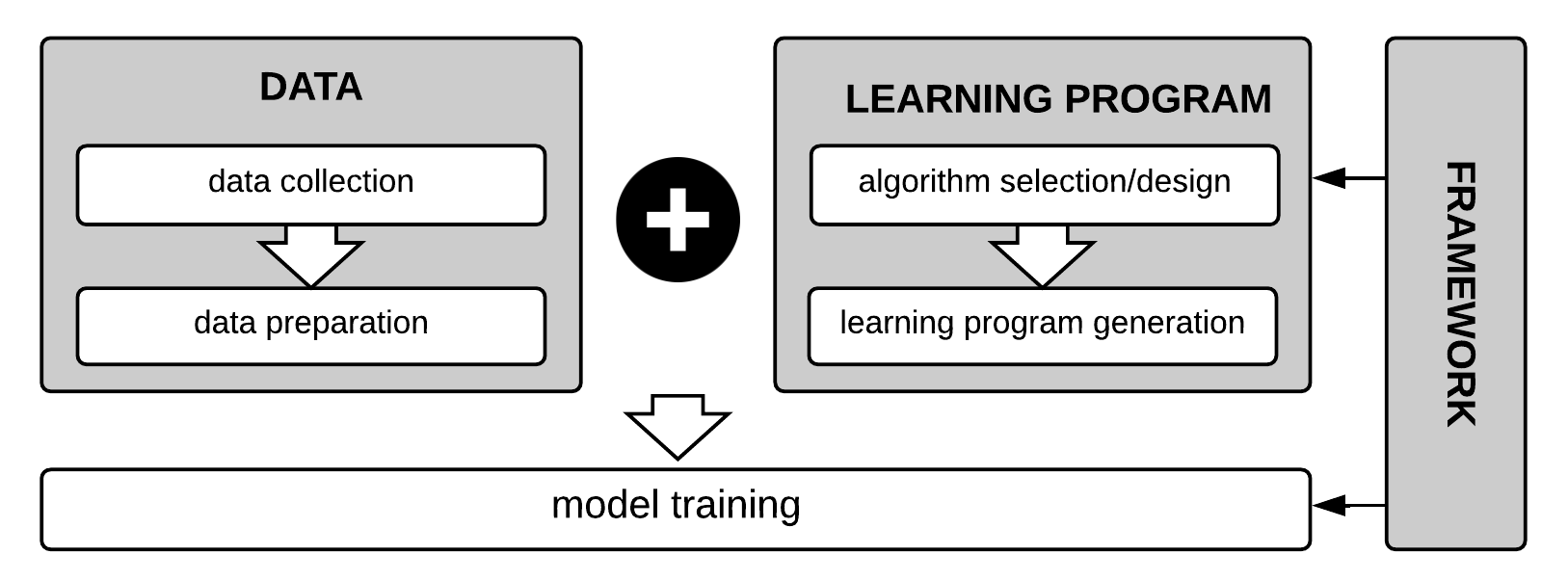}
	\caption{Components (shown by the grey boxes) involved in ML model building.}
	\label{fig:newmlcomponent}
\end{figure}


\noindent \textbf{\textbf{Bug Detection in Data}}.
The behaviours of a machine learning system largely depends on data~\cite{Amershi2019se4ai}.
Bugs in data affect the quality of the generated model, and can be amplified to yield more serious problems over a period a time~\cite{breck2019datavalidation}.
Bug detection in data checks problems such as whether the data is sufficient for training or test a model (also called completeness of the data~\cite{cheng2018towards}), whether the data is representative of future data, 
whether the data contains a lot of noise such as biased labels,
whether there is skew between training data and test data~\cite{breck2019datavalidation}, 
and whether there is data poisoning~\cite{alfeld2016data} or adversary information that may affect the model's performance.

\noindent \textbf{\textbf{Bug Detection in Frameworks}}.  
Machine Learning requires a lot of computations.
As shown by Figure~\ref{fig:newmlcomponent}, ML frameworks offer algorithms to help write the learning program, and platforms to help train the machine learning model, making it easier for developers to build solutions for designing, training and validating algorithms and models for complex problems.
They play a more important role in ML development than in traditional software development.
ML Framework testing thus checks whether the frameworks of machine learning have bugs that may lead to problems in the final system~\cite{Pham2019cradle}.

\noindent \textbf{\textbf{Bug Detection in Learning Program}}. 
A learning program can be classified into two parts: the algorithm designed by the developer or chosen from the framework, and the actual code that developers write to implement, deploy, or configure the algorithm.
A bug in the learning program may arise either because the algorithm is designed, chosen, or configured improperly, or because the developers make typos or errors when implementing the designed algorithm.

 



\subsection{ML Testing Properties}
\label{sec:mltestingproblems}
Testing properties refer to \textbf{what} to test in \mlt{}: for what conditions \mlt{} needs to guarantee for a trained model.
This section lists some \jie{typical properties that the literature has considered}.
\jienew{We classified them into} basic functional requirements (i.e., correctness and model relevance) and non-functional requirements (i.e., efficiency, robustness\footnote{we adopt the more general understanding from software engineering community~\cite{chung2012non,afzal2009systematic}, and regard robustness as a non-functional requirement.}, fairness, interpretability).

These properties are not strictly independent of each other when considering the root causes, yet they are different external manifestations of the behaviours of an ML system and deserve being treated independently in \mlt{}.

\subsubsection{Correctness}

Correctness measures the probability that the ML system under test `gets things right'.

\begin{definition}[Correctness]
Let $\mathcal{D}$ be the distribution of future unknown data.
Let $x$ be a data item belonging to $\mathcal{D}$.
Let $h$ be the machine learning model that we are testing.
$h(x)$ is the predicted label of $x$, $c(x)$ is the true label.
The model correctness $E(h)$ is the probability that $h(x)$ and $c(x)$ are identical:
\begin{equation}
E(h) = Pr_{ x \sim \mathcal{D}}[h(x)=c(x)]
\end{equation}
\end{definition}

Achieving acceptable correctness is the fundamental requirement of an ML system. 
The real performance of an ML system should be evaluated on future data. Since future data are often not available, the current best practice usually splits the data into training data and test data (or training data, validation data, and test data), and uses test data to simulate future data.
This data split approach is called cross-validation.

\begin{definition}[Empirical Correctness]
Let $\mathcal{X}=(x_1,...,x_m)$ be the set of unlabelled test data sampled from $\mathcal{D}$.
Let $h$ be the machine learning model under test.
Let $\mathcal{Y'}=(h(x_1),...,h(x_m))$ be the set of predicted labels corresponding to each training item $x_i$. 
Let $\mathcal{Y}=(y_1,...,y_m)$ be the true labels, where each $y_i\in\mathcal{Y}$ corresponds to the label of $x_i\in\mathcal{X}$.
The empirical correctness of model (denoted as $\hat{E}(h)$) is:
\begin{equation}
    \hat{E}(h)= \frac{1}{m}\sum_{i=1}^{m}\mathbb{I}(h(x_i) = y_i)
\end{equation}

\end{definition}
where $\mathbb{I}$ is the indicator function;
a predicate returns 1 if $p$ is true, and returns 0 otherwise.


\subsubsection{Model Relevance}

A machine learning model comes from the combination of a machine learning algorithm and the training data.
It is important to ensure that
the adopted machine learning algorithm be not over-complex than just needed~\cite{kirk2014thoughtful}.
Otherwise, the model may fail to have good performance on future data, or have very large uncertainty.

\jienew{
The algorithm capacity
represents the number of functions that a machine learning model can select (based on the training data at hand) as a possible solution.
It is usually approximated by VC-dimension~\cite{vapnik1994measuring} or Rademacher Complexity~\cite{rosenberg2007rademacher} for classification tasks.
VC-dimension is the cardinality of the largest set of points that the algorithm can shatter.
Rademacher Complexity is the cardinality of the largest set of training data with fixed features that the algorithm shatters.
}

\jienew{We define the relevance between a machine learning algorithm capacity and the data distribution as the problem of \emph{model relevance}.}

\begin{definition}[Model Relevance]
Let $\mathcal{D}$ be the training data distribution.
Let $R(\mathcal{D}, \mathcal{A})$ be the simplest required capacity of any machine learning algorithm $\mathcal{A}$ for $\mathcal{D}$.
$R'(\mathcal{D},\mathcal{A'})$ is the capacity of the machine learning algorithm $\mathcal{A'}$ under test.
Model relevance is the difference between 
$R(\mathcal{D}, \mathcal{A})$ and $R'(\mathcal{D}, \mathcal{A'})$.
\begin{equation}
f = |(R(\mathcal{D}, \mathcal{A})-R'(\mathcal{D}, \mathcal{A'})|
\end{equation}	
\end{definition}

\jie{Model relevance aims to measure how well a machine learning algorithm fits the data. A low model relevance is usually caused by overfitting, where the model is too complex for the data, which thereby fails to generalise to future data or to predict observations robustly.}

Of course, $R'(\mathcal{D}, \mathcal{A})$, the minimum complexity that is `just sufficient' is hard to determine, and is typically approximate~\cite{werpachowski2019detecting,zhang:hal-02139208}.
We discuss more strategies that could help alleviate the problem of overfitting in Section~\ref{sec:overfitting}. 

\subsubsection{Robustness}

Robustness is defined by the IEEE standard glossary of software engineering terminology~\cite{159342,shahrokni2013systematic} as:
`The degree to which a system or component can function correctly in the presence of invalid inputs or stressful environmental conditions'.
Adopting a similar spirit to this definition, we define the robustness of ML as follows:

\begin{definition}[Robustness]
Let $S$ be a machine learning system. 
Let $E(S)$ be the correctness of $S$. 
Let $\delta(S)$ be the machine learning system with perturbations on any machine learning components such as the data, the learning program, or the framework.
The robustness of a machine learning system is a measurement of the difference between $E(S)$ and $E(\delta(S))$:
\begin{equation}
r = E(S)-E(\delta(S))
\end{equation}	
\end{definition}

Robustness thus measures the resilience of an ML system's correctness in the presence of perturbations.

A popular sub-category of robustness is called \emph{adversarial robustness}. 
For adversarial robustness, the perturbations are designed to be hard to detect.
Following the work of Katz et al.~\cite{katz2017reluplex}, we classify adversarial robustness into local adversarial robustness and global adversarial robustness.
Local adversarial robustness is defined as follows.

\begin{definition}[Local Adversarial Robustness]
Let $x$ a test input for an ML model $h$. 
Let $x'$ be another test input generated via conducting adversarial perturbation on $x$.
Model $h$ is $\delta$-local robust at input $x$ if for any $x'$. 
\begin{equation}
 \forall x': ||x-x'||_p \leq \delta \rightarrow h(x)=h(x') 
\end{equation}
\end{definition}

$||\cdot||_p$ represents $p$-norm for distance measurement. The commonly used $p$ cases in machine learning testing are 0, 2, and $\infty$. For example, when $p=2$, i.e.\, $||x-x'||_2$ represents the Euclidean distance of $x$ and $x'$.
In the case of $p=0$, it calculates the element-wise difference between $x$ and $x'$.
When $p=\infty$, it measures the the largest element-wise distance among all elements of $x$ and $x'$.

Local adversarial robustness concerns the robustness at one specific test input, while global adversarial robustness measures robustness against all inputs. 
We define global adversarial robustness as follows.

\begin{definition}[Global Adversarial Robustness]
Let $x$ a test input for an ML model $h$. 
Let $x'$ be another test input generated via conducting adversarial perturbation on $x$.
Model $h$ is $\epsilon$-global robust if for any $x$ and $x'$.
\begin{equation}
\forall x, x':  ||x-x'||_p \leq \delta \rightarrow h(x)-h(x') \leq \epsilon
\end{equation}
\end{definition}

\subsubsection{Security}

The security of an ML system is the system's resilience against potential harm, danger, or loss made via manipulating or illegally accessing ML components.

Security and robustness are closely related. 
An ML system with low robustness may be insecure:
if it is less robust in resisting the perturbations in the data to predict, the system may more easily fall victim to adversarial attacks;
For example,
if it is less robust in resisting training data perturbations, it may also be vulnerable to data poisoning (i.e., changes to the predictive behaviour caused by adversarially modifying the training data).

Nevertheless, low robustness is just one cause of security vulnerabilities.
Except for perturbations attacks, security issues also include other aspects such as model stealing or extraction. 
This survey focuses on the testing techniques on detecting ML security problems, which narrows the security scope to robustness-related security.  
We combine the introduction of robustness and security in Section~\ref{sec:robustness}.


\subsubsection{Data Privacy} 


Privacy in machine learning is the ML system's ability to preserve private data information. 
For the formal definition, we use the most popular \emph{differential privacy} taken from the work of Dwork~\cite{dwork2011differential}.

\begin{definition}[$\epsilon$-Differential Privacy]
Let $\mathcal{A}$ be a randomised algorithm.
Let $D_1$ and $D_2$ be two training data sets that differ only on one instance. 
Let $S$ be a subset of the output set of $\mathcal{A}$.
$\mathcal{A}$ gives $\epsilon$-differential privacy if 
\begin{equation}
Pr[\mathcal{A}(D_1)\in S] \leq exp(\epsilon)*Pr[\mathcal{A}(D_2)\in S]
\end{equation}
\end{definition}

In other words, $\epsilon$-Differential privacy is a form of $\epsilon$-contained bound on output change in responding to single input change. 
It provides a way to know whether any one individual's data has has a significant effect (bounded by $\epsilon$) on the outcome.

Data privacy has been regulated by law makers, for example, the EU General Data Protection Regulation (GDPR)~\cite{VoigtGDPR} and California Consumer Privacy Act (CCPA)~\cite{californiaCCPA}.
Current research mainly focuses on how to present privacy-preserving machine learning, instead of detecting privacy violations.
We discuss privacy-related research opportunities and research directions in Section~\ref{sec:researchdirection}.

\subsubsection{Efficiency}

The efficiency of a machine learning system refers to its construction or prediction speed. An efficiency problem happens when the system executes slowly or even infinitely during the construction or the prediction phase.

With the exponential growth of data and complexity of systems, efficiency is an important feature to consider for model selection and framework selection, sometimes even more important than accuracy~\cite{8258006}.
For example, to deploy a large model to a mobile device, optimisation, compression, and device-oriented customisation may be performed to make it feasible for the mobile device execution in a reasonable time, but accuracy may sacrifice to achieve this.



\subsubsection{Fairness}

Machine learning is a statistical method and is widely adopted to make decisions, such as income prediction and medical treatment prediction. 
Machine learning tends to learn what humans teach it (i.e., in form of training data). 
However, humans may have bias over cognition, further affecting the data collected or labelled and the algorithm designed, leading to bias problems. 

The characteristics that are sensitive and need to be protected against unfairness are called \emph{protected characteristics}~\cite{corbett2018measure} or \emph{protected attributes} and \emph{sensitive attributes}.
Examples of legally recognised protected classes include 
race, colour, sex, religion, national origin, citizenship, age, pregnancy, 
familial status, disability status, veteran status, and genetic information.

Fairness is often domain specific. 
Regulated domains include credit, education, employment, housing, and public accommodation\footnote{To prohibit discrimination `in a place of public accommodation on the basis of sexual orientation, gender identity, or gender expression'~\cite{days2004feedback}.}.

To formulate fairness is the first step to solve the fairness problems and build fair machine learning models.
The literature has proposed many definitions of fairness but no firm consensus is reached at this moment.
Considering that the definitions themselves are the research focus of fairness in machine learning,
we discuss how the literature formulates and measures different types of fairness in Section~\ref{sec:fairness}.

\subsubsection{Interpretability} 

Machine learning models are often applied to assist/make decisions in medical treatment, income prediction, or personal credit assessment.
It may be important for humans to understand the `logic' behind the final decisions, so that they can build trust over the decisions made by ML~\cite{lipton2016mythos,doshi2017towards,sellam2018deepbase}.

The motives and definitions of interpretability are diverse and still somewhat discordant~\cite{lipton2016mythos}.
Nevertheless, unlike fairness, a mathematical definition of ML interpretability remains elusive~\cite{doshi2017towards}.
Referring to the work of Biran and Cotton~\cite{biran2017explanation} as well as the work of Miller~\cite{miller2018explanation}, \jienew{we describe the interpretability of ML as the degree to which an observer can understand the cause of a decision made by an ML system.}

Interpretability contains two aspects: transparency (how the model works) and post hoc explanations (other information that could be derived from the model)~\cite{lipton2016mythos}.
Interpretability is also regarded as a request by regulations like GDPR~\cite{goodman2016european}, where the user has the legal `right to explanation' to ask for an explanation of an algorithmic decision that was made about them.
A thorough introduction of ML interpretability can be referred to in the book of Christoph~\cite{molnar}.




\subsection{Software Testing vs. ML Testing}

Traditional software testing and \mlt{} are different in many aspects. 
To understand the unique features of \mlt{}, we summarise the primary differences between traditional software testing and \mlt{} in Table~\ref{tab:compare}.

\begin{table*}[h!]\small	
	\center
	\caption{\label{tab:compare}Comparison between Traditional Software Testing and ML Testing}
	\begin{tabular}{p{4.5cm} p{3.7cm} p{6.8cm}}
		\toprule
		\textbf{Characteristics}&\textbf{Traditional Testing} &\textbf{ML Testing}\\ \midrule 
		Component to test&code&data and code (learning program, framework)\\
		Behaviour under test&usually fixed&change overtime\\
		Test input&input data&data or code\\
		Test oracle&\jienew{defined by developers}&\jienew{defined by developers and labelling companies}\\
		Adequacy criteria&coverage/mutation score&unknown\\
		False positives in bugs&rare&prevalent\\
		Tester&developer&data scientist, algorithm designer, developer\\
		\bottomrule		
	\end{tabular}
\end{table*}

\noindent \textbf{1) Component to test} (where the bug may exist): traditional software testing detects bugs in the code, while \mlt{} detects bugs in the data, the learning program, and the framework, each of which play an essential role in building an ML model.

\noindent \textbf{2) Behaviours under test}: the behaviours of traditional software code are usually fixed once the requirement is fixed, while the behaviours of an ML model may frequently change as the training data is updated.

\noindent \textbf{3) Test input}: the test inputs in traditional software testing are usually the input data when testing code; in ML testing, however, the test inputs in may have more diverse forms. 
Note that we separate the definition of `test input' and `test data'.
\jie{In particular, we use `test input' to refer to the inputs in any form that can be adopted to conduct machine learning testing; while `test data' specially refers to the data used to validate ML model behaviour (see more in Section~\ref{sec:preliminaries}).
Thus,} test inputs in \mlt{} could be, but are not limited to, test data.
When testing the learning program, a test case may be a single test instance from the test data or a toy training set;
when testing the data, the test input could be a learning program.

\noindent \textbf{4) Test oracle}: traditional software testing usually assumes the presence of a test oracle. 
The output can be verified against the expected values by the developer, and thus the oracle is usually determined beforehand.
Machine learning, however, is used to generate answers based on a set of input values \jie{after being deployed online}.
\jie{The correctness of the large number of generated answers is typically manually confirmed. Currently, the identification of test oracles remains challenging, because many desired properties are difficult to formally specify. Even for a concrete domain specific problem, the oracle identification is still time-consuming and labour-intensive, because domain-specific knowledge is often required. 
In current practices, companies usually rely on third-party data labelling companies to get manual labels, which can be expensive.
}
\jie{Metamorphic relations~\cite{zhang2014search} are a type of pseudo oracle adopted to \jie{automatically} mitigate the oracle problem in machine learning testing.}

\noindent \textbf{5) Test adequacy criteria}:
test adequacy criteria are used to provide quantitative measurement on the degree of the target software that has been tested. Up to present, many adequacy criteria are proposed and widely adopted in industry, e.g., line coverage, branch coverage, dataflow coverage. However, due to fundamental differences of programming paradigm and logic representation format for machine learning software and traditional software, new test adequacy criteria are required to take the characteristics of machine learning software into consideration.


\noindent \textbf{6) False positives in detected bugs}: due to the difficulty in obtaining reliable oracles, \mlt{} tends to yield more false positives in the reported bugs. 

\noindent \textbf{7) Roles of testers}: the bugs in \mlt{} may exist not only in the learning program, but also in the data or the algorithm, and thus data scientists or algorithm designers could also play the role of testers.

\section{Paper Collection and Review Schema}
\label{sec:schema}

This section introduces the scope, the paper collection approach, an initial analysis of the collected papers, and the organisation of our survey.

\subsection{Survey Scope}
\label{sec:surveyscope}

An ML system may include both hardware and software. 
The scope of our paper is software testing (as defined in the introduction) applied to machine learning.

We apply the following inclusion criteria when collecting papers.
If a paper satisfies any one or more of the following criteria, we will include it.
When speaking of related `aspects of ML testing', we refer to the ML properties, ML components, and ML testing procedure introduced in Section~~\ref{sec:preliminaries}.

\noindent 1) The paper introduces/discusses the general idea of ML testing or one of the related aspects of ML testing.

\noindent 2) The paper proposes an approach, study, or tool/framework that targets testing one of the ML properties or components.

\noindent 3) The paper presents a dataset or benchmark especially designed for the purpose of ML testing.

\noindent 4) The paper introduces a set of measurement criteria that could be adopted to test one of the ML properties.

Some papers concern traditional validation of ML model performance such as the introduction of precision, recall, and cross-validation.
We do not include these papers because they have had a long research history and have been thoroughly and maturely studied.
Nevertheless, for completeness, we include the knowledge when introducing the background to set the context.
We do not include the papers that adopt machine learning techniques for the purpose of traditional software testing and also those target ML problems, which do not use testing techniques as a solution. 

\jienew{
Some recent papers also target the formal guarantee on the desired properties of a machine learning system, i.e., to formally verify the correctness of the machine learning systems as well as other properties.
Testing and verification of machine learning, as in traditional testing, have their own advantages and disadvantages. For example, verification usually requires a white-box scenario,
but suffers from poor scalability, while testing may scale, but lacks completeness.
The size of the space of potential behaviours may render current approaches to verification infeasible in general~\cite{o2018scalable}, but specific safety critical properties will clearly benefit from focused research activity on scalable verification, as well as testing.
In this survey, we focus on the machine learning testing.
More details for the literature review of the verification of machine learning systems can be found in the recent work of Xiang et al.~\cite{xiang2018verification}.
}

\subsection{Paper Collection Methodology}
\label{sec:papercollection}

To collect the papers across different research areas as much as possible,
we started by using exact keyword searching on popular scientific databases including Google Scholar, DBLP and arXiv one by one.
The keywords used for searching are listed below.
$[$ML properties$]$ means the set of ML testing properties including correctness, 
model relevance, robustness, efficiency, privacy, fairness, and interpretability.
We used each element in this set plus `test' or `bug' as the search query.
Similarly, $[$ML components$]$ denotes the set of ML components including data, learning program/code, and framework/library.
Altogether, we conducted $(3*3+6*2+3*2)*3=81$ searches across the three repositories before May 15th, 2019.

\begin{itemize}
	\item machine learning + test|bug|trustworthiness 
	\item deep learning + test|bug|trustworthiness
	\item neural network + test|bug|trustworthiness
	\item $[$ML properties$]$+ test|bug 
	\item $[$ML components$]$+ test|bug 	
\end{itemize}

Machine learning techniques have been applied in various domains across different research areas.
As a result, authors may tend to use very diverse terms.  
To ensure a high coverage of \mlt{} related papers,
we therefore also performed snowballing~\cite{wohlin2014guidelines} on each of the related
papers found by keyword searching. 
We checked the related work sections in these studies and continue adding the related work that satisfies the inclusion criteria introduced in Section~\ref{sec:surveyscope}, until we reached closure.

To ensure a more comprehensive and accurate survey, we emailed the authors of the papers that were collected via query and snowballing,
and let them send us other papers they are aware of which are related with machine learning testing but have not been included yet. 
We also asked them to check whether our description about their work in the survey was accurate and correct.

\begin{table}[t]\small	
	\center
	\caption{\label{tab:collectionresult}Paper Query Results}
	\begin{tabular}{p{5cm}rrr}
		\toprule
		\textbf{Key Words} & \textbf{Hits} & \textbf{Title} & \textbf{Body}\\ \midrule 
		machine learning test&211&17&13\\
		machine learning bug&28&4&4\\
		machine learning trustworthiness&1&0&0\\
		deep learning test&38&9&8\\
		deep learning bug&14&1&1\\
		deep learning trustworthiness&2&1&1\\
		neural network test&288&10&9\\
		neural network bug&22&0&0\\
		neural network trustworthiness&5&1&1\\
		$[$ML properties$]$+test&294&5&5\\
 		$[$ML properties$]$+bug&10&0&0\\  
 		$[$ML components$]$+test&77&5&5\\
 		$[$ML components$]$+bug&8&2&2\\ \midrule
 		Query&-&-&50\\
 		Snowball&-&-&59\\
 		Author feedback&-&-&35\\
 		\textbf{Overall}&-&-&\papernum\\
		
		\bottomrule		
	\end{tabular}
\end{table}

\subsection{Collection Results}

Table~\ref{tab:collectionresult} shows the details of paper collection results.
The papers collected from Google Scholar and arXiv turned out to be subsets of those from DBLP so we only present the results of DBLP.
Keyword search and snowballing resulted in 109 papers across six research areas till May 15th, 2019. 
We received over 50 replies from all the cited authors until June 4th, 2019, and added another 35 papers when dealing with the author feedback.
Altogether, we collected \papernum papers.

\begin{figure}[h!]
  \centering
\includegraphics[scale=0.37]{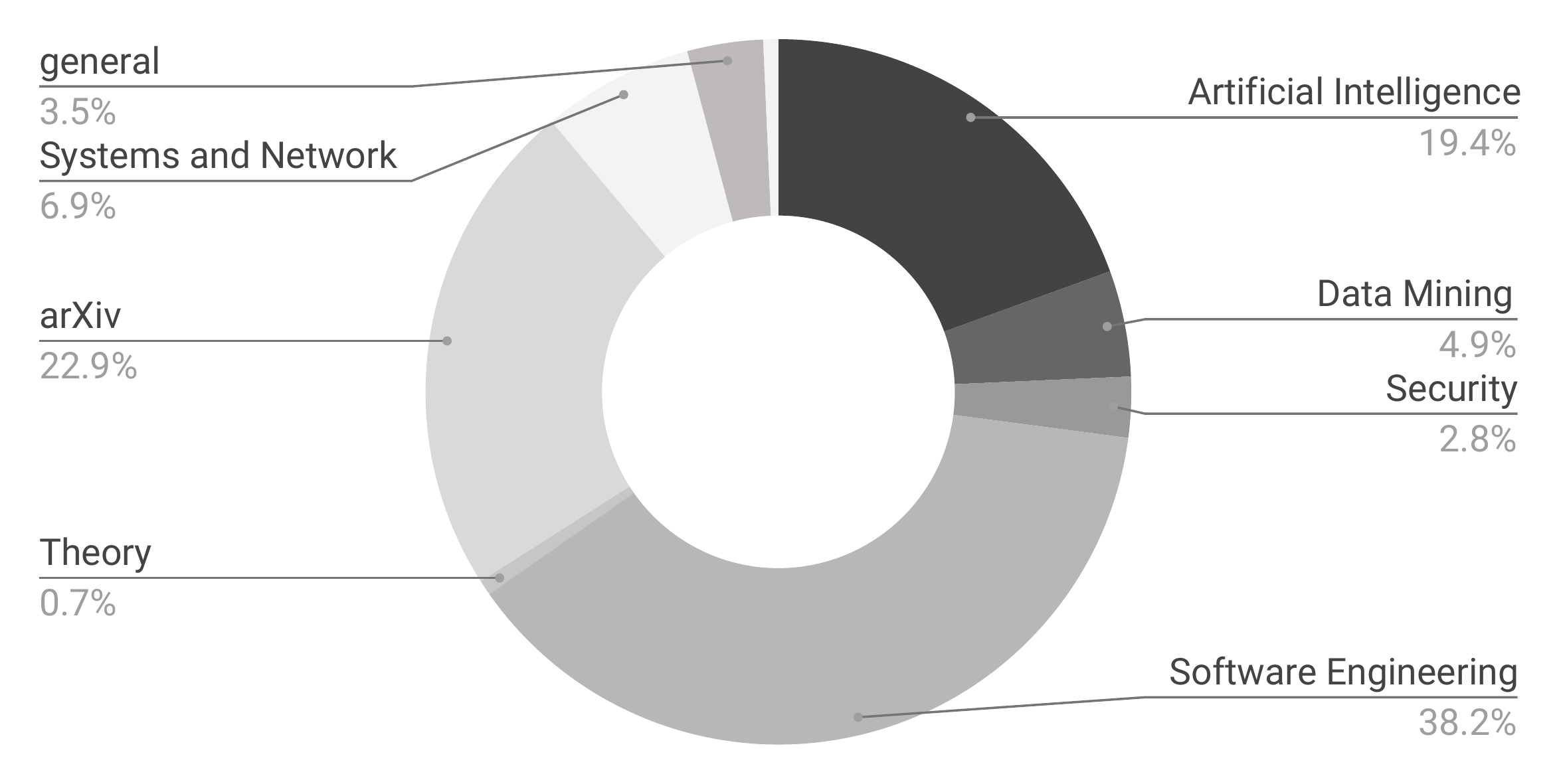}
\caption{Publication Venue Distribution}
  \label{fig:venues}
\end{figure}

Figure~\ref{fig:venues} shows the distribution of papers published in different research venues.
Among all the papers, 38.2\% papers are published in software engineering venues such as ICSE, FSE, ASE, ICST, and ISSTA;
6.9\% papers are published in systems and network venues;
surprisingly, only 19.4\% of the total papers are published in artificial intelligence venues such as AAAI, CVPR, and ICLR. 
Additionally, 22.9\% of the papers have not yet been published via peer-reviewed venues (the arXiv part).


\subsection{Paper Organisation}

We present the literature review from two aspects:
1) a literature review of the collected papers,
2) a statistical analysis of the collected papers, datasets, and tools.
The sections and the corresponding contents are presented in Table~\ref{tab:schema}.

\begin{table}[!h]\small
	
	\center
	\caption{\label{tab:schema}Review Schema}
	\resizebox{.48\textwidth}{!}{
	\begin{tabular}{p{2.5cm} c l}
		\toprule
		\textbf{Classification}&\textbf{Sec}&\textbf{Topic}\\ \midrule

		\multirow{6}{*}{Testing Workflow}
		&\ref{sec:input}&Test Input Generation\\
		&\ref{sec:oracle}&Test Oracle\\
		&\ref{sec:evaluation}&Test Adequacy\\
		&\ref{sec:testprioritisation}& Test Prioritisation and Reduction\\
		&\ref{sec:bugreport}&Bug Report Analysis\\
		&\ref{sec:debug}&Debug and Repair\\ 
		&\ref{sec:generalframework}&Testing Framework and Tools\\ 
		\midrule
		
				\multirow{8}{*}{Testing Properties}&
		\ref{sec:effectiveness}&Correctness\\
		&\ref{sec:overfitting}&Model Revelance\\
		&\ref{sec:robustness}&Robustness and Security\\
		&\ref{sec:efficiency}&Efficiency\\ 
		&\ref{sec:fairness}&Fairness \\
		&\ref{sec:inter}&Interpretability\\
		&\ref{sec:privacy}&Privacy\\
		\midrule
		
			\multirow{5}{*}{Testing Components}&
		\ref{sec:datatesting}&Bug Detection in Data\\
		&\ref{sec:trainingprogramtesting}&Bug Detection in Learning Program\\
		&\ref{sec:frameworktesting}&Bug Detection in Framework\\
		\midrule
		
		\multirow{3}{*}{Application Scenario}&
		\ref{sec:autodriving}&Autonomous Driving\\
		&\ref{sec:machinetranslation}&Machine Translation\\
		&\ref{sec:NLI}&Natural Language Inference\\
		\midrule

		\multirow{5}{*}{Summary\&Analysis}
		&\ref{sec:timeline}&Timeline\\
		&\ref{sec:type}&Research Distribution among Categories\\
		&\ref{sec:distributionproperty}&Research Distribution among Properties\\
		&\ref{sec:dataset}&Datasets\\
		&\ref{sec:tooltechnique}&Open-source Tool Support\\
		
		\bottomrule		
	\end{tabular}
	}
\end{table} 

\noindent\textbf{1) Literature Review}.
The papers in our collection are organised and presented from four angles.
We introduce the work about different testing workflow in Section~\ref{sec:testingworkflow}.
In Section~\ref{sec:testingproperties} we classify the papers based on the ML problems they target, including functional properties like correctness and model relevance and non-functional properties like robustness, fairness, privacy, and interpretability.
Section~\ref{sec:testingcomponents} introduces the testing technologies on detecting bugs in data, learning programs, and ML frameworks, libraries, or platforms.  
Section~\ref{sec:applicationsenario} introduces the testing techniques applied in particular application scenarios such as autonomous driving and machine translation.

The four aspects have different focuses of \mlt{}, each of which is a complete organisation of the total collected papers (see more discussion in Section~\ref{sec:definition}), as a single \mlt{} paper may fit multiple aspects if being viewed from different angles.


\noindent\textbf{2)  Statistical Analysis and Summary}.
We analyse and compare the number of research papers on different machine learning categories (supervised/unsupervised/reinforcement learning), machine learning structures (classic/deep learning), testing properties in Section~\ref{sec:analysissummary}.
We also summarise the datasets and tools adopted so far in ML testing.


The four different angles for presentation of related work as well as the statistical summary, analysis, and comparison, 
enable us to observe the research focus, trend, challenges, opportunities, and directions of \mlt{}.
These results are presented in Section~\ref{sec:researchdirection}.


\section{ML Testing Workflow}
\label{sec:testingworkflow}

This section organises \mlt{} research based on the testing workflow as shown by Figure~\ref{fig:testworkflow}.


\mlt{} includes offline testing and online testing.
\jie{Albarghouthi and Vinitsky~\cite{albarghouthi2019fairness} developed a fairness specification language that can be used for the development of run-time monitoring,  in detecting fairness issues. Such a kind of run-time monitoring belongs to the area of online testing.}
Nevertheless, current research mainly centres on offline testing as introduced below. 
The procedures that are not covered based on our paper collection, such as requirement analysis and regression testing and those belonging to online testing
are discussed as research opportunities in Section~\ref{sec:researchdirection}.

%

\subsection{Test Input Generation}
\label{sec:input}

We organise the test input generation research based on the techniques adopted.

\subsubsection{Domain-specific Test Input Synthesis}


Test inputs of \mlt{} can be classified into two categories: adversarial inputs and natural inputs.
Adversarial inputs are perturbed based on the original inputs. 
They may not belong to normal data distribution (i.e., maybe rarely exist in practice), but could expose robustness or security flaws. 
Natural inputs, instead, are those inputs that belong to the data distribution of a practical application scenario.
Here we introduce the related work that aims to generate natural inputs via domain-specific test input synthesis.



DeepXplore~\cite{pei2017deepxplore} proposed a white-box differential testing technique to generate test inputs for a deep learning system.
Inspired by test coverage in traditional software testing, the authors proposed neuron coverage to drive test generation (we discuss different coverage criteria for \mlt{} in Section~\ref{sec:modelevaluationmetrics}).
The test inputs are expected to have high neuron coverage.
Additionally, the inputs need to expose differences among different DNN models, as well as be like real-world data as much as possible.
The joint optimisation algorithm iteratively uses a gradient search to find a modified input that satisfies all of these goals.
The evaluation of DeepXplore indicates that it covers 34.4\% and 33.2\% more neurons than the same number of randomly picked inputs and adversarial inputs.

To create useful and effective data for autonomous driving systems, \emph{DeepTest}~\cite{tian2018deeptest} performed greedy search with nine different realistic image transformations: changing brightness, changing contrast, translation, scaling, horizontal shearing, rotation, blurring, fog effect, and rain effect.
There are three types of image transformation styles provided in OpenCV\footnote{\url{https://github.com/itseez/opencv}(2015)}: linear, affine, and convolutional.
The evaluation of DeepTest uses the Udacity self-driving car challenge dataset~\cite{udacity}.
It detected more than 1,000 erroneous behaviours on CNNs and RNNs with low false positive rates\footnote{The examples of detected erroneous behaviours are available at \url{https://deeplearningtest.github.io/deepTest/}.}.

Generative adversarial networks (GANs)~\cite{goodfellow2014generative} are algorithms to generate models that approximate the manifolds and distribution on a given set of data. GAN has been successfully applied to advanced image transformation~(e.g., style transformation, scene transformation) that look at least superficially authentic to human observers.
Zhang et al.~\cite{deeproad} applied GAN to deliver driving scene-based test generation with various weather conditions.
They sampled images from Udacity Challenge dataset~\cite{udacity} and YouTube videos (snowy or rainy scenes), and fed them into the UNIT framework\footnote{A recent
DNN-based method to perform image-to-image transformation~\cite{liu2017unsupervised}} for training. 
The trained model takes the whole Udacity images as the seed inputs and yields transformed images as generated tests.

Zhou et al.~\cite{zhou2018deepbillboard} proposed \emph{DeepBillboard} to generate real-world adversarial billboards that can trigger potential steering errors of autonomous driving systems.

To test audio-based deep learning systems, Du et al.~\cite{du2018deepcruiser} designed a set of transformations tailored to audio inputs considering background noise and volume variation. They first abstracted and extracted a probabilistic transition model from an RNN. Based on this, stateful testing criteria are defined and used to guide test generation for stateful machine learning system.

To test the image classification platform when classifying biological cell images, Ding et al.~\cite{ding2016framework} built a testing framework for biological cell classifiers.
The framework iteratively generates new images and uses metamorphic relations for testing.
For example, they generate new images by increasing the number/shape of artificial mitochondrion into the biological cell images, which can arouse easy-to-identify changes in the classification results.

Rabin et al.~\cite{rabin2019testing} discussed the possibilities of testing code2vec (a code embedding approach~\cite{alon2019code2vec}) with semantic-preserving program transformations serving as test inputs.

To test machine translation systems, Sun et al.~\cite{sun2019automatic} automatically generate test inputs via mutating the words in translation inputs.
In order to generate translation pairs that ought to yield consistent translations, their approach conducts word replacement based on word embedding similarities.
Manual inspection indicates that the test generation has a high precision (99\%) on generating input pairs with consistent translations.

\subsubsection{Fuzz and Search-based Test Input Generation}



Fuzz testing is a traditional automatic testing technique that generates random data as program inputs to detect crashes, memory leaks, failed (built-in) assertions, etc, with many sucessfully application to system security and vulnerability detection~\cite{barr2015oracle}.
As another widely used test generation technique, search-based test generation often uses metaheuristic search techniques to guide the fuzz process for more efficient and effective test generation~\cite{mcminn2004search,harman2001search,lakhotia2007multi}.
These two techniques have also been proved to be effective in exploring the input space of \mlt{}:

Odena et al.~\cite{odena2018tensorfuzz} presented \emph{TensorFuzz}.
\emph{TensorFuzz} used a simple nearest neighbour hill climbing approach to explore achievable coverage over valid input space for Tensorflow graphs, and to discover numerical errors, disagreements between neural networks and their quantized versions, and surfacing undesirable behaviour in RNNs.

DLFuzz, proposed by Guo et al.~\cite{guo2018dlfuzz}, is another fuzz test generation tool based on the implementation of DeepXplore with nueron coverage as guidance.
DLFuzz aims to generate adversarial examples.
The generation process thus does not require similar functional deep learning systems for
cross-referencing check like DeepXplore and TensorFuzz.
Rather, it needs only minimum changes over the original inputs to find those new inputs that improve neural coverage but have different predictive results from the original inputs.
The preliminary evaluation on MNIST and ImageNet shows that compared with DeepXplore, DLFuzz is able to generate 135\% to 584.62\% more inputs with 20.11\% less time consumption.


Xie et al.~\cite{xie2018coverage} presented a metamorphic transformation based coverage guided fuzzing technique, DeepHunter, which leverages both neuron coverage and coverage criteria  presented by DeepGauge~\cite{Ma2018ase}.
DeepHunter uses a more fine-grained metamorphic mutation strategy to generate tests, which demonstrates the advantage in reducing the false positive rate. It also demonstrates its advantage in achieving high coverage and bug detection capability. 



Wicker et al.~\cite{wicker2018feature} proposed feature-guided test generation.
They adopted Scale-Invariant Feature Transform (SIFT) to identify features that represent an image with a Gaussian mixture model, then transformed the problem of finding adversarial examples into a two-player turn-based stochastic game.
They used Monte Carlo Tree Search to identify those elements of an image most vulnerable as the means of generating adversarial examples.
The experiments show that their black-box approach is competitive with some state-of-the-art white-box methods.

Instead of targeting supervised learning, Uesato et al.~\cite{uesato2018rigorous} proposed to evaluate reinforcement learning with adversarial example generation.
The detection of catastrophic failures is expensive because failures are rare. 
To alleviate the consequent cost of finding such failures, the authors proposed to use a failure probability predictor to estimate the probability that the agent fails, which was demonstrated to be both effective and efficient.

There are also fuzzers for specific application scenarios other than image classifications.
Zhou et al.~\cite{zhou2019metamorphic} 
combined fuzzing and metamorphic testing to test the LiDAR obstacle-perception module of real-life self-driving cars, and reported previously unknown software faults.
Jha et al.~\cite{jhaml2019mlbased} investigated how to generate the most effective test cases (the faults that are most likely to lead to violations of safety conditions) via modelling the fault injection as a Bayesian network. 
The evaluation, based on two production-grade AV systems from NVIDIA and Baidu, revealed many situations where faults lead to safety violations. 

Udeshi and Chattopadhyay~\cite{udeshi2019grammar} generate inputs for text classification tasks and produce a fuzzing approach that considers the grammar under test as well as the distance between inputs.
Nie et al.~\cite{nie2018analyzing} and Wang et al.~\cite{wang2018if} mutated the sentences in NLI (Natural Language Inference) tasks to generate test inputs for robustness testing.
Chan et al.~\cite{chan2018metamorphic} generated adversarial examples for DNC to expose its robustness problems.
Udeshi et al.~\cite{udeshi2018automated} focused much on individual fairness and generated test inputs that highlight the discriminatory nature of the model under test.
We give details about these domain-specific fuzz testing techniques in Section~\ref{sec:applicationsenario}.

Tuncali et al.~\cite{tuncali2018simulation} proposed a framework for testing autonomous driving systems.
In their work they compared three test generation strategies: 
random fuzz test generation,
covering array~\cite{hartman2005software} $+$ fuzz test generation,
and covering array $+$ search-based test generation (using Simulated Annealing algorithm~\cite{kirkpatrick1983optimization}).
The results indicated that the test generation strategy with search-based technique involved has the best performance in detecting glancing behaviours.


\subsubsection{Symbolic Execution Based Test Input Generation}

Symbolic execution is a program analysis technique to test
whether certain properties can be violated by the software under test~\cite{king1976symbolic}. 
Dynamic Symbolic Execution (DSE, also called concolic testing) is a technique used to automatically generate test inputs that achieve high code coverage. 
DSE executes the program under test with random test inputs and performs symbolic execution in parallel to collect symbolic constraints obtained from predicates in branch statements along the execution traces. 
The conjunction of all symbolic constraints along a path is called a path condition. 
When generating tests, DSE randomly chooses one test input from the input domain,
then uses constraint solving to reach a target branch condition in the path~\cite{zhang2010test}.
DSE has been found to be accurate and effective, and has been the primary technique used by some vulnerability discovery tools~\cite{chen2013state}.

In ML testing, the model's performance is decided, not only by the code, but also by the data, and thus symbolic execution has two application scenarios: either on the data or on the code.

Symbolic analysis was applied to generate more effective tests to expose bugs by Ramanathan and Pullum~\cite{Ramanathan2016}.
They proposed a combination of symbolic and statistical approaches to efficiently find test cases.
The idea is to distance-theoretically abstract the data using symbols to help search for those test inputs where minor changes in the input will cause the algorithm to fail.
The evaluation of the implementation of a $k$-means algorithm indicates that the approach is able to detect subtle errors such as bit-flips. 
The examination of false positives may also be a future research interest.

When applying symbolic execution on the machine learning code, there are many challenges.
Gopinath~\cite{gopinath2018symbolic} listed three such challenges for neural networks in their paper, which work for other ML modes as well:
(1) the networks have no explicit branching;
(2) the networks may be highly non-linear, with no well-developed solvers for constraints; and 
(3) there are scalability issues because the structures of the ML models are usually very complex and are beyond the capabilities of current symbolic reasoning tools.

Considering these challenges, Gopinath~\cite{gopinath2018symbolic} introduced DeepCheck.
It transforms a Deep Neural Network (DNN) into a program to enable symbolic execution to find pixel attacks that have the same activation pattern as the original image.
In particular, the activation functions in DNN follow an IF-Else branch structure, which can be viewed as a path through the translated program.
DeepCheck is able to create 1-pixel and 2-pixel attacks by identifying most of the pixels or pixel-pairs that the neural network fails to classify the corresponding modified images.

Similarly, Agarwal et al.~\cite{agarwal2018automated} apply LIME~\cite{ribeiro2016should}, a local explanation tool that approximates a model with linear models, decision trees, or falling rule lists, to help get the path used in symbolic execution.
Their evaluation based on 8 open source fairness benchmarks shows that the algorithm generates 3.72 times more successful test cases than the random test generation approach THEMIS~\cite{galhotra2017fairness}.

Sun et al.~\cite{Sunase2018} presented DeepConcolic, 
a dynamic symbolic execution testing method for DNNs.
Concrete execution is used to direct the symbolic analysis to particular MC/DC criteria' condition, through concretely evaluating given properties of the ML models. 
DeepConcolic explicitly takes coverage requirements as input.
The authors report that it yields over 10\% higher neuron coverage than DeepXplore for the evaluated models.

\subsubsection{Synthetic Data to Test Learning Program}

Murphy et al.~\cite{Murphy2007PRT} generated data with repeating values, missing values, or categorical data for testing two ML ranking applications.
Breck et al.~\cite{breck2019datavalidation} used synthetic training data that adhere to schema constraints to trigger the hidden assumptions in the code that do not agree with the constraints.
Zhang et al.~\cite{zhang:hal-02139208} used synthetic data with known distributions to test overfitting.
Nakajima and Bui~\cite{nakajima2016dataset} also mentioned the possibility of generating simple datasets with some predictable characteristics that can be adopted as pseudo oracles.

\subsection{Test Oracle}
\label{sec:oracle}

Test oracle identification is one of the key problems in \mlt{}.
It is needed in order to enable the judgement of whether a bug exists.
This is the so-called `Oracle Problem'~\cite{barr2015oracle}. 

In \mlt{}, the oracle problem is challenging, because many machine learning algorithms are probabilistic programs. 
In this section, we list several popular types of test oracle that have been studied for ML testing, i.e., metamorphic relations, cross-referencing, and model evaluation metrics.


\subsubsection{Metamorphic Relations as Test Oracles}
\label{sec:metamorphicrelations}

Metamorphic relations was proposed by Chen et al.~\cite{chen1998metamorphic} to ameliorate the test oracle problem in traditional software testing.
A metamorphic relation refers to the relationship between the software input change and output change during multiple program executions.
For example, to test the implementation of the function $sin(x)$, one may check how the function output changes when the input is changed from $x$ to $\pi -x$. 
If $sin(x)$ differs from $sin(\pi - x)$, this observation signals an error without needing to examine the specific values computed by the implementation.
$sin(x) = sin(\pi - x)$ is thus a metamorphic relation that plays the role of test oracle (also named `pseudo oracle') to help bug detection.

In \mlt{}, metamorphic relations are widely studied to tackle the oracle problem.
Many metamorphic relations are based on transformations of training or test data that are expected to yield unchanged or certain expected changes in the predictive output.
There are different granularities of data transformations when studying the corresponding metamorphic relations.
Some transformations conduct coarse-grained changes such as enlarging the dataset or changing the data order, without changing each single data instance. 
We call these transformations `Coarse-grained data transformations'.
Some transformations conduct data transformations via smaller changes on each data instance, such as mutating the attributes, labels, or pixels of images, and are referred to as `fine-grained' data transformations in this paper.
The related works of each type of transformations are introduced below.

\noindent \textbf{Coarse-grained Data Transformation}.
As early as in 2008, Murphy et al.~\cite{Murphy2008PropertiesOM} discuss the properties of machine learning algorithms that may be adopted as metamorphic relations.
Six transformations of input data are introduced: additive, multiplicative, permutative, invertive,
inclusive, and exclusive.
The changes include adding a constant to numerical values; 
multiplying numerical values by a constant; 
permuting the order of the input data; 
reversing the order of the input data; 
removing a part of the input data; 
adding additional data.
Their analysis is on MartiRank, SVM-Light, and PAYL.
Although unevaluated in the initial 2008 paper, this work provided a foundation for determining the relationships and transformations that can be used for conducting metamorphic testing for machine learning.

Ding et al.~\cite{Ding2017} proposed 11 metamorphic relations to test deep learning systems.
At the dataset level, the metamorphic relations were also based on training data or test data transformations that were not supposed to affect classification accuracy, such as adding 10\% training images into each category of the training data set or removing one category of data from the dataset.
The evaluation is based on a classification of biological cell images.

Murphy et al.~\cite{murphy2009UJR} presented function-level metamorphic relations. The evaluation on 9 machine learning applications indicated that functional-level properties were 170\% more effective than application-level properties.

\noindent \textbf{Fine-grained Data Transformation}.
In 2009, Xie et al.~\cite{xie2009application} proposed to use metamorphic relations that were specific to a certain model to test the implementations of supervised classifiers.
The paper presents five types of metamorphic relations that enable the prediction of expected changes to the output (such as changes in classes, labels, attributes) based on particular changes to the input.
Manual analysis of the implementation of KNN and Naive Bayes from Weka~\cite{hall2009weka} indicates that not all metamorphic relations are necessary.
The differences in the metamorphic relations between SVM and neural networks are also discussed in ~\cite{nakajima2017generalized}.
Dwarakanath et al.~\cite{Dwarakanathissta2018} applied metamorphic relations to image classifications with SVM and deep learning systems.
The changes on the data include changing the feature or instance orders, linear scaling of the test features, normalisation or scaling up the test data, or changing the convolution operation order of the data. 
The proposed MRs are able to find 71\% of the injected bugs.
Sharma and Wehrheim~\cite{arnab2019balancedata} considered fine-grained data transformations such as changing feature names, renaming feature values to test fairness. They studied 14 classifiers, none of them were found to be sensitive to feature name shuffling.

Zhang et al.~\cite{zhang:hal-02139208} proposed Perturbed Model Validation (PMV) which combines metamorphic relation and data mutation to detect overfitting.
PMV mutates the training data via injecting noise in the training data to create perturbed training datasets, then checks the training accuracy decrease rate when the noise degree increases.
The faster the training accuracy decreases, the less the machine learning model overfits.

Al-Azani and Hassine~\cite{al2017validation} studied the metamorphic relations of Naive Bayes, $k$-Nearest Neighbour, as well as their ensemble classifier.
It turns out that the metamorphic relations necessary for Naive Bayes and k-Nearest Neighbour may be not necessary for their ensemble classifier.

Tian et al.~\cite{tian2018deeptest} and Zhang et al.~\cite{deeproad} stated that the autonomous vehicle steering angle should not change significantly or stay the same for the transformed images under different weather conditions. 
Ramanagopal et al.~\cite{ramanagopal2018failing} used the classification consistency of similar images to serve as test oracles for testing self-driving cars. 
The evaluation indicates a precision of 0.94 when detecting errors in unlabelled data.

Additionally, Xie et al.~\cite{xie2018mettle} proposed METTLE, a metamorphic testing approach for unsupervised learning validation.
METTLE has six types of different-grained metamorphic relations that are specially designed for unsupervised learners.
These metamorphic relations manipulate instance order, distinctness, density, attributes, or inject outliers of the data. The evaluation was based on synthetic data generated by Scikit-learn, showing that METTLE is practical and effective in validating unsupervised learners.
Nakajima et al.~\cite{nakajima2016dataset,nakajima2018dataset} discussed the possibilities of using different-grained metamorphic relations to find problems in SVM and neural networks, such as to manipulate instance order or attribute order and to reverse labels and change attribute values, or to manipulate the pixels in images.

\noindent \textbf{Metamorphic Relations between Different Datasets}.
The consistency relations between/among different datasets can also be regarded as metamorphic relations that could be applied to detect data bugs. 
Kim et al.~\cite{kim2018guiding} and Breck et al.~\cite{breck2019datavalidation} studied the metamorphic relations between training data and new data. If the training data and new data have different distributions, the training data may not be adequate.
Breck et al.~\cite{breck2019datavalidation} also studied the metamorphic relations among different datasets that are close in time: these datasets are expected to share some characteristics because it is uncommon to have frequent drastic changes to the data-generation code.

\noindent \textbf{Frameworks to Apply Metamorphic Relations}.
Murphy et al.~\cite{Murphy2009AST} implemented a framework called Amsterdam to automate the process of using metamorphic relations to detect ML bugs. 
The framework reduces false positives via setting thresholds when doing result comparison.
They also developed Corduroy~\cite{murphy2009UJR}, which extended Java Modelling Language to let developers specify metamorphic properties and generate test cases for ML testing.

We introduce more related work on domain-specific metamorphic relations of testing autonomous driving, Differentiable Neural Computer (DNC)~\cite{chan2018metamorphic}, machine translation systems~\cite{sun2019automatic,Zhou2018metamorphic,pesu2018monte}, biological cell classification ~\cite{ding2016framework}, and audio-based deep learning systems~\cite{du2018deepcruiser} in Section~\ref{sec:applicationsenario}.

\subsubsection{Cross-Referencing as Test Oracles}
\label{sec:oracle_differential}

Cross-Referencing is another type of test oracle for ML testing, including differential Testing and N-version Programming.
Differential testing is a traditional software testing technique that detects bugs by observing whether similar applications yield different outputs regarding identical inputs~\cite{mckeeman1998differential,Davis1981}.
It is a testing oracle for detecting compiler bugs~\cite{le2014compiler}.
According to the study of Nejadgholi and Yang~\cite{nejadgholistudy}, 5\% to 27\% test oracles for deep learning libraries use differential testing.

Differential testing is closely related with N-version programming~\cite{avizienis1995methodology}: N-version programming aims to generate multiple functionally-equivalent programs based on one specification, so that the combination of different versions are more fault-tolerant and robust.

Davis and Weyuker~\cite{Davis1981} discussed the possibilities of differential testing for `non-testable' programs. 
The idea is that if multiple implementations of an algorithm yield different outputs on one identical input, then at least one of the implementation contains a defect.
Alebiosu et al.~\cite{srisakaokul2018multiple} evaluated this idea on machine learning, and successfully found 16 faults from 7 Naive Bayes implementations and 13 faults from 19 $k$-nearest neighbour implementation.

Pham et al.~\cite{Pham2019cradle} also adopted cross referencing to test ML implementations, but focused on the implementation of deep learning libraries.
They proposed CRADLE, the first approach that focuses on finding and localising bugs in deep learning software libraries. The evaluation was conducted on three libraries (TensorFlow, CNTK, and Theano), 11 datasets (including ImageNet, MNIST, and KGS Go game), and 30 pre-trained models. It turned out that CRADLE detects 104 unique inconsistencies and 12 bugs.

DeepXplore~\cite{pei2017deepxplore} and DLFuzz~\cite{guo2018dlfuzz} used differential testing as test oracles to find effective test inputs. Those test inputs causing different behaviours among different algorithms or models were preferred during test generation. 

Most differential testing relies on multiple implementations or versions, while 
Qin et al.~\cite{qin2018syneva} used the behaviours of `mirror' programs, generated from the training data as pseudo oracles.
A mirror program is a program generated based on training data, so that the behaviours of the program represent the training data.
If the mirror program has similar behaviours on test data, it is an indication that the behaviour extracted from the training data suit test data as well.

Sun et al.~\cite{sun2019automatic} applied cross reference in repairing machine translation systems.
Their approach, TransRepair, compares the outputs (i.e., translations) of different mutated inputs, and picks the output that shares the most similarity with others as a superior translation candidate.

\subsubsection{Measurement Metrics for Designing Test Oracles}
\label{sec:modelevaluationmetrics}

Some work has presented definitions or statistical measurements of non-functional features of ML systems including robustness~\cite{tjeng2017evaluating}, fairness~\cite{dwork2012fairness,hardt2016equality,zliobaite2017fairness}, and interpretability~\cite{doshi2017towards,herman2017promise}.
These measurements are not direct oracles for testing, but are essential for testers to understand and evaluate the property under test, and to provide some actual statistics that can be compared with the expected ones.
For example, the definitions of different types of fairness ~\cite{dwork2012fairness,hardt2016equality,zliobaite2017fairness} (more details are in Section~\ref{sec:fairnessdefinitions}) define the conditions an ML system has to satisfy without which the system is not fair. 
These definitions can be adopted directly to detect fairness violations. 



Except for these popular test oracles in \mlt{}, 
there are also some domain-specific rules that could be applied to design test oracles.
We discussed several domain-specific rules that could be adopted as oracles to detect data bugs in Section~\ref{sec:bugdetectionintrainingdata}.
Kang et al.~\cite{kang2018model} discussed two types of model assertions under the task of car detection in videos:
flickering assertion to detect the flickering in car bounding box, and multi-box assertion to detect nested-car bounding. 
For example, if a car bounding box contains other boxes, the multi-box assertion fails.
They also proposed some automatic fix rules to set a new predictive result when a test assertion fails.

There has also been a discussion about the possibility of evaluating ML learning curve in lifelong machine learning as the oracle~\cite{li2015lifelong}.
An ML system can pass the test oracle if it can grow and increase its knowledge level over
time.

\subsection{Test Adequacy}
\label{sec:evaluation}

Test adequacy evaluation aims to discover whether the existing tests have a good fault-revealing ability.
It provides an objective confidence measurement on testing activities.
The adequacy criteria can also be adopted to guide test generation.
Popular test adequacy evaluation techniques in traditional software testing include code coverage and mutation testing, which are also adopted in \mlt{}.

\subsubsection{Test Coverage}
In traditional software testing, code coverage measures the degree to which the source code of a program is executed by a test suite~\cite{zhang2016PMT}. 
The higher coverage a test suite achieves, it is more probable that the hidden bugs could be uncovered. In other words, covering the code fragment is a necessary condition to detect the defects hidden in the code. It is often desirable to create test suites to achieve higher coverage.


Unlike traditional software, code coverage is seldom a demanding criterion for ML testing, since the decision logic of an ML model is not written manually but rather it is learned from training data.
For example, in the study of Pei et al.~\cite{pei2017deepxplore}, 100\,\% traditional code coverage is easy to achieve with a single randomly chosen test input.
Instead, researchers propose various types of coverage for ML models beyond code coverage.

\noindent \textbf{Neuron coverage}.
Pei et al.~\cite{pei2017deepxplore} proposed the first coverage criterion, neuron coverage, particularly designed for deep learning testing.
Neuron coverage is calculated as the ratio of the number of unique neurons activated by all test inputs and the total number of neurons in a DNN. In particular, a neuron is activated if its output value is larger than a user-specified threshold. 

Ma et al.~\cite{Ma2018ase} extended the concept of neuron coverage. They first profile a DNN based on the training data, so that they obtain the activation behaviour of each neuron against the training data. Based on this, they propose more fine-grained criteria, $k$-multisection neuron coverage, neuron boundary coverage, and strong neuron activation coverage, to represent the major functional behaviour and corner behaviour of a DNN.

\noindent \textbf{MC/DC coverage variants}.
Sun et al.~\cite{sun2018testing} proposed four test coverage criteria that are tailored to the distinct features of DNN inspired by the MC/DC coverage criteria~\cite{dupuy2000empirical}.
MC/DC observes the change of a Boolean variable, while their proposed criteria observe a sign, value, or distance change of a neuron, in order to capture the causal changes in the test inputs.
The approach assumes the DNN to be a fully-connected network, and does not consider the context of a neuron in its own layer as well as different neuron combinations within the same layer~\cite{sekhon2019improved}.

\noindent\textbf{Layer-level coverage}.
Ma et al.~\cite{Ma2018ase} also presented layer-level coverage criteria, which considers the top hyperactive neurons and their combinations (or the sequences) to characterise the behaviours of a DNN.
The coverage is evaluated to have better performance together with neuron coverage based on dataset MNIST and ImageNet.
In their following-up work~\cite{ma2018combinatorial,LeideepCT}, they further proposed combinatorial testing coverage, which checks the combinatorial activation status of the neurons in each layer via checking the fraction of neurons activation interaction in a layer.
Sekhon and Fleming~\cite{sekhon2019improved} defined a coverage criteria that looks for 1) all pairs of neurons in the same layer having all possible value combinations, and 2) all pairs of neurons in consecutive layers having all possible value combinations.

\noindent \textbf{State-level coverage}.
While aftermentioned criteria, to some extent, capture the behaviours of feed-forward neural networks, they do not explicitly characterise stateful machine learning system like recurrent neural network (RNN).
The RNN-based ML approach has achieved notable success in applications that handle sequential inputs, e.g., speech audio, natural language, cyber physical control signals.
In order to analyse such stateful ML systems, Du et al.~\cite{du2018deepcruiser} proposed the first set of testing criteria specialised for RNN-based stateful deep learning systems.
They first abstracted a stateful deep learning system as a probabilistic transition system. 
Based on the modelling, they proposed criteria based on the state and traces of the transition system, to capture the dynamic state transition behaviours.

\noindent\textbf{Limitations of Coverage Criteria}.
Although there are different types of coverage criteria, most of them focus on DNNs.
Sekhon and Fleming~\cite{sekhon2019improved} examined the existing testing methods for DNNs and discussed the limitations of these criteria.

Most proposed coverage criteria are based on the structure of a DNN. Li et al.~\cite{li2019structural} pointed out the limitations of structural coverage criteria for deep networks caused by the fundamental differences
between neural networks and human-written programs.
Their initial experiments with natural inputs found no strong correlation between the number of misclassified
inputs in a test set and its structural coverage.
Due to the black-box nature of a machine learning system, it is not clear how such criteria directly relate to the system's decision logic.

\subsubsection{Mutation Testing}
In traditional software testing, mutation testing evaluates the fault-revealing ability of a test suite via injecting faults~\cite{jia2011analysis,zhang2016PMT}. 
The ratio of detected faults against all injected faults is called the \emph{Mutation Score}.

In \mlt{}, the behaviour of an ML system depends on, not only the learning code, but also data and model structure.
Ma et al.~\cite{ma2018deepmutation} proposed DeepMutation, which mutates DNNs at the source level or model level, to make minor perturbation on the decision boundary of a DNN.
Based on this, a mutation score is defined as the ratio of test instances of which results are changed against the total number of instances. 

Shen et al.~\cite{shen2018munn} proposed five mutation operators for DNNs and evaluated properties of mutation on the MINST dataset.
They pointed out that domain-specific mutation operators are needed to enhance mutation analysis.

Compared to structural coverage criteria, mutation testing based criteria are more directly relevant to the decision boundary of a DNN. For example, an input data that is near the decision boundary of a DNN, could more easily detect the inconsistency between a DNN and its mutants. 


\subsubsection{Surprise Adequacy}

Kim et al.~\cite{kim2018guiding} introduced 
\textbf{surprise adequacy} to measure the coverage of discretised input surprise range for deep learning systems.
They argued that \jie{test diversity is more meaningful when being measured with respect to the training data.}
A `good' test input should be `sufficiently but not overly surprising' comparing with the training data.
Two measurements of surprise were introduced: one is based on Keneral Density Estimation~(KDE) to approximate the likelihood of the system having seen a similar input during
training, 
the other is based on the distance between vectors representing
the neuron activation traces of the given input and the training data (e.g., Euclidean distance).
\jie{These criteria can be adopted to detect adversarial examples. Further investigation is required to determine whether such criteria enable the behaviour boundaries of ML models to be approximated in terms of surprise. It will also be interesting for future work to study the relationship between adversarial examples, natural error samples, and surprise-based criteria.}

\subsubsection{Rule-based Checking of Test Adequacy}
To ensure the functionality of an ML system, there may be some `typical' rules that are necessary.
Breck et al.~\cite{breck2017ml} offered 28 test aspects to consider and a scoring system used by Google.
Their focus is to measure how well a given machine learning system is tested. 
The 28 test aspects are classified into four types: 
1) the tests for the ML model itself, 2) the tests for ML infrastructure used to build the model, 
3) the tests for ML data used to build the model,
and 4) the tests that check whether the ML system works correctly over time.
Most of them are some must-to-check rules that could be applied to guide test generation.
For example, the training process should be reproducible; all features should be beneficial; 
there should be no other model that is simpler but better in performance than the current one.
Their research indicates that, although ML testing is complex, there are shared properties to design some basic test cases to test the fundamental functionality of the ML system.

\subsection{Test Prioritisation and Reduction}
\label{sec:testprioritisation}
Test input generation in ML has a very large input space to cover.
On the other hand, we need to label every test instance so as to judge predictive accuracy. 
These two aspects lead to high test generation costs.
Byun et al.~\cite{byun2019input} used DNN metrics like cross entropy, surprisal, and Bayesian uncertainty to prioritise test inputs.
They experimentally showed that these are good indicators of inputs that expose unacceptable behaviours, which are also useful for retraining.

Generating test inputs is also computationally expensive.
Zhang et al.~\cite{zhang2019noise} proposed to reduce costs by identifying those test instances that denote the more effective adversarial examples.
The approach is a test prioritisation technique that ranks the test instances based on their sensitivity to noise, because the instances that are more sensitive to noise is more likely to yield adversarial examples.

Li et. al~\cite{li2019boosting} focused on test data reduction in operational DNN testing.
They proposed a sampling technique guided by the neurons in the last hidden layer of a DNN, using a cross-entropy minimisation based distribution approximation technique.
The evaluation was conducted on pre-trained models with three image datasets: MNIST, Udacity challenge, and ImageNet. 
The results show that, compared to random sampling, their approach samples only half the test inputs, yet it achieves a similar level of precision.

Ma et al.~\cite{ma2019test} proposed a set of test selection metrics based on the notion of model confidence.
Test inputs that are more uncertain to the models are preferred, because they are more informative and should be used to improve the model if being included during retraining.
The evaluation shows that their test selection approach has 80\% more gain than random selection.

\subsection{Bug Report Analysis}
\label{sec:bugreport}

Thung et al.~\cite{thung2012empirical} were the first to study machine learning bugs via analysing the bug reports of machine learning systems.
500 bug reports from Apache Mahout, Apache Lucene, and Apache OpenNLP were studied.
The explored problems included bug frequency, bug categories, bug severity, and bug resolution characteristics such as bug-fix time, effort, and file number.
The results indicated that incorrect implementation counts for the largest proportion of ML bugs, i.e., 22.6\% of bugs are due to incorrect implementation of defined algorithms.
Implementation bugs are also the most severe bugs, and take longer to fix.
In addition,
15.6\% of bugs are non-functional bugs. 
5.6\% of bugs are data bugs.

Zhang et al.~\cite{Zhangyuhaotensor} studied 175 TensorFlow bugs, based on the bug reports from Github or StackOverflow.
They studied the symptoms and root causes of the bugs, the existing challenges to detect the bugs and how these bugs are handled.
They classified TensorFlow bugs into exceptions or crashes, low correctness,
low efficiency, and unknown.
The major causes were found to be in algorithm design and implementations such as TensorFlow API misuse (18.9\%), unaligned tensor (13.7\%), and incorrect model parameter or structure (21.7\%)

Banerjee et al.~\cite{Banerjee2018DSN} analysed the bug reports of autonomous driving systems from 12 autonomous vehicle manufacturers that drove a cumulative total of 1,116,605 miles in California.
They used NLP technologies to classify the causes of disengagements into 10 types (A disagreement is a failure that causes the control of the vehicle to switch from the software
to the human driver).
The issues in machine learning systems and decision control account for the primary cause of 64\,\% of all disengagements based on their report analysis.

\subsection{Debug and Repair}
\label{sec:debug}

\noindent \textbf{Data Resampling}.
The generated test inputs introduced in Section~\ref{sec:input} only expose ML bugs, but are also studied as a part of the training data and can improve the model's correctness through retraining.
For example, DeepXplore achieves up to 3\% improvement in classification accuracy by retraining a deep learning model on generated inputs.
DeepTest~\cite{tian2018deeptest} improves the model's accuracy by 46\%.

Ma et al.~\cite{ma2018mode} identified the neurons responsible for the misclassification and call them `faulty neurons'.
They resampled training data that influence such faulty neurons to help improve model performance.

\noindent\textbf{Debugging Framework Development}.
\jie{
Dutta et al.~\cite{dutta2019storm} proposed Storm, a program transformation framework to generate smaller programs that can support debugging for machine learning testing. 
To fix a bug, developers usually need to shrink the program under test to write better bug reports and to facilitate debugging and regression testing. Storm applies program analysis and probabilistic reasoning to simplify probabilistic programs, which helps to pinpoint the issues more easily.
}

Cai et al.~\cite{cai2016tensorflow} presented \emph{tfdbg},
a debugger for ML models built on TensorFlow, containing three key components:
1) the Analyzer, which makes the structure and intermediate state of the runtime graph visible;
2) the NodeStepper, which enables clients to pause, inspect, or modify at a given node of the graph;
3) the RunStepper, which enables clients to take higher level actions between iterations of model training.
Vartak et al.~\cite{vartak2018mistique} proposed the MISTIQUE system to capture, store, and query model intermediates to help the debug.
Krishnan and Wu~\cite{krishnan2017palm} presented PALM, a tool that explains a complex model with a two-part surrogate model: a meta-model that partitions the training data and a set of sub-models that approximate the patterns within each partition.
PALM helps developers find out the training data that impacts prediction the most, and thereby target the subset of training data that account for the incorrect predictions to assist debugging.

\noindent \textbf{Fix Understanding}.
Fixing bugs in many machine learning systems is difficult because bugs can occur at multiple points in different components.
Nushi et al.~\cite{nushi2017human} proposed a human-in-the-loop approach that simulates potential fixes in different components through human computation tasks: humans were asked to simulate improved component states.
The improvements of the system are recorded and compared, to provide guidance to designers about how they can best improve the system. 

\jienew{
\noindent \textbf{Program Repair}.
Albarghouthi et al.~\cite{albarghouthi2017repairing} proposed a distribution-guided inductive synthesis approach to repair decision-making programs such as machine learning programs. 
The purpose is to construct a new program with correct predictive output, but with similar semantics with the original program.
Their approach uses sampled instances and the predicted outputs to drive program synthesis in which the program is encoded based on SMT. 
}

\subsection{General Testing Framework and Tools}
\label{sec:generalframework}
There has also been work focusing on providing a testing tool or framework that helps developers to implement testing activities in a testing workflow.
There is a test framework to generate and validate test inputs for security testing~\cite{Yang2018TelemadeAT}.
Dreossi et al.~\cite{dreossi2017systematic} presented a CNN testing framework that consists of three main modules: an image generator, a collection of sampling methods, and a suite of visualisation tools. 
Tramer et al.~\cite{tramer2017fairtest} proposed a comprehensive testing tool to help developers to test and debug fairness bugs with an easily interpretable bug report.
Nishi et al.~\cite{nishi2018test} proposed a testing
framework including different evaluation aspects such as allowability, achievability, robustness, avoidability and improvability.
They also discussed different levels of ML testing, such as system, software, component, and data testing.

Thomas et al. ~\cite{thomas2019preventing} recently proposed a framework for designing machine learning algorithms, which simplifies the regulation of undesired behaviours. 
The framework is demonstrated to be suitable for regulating regression, classification, and reinforcement algorithms.
It allows one to learn from (potentially biased) data
while guaranteeing that, with high
probability, the model will exhibit no bias when applied to unseen data.  
The definition of bias is user-specified, allowing for a large family of definitions.  
For a learning algorithm and a training data, the framework will either returns a model with this guarantee, or a warning that it fails to find such a model with the required
guarantee.

\section{ML Properties to be Tested}
\label{sec:testingproperties}

ML properties concern the conditions we should care about for \mlt{}, and are usually connected with the behaviour of an ML model after training.
The poor performance in a property, however, may be due to bugs in any of the ML components (see more in Introduction~\ref{sec:testingcomponents}).

This section presents the related work of testing both functional ML properties and non-functional ML properties.
Functional properties include correctness (Section~\ref{sec:effectiveness}) and overfitting (Section~\ref{sec:overfitting}). 
Non-functional properties include robustness and security (Section~\ref{sec:robustness}), efficiency (Section~\ref{sec:efficiency}), fairness (Section~\ref{sec:fairness}).

\subsection{Correctness}
\label{sec:effectiveness} 
Correctness concerns the fundamental function accuracy of an ML system.
Classic machine learning validation is the most well-established and widely-used technology for correctness testing.
Typical machine learning validation approaches are cross-validation and bootstrap. The principle is to isolate test data via data sampling to check whether the trained model fits new cases.
There are several approaches to perform cross-validation.
In hold out cross-validation~\cite{kohavi1995study}, the data are split into two parts: one part becomes the training data and the other part becomes test data\footnote{Sometimes a validation set is also needed to help train the model, in which circumstances the validation set will be isolated from the training set.}.
In $k$-fold cross-validation, the data are split into $k$ equal-sized subsets: one subset used as the test data and the remaining $k-1$ as the training data. 
The process is then repeated $k$ times, with each of the subsets serving as the test data.
In leave-one-out cross-validation, $k$-fold cross-validation is applied, where $k$ is the total number of data instances.
In Bootstrapping, the data are sampled with replacement~\cite{efron1994introduction}, and thus the test data may contain repeated instances.

There are several widely-adopted correctness measurements such as accuracy, precision, recall, and Area Under Curve (AUC).
There has been work~\cite{japkowicz2006question} analysing the disadvantages of each measurement criterion. 
For example, accuracy does not distinguish between the types of errors it makes (False Positive versus False Negatives). 
Precision and Recall may be misled when data is unbalanced. 
An implication of this work is that we should carefully choose performance metrics.
Chen et al. studied the variability of both training data and test data when assessing the correctness of an ML classifier~\cite{chen2012classifier}.
They derived analytical expressions for the variance of the estimated performance and provided an open-source software implemented with an efficient computation algorithm.
They also studied the performance of different statistical methods when comparing AUC, and found that the $F$-test has the best performance~\cite{chen2013assessment}.




To better capture correctness problems, Qin et al.~\cite{qin2018syneva} proposed to generate a mirror program from the training data, and then use the behaviours this mirror program to serve as a correctness oracle.
The mirror program is expected to have similar behaviours as the test data.


There has been a study of the prevalence of correctness problems among all the reported ML bugs: 
Zhang et al.~\cite{Zhangyuhaotensor} studied 175 Tensorflow bug reports from StackOverflow QA (Question and Answer) pages and from Github projects.
Among the 175 bugs, 40 of them concern poor correctness.

Additionally, there have been many works on 
detecting data bug that may lead to low correctness~\cite{hynes2017data,krishnan2017boostclean,schelter2018automating} (see more in Section~\ref{sec:datatesting}),
test input or test oracle design~\cite{Ding2017,breck2017ml,nakajima2017generalized,al2017validation,pesu2018monte,Dwarakanathissta2018,ma2018mode},
and test tool design~\cite{baylor2017tfx,nushi2017human,vartak2018mistique} (see more in Section~\ref{sec:testingworkflow}).

\subsection{Model Relevance}
\label{sec:overfitting} 

\jienew{Model relevance evaluation detects mismatches between model and data. A poor model relevance is usually associated with overfitting or underfitting.}
When a model is too complex for the data, even the noise of training data is fitted by the model~\cite{hawkins2004problem}. 
Overfitting can easily happen, especially when the training data is insufficient, ~\cite{chan1999classifier,sahiner2000feature,fukunaga1989effects}.

Cross-validation is traditionally considered to be a useful way to detect overfitting.
However, it is not always clear how much overfitting is acceptable and cross-validation may be unlikely to detect overfitting if the test data is unrepresentative of potential unseen data.

Zhang et al.~\cite{zhang:hal-02139208} introduced Perturbed Model Validation (PMV) to help model selection. 
PMV injects noise to the training data, re-trains the model against the perturbed data, then uses the training accuracy decrease rate to detect overfitting/underfitting.
The intuition is that an overfitted learner tends to fit noise in the training sample, while an underfitted learner will have low training accuracy regardless the presence of injected noise.
Thus, both overfitting and underfitting tend to be less insensitive to noise and exhibit a small accuracy decrease rate against noise degree on perturbed data.
PMV was evaluated on four real-world datasets (breast cancer, adult, connect-4, and MNIST) and nine synthetic datasets in the classification setting. 
The results reveal that PMV has better performance and provides a more recognisable signal for detecting both overfitting/underfitting compared to 10-fold cross-validation.


An ML system usually gathers new data after deployment, which will be added into the training data to improve correctness. 
The test data, however, cannot be guaranteed to represent the future data.
Werpachowski et al.~\cite{werpachowski2019detecting} presents an overfitting detection approach via generating adversarial examples from test data.
If the reweighted error estimate on adversarial examples is sufficiently different from that of the original test set, overfitting is detected.
They evaluated their approach on ImageNet and CIFAR-10.

Gossmann et al. ~\cite{gossmann2018test} studied the threat of test data reuse practice in the medical domain with extensive simulation
studies, and found that the repeated reuse of the same test data will inadvertently result in overfitting under all considered simulation settings.

Kirk~\cite{kirk2014thoughtful} mentioned that we could use training time as a complexity proxy for an ML model;
it is better to choose the algorithm with equal correctness but relatively small training time.


Ma et al.~\cite{ma2018mode} tried to relieve the overfitting problem via re-sampling the training data.
Their approach was found to improve test accuracy from 75\% to 93\% on average, based on an evaluation using three image classification datasets.


\subsection{Robustness and Security}
\label{sec:robustness}


\subsubsection{Robustness Measurement Criteria}

Unlike correctness or overfitting, robustness is a non-functional characteristic of a machine learning system.
A natural way to measure robustness is to check the correctness of the system with the existence of noise~\cite{tjeng2017evaluating}; a robust system should maintain performance in the presence of noise.

Moosavi-Dezfooli et al.~\cite{moosavi2016deepfool} proposed DeepFool that computes perturbations (added noise) that `fool' deep networks so as to quantify their robustness.
Bastani et al.~\cite{bastani2016measuring} presented three metrics to measure robustness: 
1) pointwise robustness, indicating the minimum input change a classifier fails to be robust;
2) adversarial frequency, indicating how often changing an input changes a classifier's results;
3) adversarial severity, indicating the distance between an input and its nearest adversarial example.

Carlini and Wagner~\cite{carlini2017towards} created a set of attacks that can be used
to construct an upper bound on the robustness of a neural network.
Tjeng et al.~\cite{tjeng2017evaluating} proposed to use the distance between a test input and its closest adversarial example to measure robustness.
Ruan et al.~\cite{ruan2018global} provided global robustness lower and upper bounds based on the test data to quantify the robustness. 
Gopinath~\cite{gopinath2018deepsafe} et al. proposed DeepSafe, a data-driven approach for assessing DNN robustness: inputs that are clustered into the same group should share the same label.

More recently, Mangal et al.~\cite{Mangal2019robustness} proposed the definition of probabilistic robustness. 
Their work used abstract interpretation to approximate the behaviour of a neural network and to compute an over-approximation of the input regions where the network may exhibit non-robust behaviour.

Banerjee et al.~\cite{Banerjee2019towards} explored the use of Bayesian Deep Learning to model the propagation of errors inside deep neural networks to mathematically model the sensitivity of neural networks to hardware errors, without performing extensive fault injection experiments.

\subsubsection{Perturbation Targeting Test Data}


The existence of adversarial examples allows attacks that may lead to serious consequences in safety-critical applications such as self-driving cars.
There is a whole separate literature on adversarial example generation that deserves a survey of its own, and so this paper does not attempt to fully cover it. Rather, we focus on those promising aspects that could be fruitful areas for future research at the intersection of traditional software testing and machine learning.

Carlini and Wagner~\cite{carlini2017towards}
developed adversarial example generation approaches using distance metrics to quantify similarity.
The approach succeeded in generating adversarial examples for all images on the recently proposed defensively distilled networks~\cite{papernot2016distillation}. 

Adversarial input generation has been widely adopted to test the robustness of autonomous driving systems~\cite{pei2017deepxplore,tian2018deeptest,Ma2018ase,deeproad,wicker2018feature}.
There has also been research on generating adversarial inputs for NLI models~\cite{nie2018analyzing,wang2018if}(Section~\ref{sec:NLI}), malware detection~\cite{Yang2018TelemadeAT}, and Differentiable Neural Computer (DNC)~\cite{chan2018metamorphic}.

Papernot et al.~\cite{papernot2016cleverhans,papernot2018cleverhans} designed a library to standardise the implementation of adversarial example construction.
They pointed out that standardising adversarial example generation is important because `benchmarks constructed without a
standardised implementation of adversarial example construction are not
comparable to each other': it is not easy to tell whether a good result is caused by a high level of robustness or by the differences in the adversarial example construction procedure.




Other techniques to generate test data that check the neural network robustness include symbolic execution~\cite{Sunase2018,gopinath2018symbolic}, fuzz testing~\cite{guo2018dlfuzz}, combinatorial Testing~\cite{ma2018combinatorial},
and abstract interpretation~\cite{Mangal2019robustness}. 
In Section~\ref{sec:input}, we cover these test generation techniques in more detail.

\subsubsection{Perturbation Targeting the Whole System}

Jha et al.~\cite{jha2018avfi} presented AVFI, which used application/software fault injection to approximate hardware errors in the sensors, processors, or memory of the autonomous vehicle (AV) systems to test the robustness.
They also presented Kayotee~\cite{jha2018kayotee}, a fault injection-based tool to systematically inject faults into software and hardware components of the autonomous driving systems. Compared with AVFI, Kayotee is capable of
characterising error propagation and masking
using a closed-loop simulation environment, which is also capable
of injecting bit flips directly into GPU and CPU architectural
state.
DriveFI~\cite{jhaml2019mlbased}, further presented by Jha et al., is a fault-injection engine that mines situations and faults that maximally impact AV safety.

Tuncali et al.~\cite{tuncali2018simulation} considered the closed-loop behaviour of the whole system to support adversarial example generation for autonomous driving systems, not only in image space, but also in configuration space.


\subsection{Efficiency}
\label{sec:efficiency} 

The empirical study of Zhang et al.~\cite{Zhangyuhaotensor} on Tensorflow bug-related artefacts (from StackOverflow QA page and Github) found that nine out of 175 ML bugs (5.1\%) belong to efficiency problems.
The reasons may either be that efficiency problems rarely occur or that these issues are difficult to detect.

Kirk~\cite{kirk2014thoughtful} pointed out that it is possible to use the efficiency of different machine learning algorithms when training the model to compare their complexity.

Spieker and Gotlieb~\cite{spieker2019towards} studied three training data reduction approaches, the goal of which was to find a smaller subset of the original training data set with similar characteristics during model training, so that model building speed could be improved for faster machine learning testing.

\subsection{Fairness}
\label{sec:fairness}

Fairness is a relatively recently-emerging non-functional characteristic. 
According to the work of Barocas and Selbst~\cite{barocas2016big}, there are the following five major causes of unfairness.

\noindent1) \emph{Skewed sample}: once some initial bias happens, such bias may compound over time.

\noindent2) \emph{Tainted examples}: the data labels are biased because of biased labelling activities of humans.

\noindent3) \emph{Limited features}: features may be less informative or reliably collected, misleading the model in building the connection between the features and the labels.

\noindent4) \emph{Sample size disparity}: if the data from the minority group and the majority group are highly imbalanced, ML model may the minority group less well.

\noindent5) \emph{Proxies}: some features are proxies of sensitive attributes (e.g., neighbourhood in which a person resides), and may cause bias to the ML model even if sensitive attributes are excluded.

Research on fairness focuses on measuring, discovering, understanding, and coping with the observed differences regarding different groups or individuals in performance.
Such differences are associated with fairness bugs, which can offend and even harm users, and cause programmers and businesses embarrassment, mistrust, loss of revenue, and even legal violations~\cite{tramer2017fairtest}.

\subsubsection{Fairness Definitions and Measurement Metrics}
\label{sec:fairnessdefinitions}

There are several definitions of fairness proposed in the literature, yet no firm consensus~\cite{gajane2017formalizing,verma2018fairness,saxena2018fairness,afetal:re08}.
Nevertheless, these definitions can be used as oracles to detect fairness violations in \mlt{}.

To help illustrate the formalisation of ML fairness, we use 
$X$ to denote a set of individuals,
$Y$ to denote the true label set when making decisions regarding each individual in $X$.
Let $h$ be the trained machine learning predictive model.
Let $A$ be the set of sensitive attributes, and $Z$ be the remaining attributes.

\noindent \textbf{1) Fairness Through Unawareness}.
Fairness Through Unawareness (FTU) means that an algorithm is fair so long as the protected
attributes are not explicitly used in the decision-making process~\cite{kusner2017counterfactual}.
It is a relatively low-cost way to define and ensure fairness. 
Nevertheless, sometimes the non-sensitive attributes in $X$ may contain
information correlated to sensitive attributes that may thereby lead to discrimination~\cite{gajane2017formalizing,kusner2017counterfactual}. 
Excluding sensitive attributes may also impact model accuracy and yield less effective predictive results~\cite{grgic2016case}.

\noindent \textbf{2) Group Fairness}.
A model under test has group fairness if groups selected based on 
sensitive attributes have an equal probability of decision outcomes.
There are several types of group fairness.

\textbf{\emph{Demographic Parity}} is a popular group fairness measurement~\cite{zafar2015fairness}. 
It is also named \emph{Statistical Parity} or \emph{Independence Parity}.
It requires that a decision should be independent of the protected attributes. 
Let $G_1$ and $G_2$ be the two groups belonging to $X$ divided by a sensitive attribute $a\in A$.
A model $h$ under test satisfies demographic parity if $P\{h(x_i)=1|x_i\in G_1\} = P\{h(x_j)=1|x_j \in G_2\}$.

\textbf{\emph{Equalised Odds}} is another group fairness approach proposed by Hardt et al.~\cite{hardt2016equality}.
A model under test $h$ satisfies \emph{Equalised Odds} if $h$ is independent of the protected attributes when a target label $Y$ is fixed as $y_i$:
$P\{h(x_i)=1|x_i\in G_1,Y=y_i\} = P\{h(x_j)=1|x_j \in G_2,Y=y_i\}$.

When the target label is set to be positive, Equalised Odds becomes \textbf{\emph{Equal Opportunity}}~\cite{hardt2016equality}.
It requires that the true positive rate should be the same for all the groups.
A model $h$ satisfies \emph{Equal Opportunity} if $h$ is independent of the protected attributes when a target class $Y$ is fixed as being positive:
$P\{h(x_i)=1|x_i\in G_1,Y=1\} = P\{h(x_j)=1|x_j \in G_2,Y=1\}$.

\noindent \textbf{3) Counter-factual Fairness}.
Kusner et al.~\cite{kusner2017counterfactual} introduced \textbf{\emph{Counter-factual Fairness}}.
A model satisfies Counter-factual Fairness if its output remains the same when the protected
attribute is flipped to a counter-factual value, and other variables are modified as determined by the assumed causal model.
\jie{Let $a$ be a protected attribute, $a'$ be the counterfactual attribute of $a$, $x_i'$ be the new input with $a$ changed into $a'$.
Model $h$ is counter-factually fair if, for any input $x_i$ and protected attribute $a$:
}
$P\{h(x_i)_{a}=y_i|a\in A,x_i \in X\} = P\{h(x_i')_{a'}=y_i|a \in A,x_i \in X\}$.
This measurement of fairness additionally provides a mechanism to interpret the causes of bias, \jie{because the variables other than the protected attributes are controlled, and thus the differences in $h(x_i)$ and $h(x_i')$ must be caused by variations in $A$.}

\noindent \textbf{4) Individual Fairness}.
Dwork et al.~\cite{dwork2012fairness} proposed a use task-specific similarity metric to describe the pairs of individuals that should be regarded as similar. 
According to Dwork et al., a model $h$ with individual fairness should give similar predictive results among similar individuals: $P\{h(x_i)|x_i \in X\} = P\{h(x_j)=y_i|x_j \in X\}$ iff $d(x_i,x_j)<\epsilon$, where $d$ is a distance metric for individuals that measures their similarity, and $\epsilon$ is tolerance to such differences.


\noindent \textbf{Analysis and Comparison of Fairness Metrics}.
Although there are many existing definitions of fairness, each has its advantages and disadvantages.
Which fairness is the most suitable remains controversial.
There is thus some work surveying and analysing the existing fairness metrics, or investigate and compare their performance based on experimental results, as introduced below.

Gajane and Pechenizkiy~\cite{gajane2017formalizing} surveyed how fairness is defined and formalised in the literature.
Corbett-Davies and Goel~\cite{corbett2018measure} studied three types of fairness definitions: anti-classification, classification parity, and calibration.
They pointed out the deep statistical limitations of each type with examples.
Verma and Rubin~\cite{verma2018fairness} explained and illustrated the existing most prominent fairness definitions based on a common, unifying dataset.

Saxena et al.~\cite{saxena2018fairness} investigated people's perceptions of three of the fairness definitions. 
About 200 recruited participants from Amazon's Mechanical Turk were asked to choose their preference over three allocation rules on two individuals having each applied for a loan. 
The results demonstrate a clear preference for the way of allocating resources in proportion to the applicants' loan repayment rates.

\noindent \textbf{\jienew{Support for Fairness Improvement.}}
\jie{
Metevier et al.~\cite{Metevier19neurips} proposed RobinHood, an algorithm that supports multiple user-defined fairness definitions under the scenario of offline contextual bandits\footnote{\jienew{A contextual bandit is a type of algorithm that learns to take actions based on rewards such as user click rate~\cite{li2010contextual}.}}.
RobinHood makes use of concentration inequalities~\cite{massart2007concentration} to calculate high-probability bounds and to search for solutions that satisfy the fairness requirements. It gives user warnings when the requirements are violated. 
The approach is evaluated under three application scenarios: a tutoring system, a loan approval setting, and the criminal recidivism, all of which demonstrate the superiority of RobinHood over other algorithms.
}

\jie{Albarghouthi and Vinitsky~\cite{albarghouthi2019fairness} proposed the concept of `fairness-aware programming', in which fairness is a first-class concern. 
To help developers define their own fairness specifications, they developed a specification language.
Like assertions in traditional testing, the fairness specifications are developed into the run-time monitoring code to enable multiple executions to catch violations.
A prototype was implemented in Python.
}

\jienew{
Agarwal et al.~\cite{agarwal2018reductions} proposed to reduce fairness classification into a problem of cost-sensitive classification (where the costs of different types of errors are differentiated).
The application scenario is binary classification, with the underlying classification method being treated as a black box.
The reductions optimise the trade-off between accuracy and fairness constraints. 
}

\jienew{Albarghouthi et al.~\cite{albarghouthi2017repairing} proposed an approach to repair decision-making programs using distribution-guided inductive synthesis.
}


\subsubsection{Test Generation Techniques for Fairness Testing}

Galhotra et al.~\cite{galhotra2017fairness,angell2018themis} proposed Themis which considers group fairness using causal analysis~\cite{Brittany2018causaltesting}.
It defines fairness scores as measurement criteria for fairness and uses random test generation techniques to evaluate the degree of discrimination (based on fairness scores).
Themis was also reported to be more efficient on systems that exhibit more discrimination.

Themis generates tests randomly for group fairness, while Udeshi et al.~\cite{udeshi2018automated} proposed Aequitas, focusing on test generation to uncover discriminatory inputs and those inputs essential to understand individual fairness.
The generation approach first randomly samples the input space to discover the presence of discriminatory inputs, then searches the neighbourhood of these inputs to find more inputs.
As well as detecting fairness bugs, \emph{Aeqitas} also retrains the machine-learning models and reduce discrimination in the decisions made by these models.

Agarwal et al.~\cite{agarwal2018automated} used symbolic execution together with local explainability to generate test inputs.
The key idea is to use the local explanation, specifically Local Interpretable
Model-agnostic Explanations\footnote{Local Interpretable
Model-agnostic Explanations produces decision trees corresponding to an input that could provide paths in symbolic execution~\cite{ribeiro2016should}} to identify whether factors that drive decisions include protected attributes.
The evaluation indicates that the approach generates 3.72 times more successful test cases than THEMIS across 12 benchmarks.

Tramer et al.~\cite{tramer2017fairtest} were the first to proposed the concept of `fairness bugs'.
They consider a statistically significant association between a protected attribute and an algorithmic output to be a fairness bug, specially named `Unwarranted Associations' in their paper.
They proposed the first comprehensive testing tool, aiming to help developers test and debug fairness bugs with an `easily interpretable' bug report.
The tool is available for various application areas including image classification, income prediction, and health care prediction.

Sharma and Wehrheim~\cite{arnab2019balancedata} sought to identify causes of unfairness  via checking whether the algorithm under test is sensitive to training data changes.
They mutated the training data in various ways to generate new datasets, such as changing the order of rows, columns, and shuffling feature names and values.
12 out of 14 classifiers were found to be sensitive to these changes.




\subsection{Interpretability}
\label{sec:inter}



\noindent\textbf{Manual Assessment of Interpretability}.
The existing work on empirically assessing the interpretability property usually includes humans in the loop. 
That is, manual assessment is currently the primary approach to evaluate interpretability.
Doshi-Velez and Kim~\cite{doshi2017towards} gave a taxonomy of evaluation (testing) approaches for interpretability: application-grounded, human-grounded, and functionally-grounded.
Application-grounded evaluation involves human experimentation with a real application scenario.
Human-grounded evaluation uses results from human evaluation on simplified tasks.
Functionally-grounded evaluation requires no human experiments but uses a quantitative metric as a proxy for explanation quality, for example, a proxy for the explanation of a decision tree model may be the depth of the tree. 

Friedler et al.~\cite{Friedler2019interpretability} introduced two types of interpretability:
global interpretability means understanding the entirety of a trained model;
local interpretability means understanding the results of a trained model on a specific input and the corresponding output.
They asked 1000 users to produce the expected
output changes of a model given an input change, and then recorded accuracy and completion time over
varied models.
Decision trees and logistic
regression models were found to be more locally interpretable than neural networks.

\noindent\textbf{Automatic Assessment of Interpretability}.
Cheng et al.~\cite{cheng2018towards} presented a metric to understand the behaviours of an ML model.
The metric measures whether the learner has learned the object in object identification scenario via occluding the surroundings of the objects.

Christoph~\cite{molnar} proposed to measure interpretability based on the category of ML algorithms.
He claimed that `the easiest way to achieve interpretability is to use only a subset of algorithms that create interpretable models'. 
He identified several models with good interpretability, including
linear regression, logistic regression and decision tree models.

Zhou et al.~\cite{zhou2018metamorphicrelation} defined the concepts of Metamorphic Relation Patterns (MRPs) and Metamorphic Relation Input Patterns (MRIPs) that can be adopted to help end users understand how an ML system works.
They conducted case studies of various systems, including large commercial websites, Google Maps navigation, Google Maps location-based search, image analysis for face recognition (including Facebook, MATLAB, and OpenCV), and the Google video analysis service Cloud Video Intelligence. 

\noindent\textbf{Evaluation of Interpretability Improvement Methods}.
Machine learning classifiers are widely used in many medical applications, yet the clinical meaning of the predictive outcome is often unclear.
Chen et al.~\cite{chen2018calibration} investigated several interpretability-improving methods which transform classifier scores to a probability of disease scale. 
They showed that classifier scores on arbitrary scales can be calibrated to the probability scale without affecting their discrimination performance.


\subsection{Privacy} 
\label{sec:privacy}

\jienew{ 
Ding et al.~\cite{ding2018detecting} treat programs as grey boxes, and detect differential privacy violations via statistical tests. 
For the detected violations, they generate counter examples to illustrate these violations as well as to help developers understand and fix bugs.
Bichsel et al. ~\cite{bichsel2018dp} proposed to estimate the $\epsilon$ parameter in differential privacy, aiming to find a triple $(x,x',\Phi)$ that witnesses the largest possible privacy violation, where $x$ and $x'$ are two test inputs and $\Phi$ is a possible set of outputs.}

\section{ML Testing Components}
\label{sec:testingcomponents}

This section organises the work on \mlt{} by identifying the component (data, learning program, or framework) for which \mlt{} may reveal a bug. 

\subsection{Bug Detection in Data}
\label{sec:datatesting}
Data is a `component' to be tested in \mlt{},
since the performance of the ML system largely depends on the data. 
Furthermore, as pointed out by Breck et al.~\cite{breck2019datavalidation}, 
it is important to detect data bugs early because
predictions from the trained model are often logged and used to generate further data.
This subsequent generation creates a feedback loop that may amplify even small data bugs over time.

Nevertheless, data testing is challenging~\cite{polyzotis2017data}.
According to the study of Amershi et al.~\cite{Amershi2019se4ai}, the management and evaluation of data is among the most challenging tasks when developing an AI application in Microsoft. 
Breck et al.~\cite{breck2019datavalidation} mentioned that data generation logic often lacks visibility in the ML pipeline; the data are often stored in a raw-value format (e.g., CSV) that strips out semantic information that can help identify bugs.




\subsubsection{Bug Detection in Training Data}
\label{sec:bugdetectionintrainingdata}

\noindent \textbf{Rule-based Data Bug Detection}. Hynes et al.~\cite{hynes2017data} proposed \emph{data linter}-- an ML tool, inspired by code linters, to automatically inspect ML datasets.
They considered three types of data problems:
1) miscoded data, such as mistyping a number or date as a string;
2) outliers and scaling, such as uncommon list length;
3) packaging errors, such as duplicate values, empty examples, and other data organisation issues.

Cheng et al.~\cite{cheng2018towards} presented a series of metrics to evaluate whether the training data have covered all important scenarios. 

\noindent \textbf{Performance-based Data Bug Detection}.
To solve the problems in training data, Ma et al.~\cite{ma2018mode} proposed MODE.
MODE identifies the `faulty neurons' in neural networks that are responsible for the classification errors, and tests the training data via data resampling to analyse whether the faulty neurons are influenced.
MODE allows to improve test effectiveness from 75\,\% to 93\,\% on average, based on evaluation using the MNIST, Fashion MNIST, and CIFAR-10 datasets.

\subsubsection{Bug Detection in Test Data}

Metzen et al.~\cite{metzen2017detecting} proposed to augment
DNNs with a small sub-network, specially designed to distinguish genuine data from data containing adversarial perturbations.
Wang et al.~\cite{wang2018detecting} used DNN model mutation to expose adversarial examples motivated by their observation that adversarial samples are more sensitive to perturbations~\cite{jingyiwang2018detecting}.
The evaluation was based on the MNIST and CIFAR10 datasets.
The approach detects 96.4\,\%/90.6\,\% adversarial samples with 74.1/86.1 mutations for MNIST/CIFAR10.

Adversarial examples in test data raise security risks. Detecting adversarial examples is thereby similar to bug detection.
Carlini and Wagner~\cite{Carlini2017} surveyed ten proposals that are designed for detecting adversarial examples and compared their efficacy.
They found that detection approaches rely on loss functions and can thus be bypassed when constructing new loss functions. 
They concluded that adversarial examples are significantly harder to detect than previously appreciated.

The insufficiency of the test data may not able to detect overfitting issues, and
could also be regarded as a type of data bugs.
The approaches for detecting test data insufficiency were discussed in Section~\ref{sec:evaluation}, such as coverage~\cite{pei2017deepxplore,tian2018deeptest,Ma2018ase} and mutation score~\cite{ma2018deepmutation}. 



\subsubsection{Skew Detection in Training and Test Data}

The training instances and the instances that the model predicts should exhibit consistent features and distributions.
Kim et al.~\cite{kim2018guiding} proposed two measurements to evaluate the skew between training and test data:
one is based on Kernel Density Estimation~(KDE) to approximate the likelihood of the system having seen a similar input during
training, 
the other is based on the distance between vectors representing
the neuron activation traces of the given input and the training
data (e.g., Euclidean distance).

Breck~\cite{breck2019datavalidation} investigated the skew in training data and serving data (the data that the ML model predicts after deployment).
To detect the skew in features, they do key-join feature comparison.
To quantify the skew in distribution,
they argued that general approaches such as KL divergence or cosine similarity might not be sufficiently intuitive for produce teams. 
Instead, they proposed to use the largest change in probability as a value in the two distributions as a measurement of their distance.

\subsubsection{Frameworks in Detecting Data Bugs}

Breck et al.~\cite{breck2019datavalidation} proposed a data validation system for detecting data bugs. 
The system applies constraints (e.g., type, domain, valency) to find bugs in single-batch (within the training data or new data),
and quantifies the distance between training data and new data.
Their system is deployed as an integral part of TFX (an end-to-end machine learning platform at Google).
The deployment in production provides evidence that the system helps early detection and debugging of data bugs.
They also summarised the type of data bugs, in which new feature column, unexpected string values, and missing feature columns are the three most common.

Krishnan et al.~\cite{krishnan2016activeclean, krishnan2016activeclean2} proposed a model training framework, ActiveClean, that allows for iterative data cleaning while preserving provable convergence properties.
ActiveClean suggests a sample of data to clean based on the data's value to the model and the likelihood that it is `dirty'. 
The analyst can then apply value transformations and filtering operations to the sample to `clean' the identified dirty data.

In 2017, Krishnan et al.~\cite{krishnan2017boostclean} presented a system named BoostClean to detect domain value violations (i.e., when an attribute value is outside of an allowed domain) in training data.
The tool utilises the available cleaning resources such as Isolation Forest~\cite{liu2008isolation} to improve a model's performance.
After resolving the problems detected, the tool is able to improve prediction accuracy by up to 9\% in comparison to the best non-ensemble alternative.

ActiveClean and BoostClean may involve a human in the loop of testing process.
Schelter et al.~\cite{schelter2018automating} focus on the automatic `unit' testing of large-scale datasets.
Their system provides a declarative API that combines common as well as user-defined quality constraints for data testing.
Krishnan and Wu~\cite{krishnan2019alphaclean} also targeted automatic data cleaning and proposed AlphaClean.
They used a greedy tree search algorithm to automatically tune the parameters for data cleaning pipelines.
With AlphaClean, the user could focus on defining cleaning requirements and let the system find the best configuration under the defined requirement.
The evaluation was conducted on three datasets, demonstrating that AlphaClean finds solutions of up to 9X better than state-of-the-art parameter tuning methods.

Training data testing is also regarded as a part of a whole machine learning workflow in the work of Baylor et al.~\cite{baylor2017tfx}.
They developed a machine learning platform that enables data testing, based on a property description schema that captures properties such as the features present in the data and the expected type or valency of each feature.

There are also data cleaning technologies such as statistical or learning approaches from the domain of traditional database and data warehousing.
These approaches are not specially designed or evaluated for ML, but they can be re-purposed for \mlt{}~\cite{rahm2000data}.

\subsection{Bug Detection in Learning Program}
\label{sec:trainingprogramtesting}

Bug detection in the learning program checks whether the algorithm is correctly implemented and configured, e.g., the model architecture is designed well, and whether there exist coding errors.

\noindent \textbf{Unit Tests for ML Learning Program}.
McClure~\cite{mcclure2017tensorflow} introduced ML unit testing with TensorFlow built-in testing functions to help ensure that `code will function as expected' to help build developers' confidence.

Schaul et al.~\cite{schaul2014unit} developed a collection of unit tests specially designed for stochastic optimisation. 
The tests are small-scale, isolated with well-understood difficulty. 
They could be adopted in the beginning learning stage to test the learning algorithms to detect bugs as early as possible.

\noindent \textbf{Algorithm Configuration Examination}.
Sun et al.~\cite{sunempirical2017} and Guo et al.~\cite{Guo2018study} identified operating systems, language, and hardware Compatibility issues.
Sun et al.~\cite{sunempirical2017} studied 329 real bugs from three machine learning frameworks: 
Scikit-learn, Paddle, and Caffe. 
Over 22\% bugs are found to be compatibility problems due to incompatible operating systems, language versions, or conflicts with hardware.
Guo et al.~\cite{Guo2018study} investigated deep learning frameworks such as TensorFlow, Theano, and Torch.
They compared the learning accuracy, model size, robustness with different models classifying dataset MNIST and CIFAR-10.

The study of Zhang et al.~\cite{Zhangyuhaotensor} indicates that the most common learning program bug is due to the change of TensorFlow API when the implementation has not been updated accordingly.
Additionally, 23.9\% (38 in 159) of the bugs from ML projects in their study built based on TensorFlow arise from problems in the learning program.

Karpov et al.~\cite{Karpov2018} also highlighted testing algorithm parameters in all neural network testing problems. 
The parameters include the number of neurons and their types based on the neuron layer types, the ways the neurons interact with each other, the synapse weights, and the activation functions.
However, the work currently remains unevaluated.

\noindent \textbf{Algorithm Selection Examination}.
Developers usually have more than one learning algorithm to choose from.
Fu and Menzies~\cite{Fu2017fse} compared deep learning and classic learning on the task of linking Stack Overflow questions, and found that classic learning algorithms (such as refined SVM) could achieve similar (and sometimes better) results at a lower cost. 
Similarly, the work of Liu et al.~\cite{Liu2018} found that the $k$-Nearest Neighbours (KNN) algorithm achieves similar results to deep learning for the task of commit message generation.

\noindent \textbf{Mutant Simulations of Learning Program Faults}.
Murphy et al.~\cite{murphy2009UJR,Murphy2009AST} used mutants to simulate programming code errors to investigate whether the proposed metamorphic relations are effective at detecting errors.
They introduced three types of mutation operators: switching comparison operators, mathematical operators, and off-by-one errors for loop variables.

Dolby et al.~\cite{Dolby2018Ariadne} extended WALA to support static analysis of the behaviour of tensors in Tensorflow learning programs written in Python.
They defined and tracked tensor types for machine learning, and changed WALA to produce a dataflow graph to abstract possible program behavours.

\subsection{Bug Detection in Framework}
\label{sec:frameworktesting}

The current research on framework testing focuses on studying framework bugs (Section~\ref{sec:studyframeworkbugs}) and detecting bugs in framework implementation (Section~\ref{sec:implementationtesting}).

\subsubsection{Study of Framework Bugs}
\label{sec:studyframeworkbugs}


Xiao et al.~\cite{xiao2018security} focused on the security vulnerabilities of popular deep learning frameworks including Caffe, TensorFlow, and Torch.
They examined the code of popular deep learning frameworks.
The dependency of these frameworks was found to be very complex.
Multiple vulnerabilities were identified in their implementations. 
The most common vulnerabilities are bugs that cause programs to crash, non-terminate, or exhaust memory.

Guo et al.~\cite{Guo2018study} tested deep learning frameworks, including TensorFlow, Theano, and Torch, by comparing their runtime behaviour, training accuracy, and robustness, under identical algorithm design and configuration. 
The results indicate that runtime training behaviours are different for each framework, while the prediction accuracies remain similar.

Low Efficiency is a problem for ML frameworks, which may directly lead to inefficiency of the models built on them. 
Sun et al.~\cite{sunempirical2017} found that approximately 10\% of reported framework bugs concern low efficiency.
These bugs are usually reported by users.
Compared with other types of bugs, they may take longer for developers to resolve.


\subsubsection{Implementation Testing of Frameworks}
\label{sec:implementationtesting}
  
Many learning algorithms are implemented inside ML frameworks.
Implementation bugs in ML frameworks may cause neither crashes, errors, nor efficiency problems~\cite{chase2017unitest}, making their detection challenging.

\noindent\textbf{Challenges in Implementation Bug Detection}.
Thung et al.~\cite{thung2012empirical} studied machine learning bugs in 2012.
Their results, regarding 500 bug reports from three machine learning systems, indicated that approximately 22.6\% bugs are due to incorrect algorithm implementations.
Cheng et al.~\cite{cheng2018manifesting} injected implementation bugs into classic machine learning code in Weka and observed the performance changes that resulted.
They found that 
8\% to 40\% of the logically non-equivalent executable mutants (injected implementation bugs) were statistically indistinguishable from their original versions. 


\noindent\textbf{Solutions for Implementation Bug Detection}.
Some work has used multiple implementations or differential testing to detect bugs.
For example, Alebiosu et al.~\cite{srisakaokul2018multiple} found five faults in 10 Naive Bayes implementations and four faults in 20 $k$-nearest neighbour implementations.
Pham et al.~\cite{Pham2019cradle} found 12 bugs in three libraries (i.e., TensorFlow, CNTK, and Theano), 11 datasets (including ImageNet, MNIST, and KGS Go game), and 30 pre-trained models (see Section~\ref{sec:oracle_differential}).

However, not every algorithm has multiple implementations. Murphy et al.~\cite{murphy2007approach, Murphy2008PropertiesOM} were the first to discuss the possibilities of applying metamorphic relations to machine learning implementations.
They listed several transformations of the input data that should ought not to bring changes in outputs, such as multiplying numerical values by a constant, permuting or reversing the order of the input data, and adding additional data.
Their case studies found that their metamorphic relations held on three machine learning applications.

Xie et al.~\cite{xie2009application} focused on supervised learning. They proposed to use more specific metamorphic relations to test the implementations of supervised classifiers.
They discussed five types of potential metamorphic relations on KNN and Naive Bayes on randomly generated data.
In 2011, they further evaluated their approach using mutated machine learning code~\cite{Xie2011JSS}.
Among the 43 injected faults in Weka~\cite{hall2009weka} (injected by MuJava~\cite{ma2005mujava}), the metamorphic relations were able to reveal 39.
In their work, the test inputs were randomly generated data. 

Dwarakanath et al.~\cite{Dwarakanathissta2018} applied metamorphic relations to find implementation bugs in image classification.
For classic machine learning such as SVM, they conducted mutations such as changing feature or instance orders, and linear scaling of the test features. 
For deep learning models such as residual networks (which the data features are not directly available),
they proposed to normalise or scale the test data, or to change the convolution operation order of the data. 
These changes were intended to bring no change to the model performance when there are no implementation bugs. 
Otherwise, implementation bugs are exposed. 
To evaluate, they used MutPy to inject mutants that simulate implementation bugs, of which the proposed metamorphic relations are able to find 71\%.

\subsubsection{Study of Frameworks Test Oracles}
Nejadgholi and Yang~\cite{nejadgholistudy} studied the approximated oracles of four popular deep learning libraries: Tensorflow, Theano, PyTorch, and Keras. 
5\% to 24\% oracles were found to be approximated oracles with a flexible threshold (in contrast to certain oracles).
5\%-27\% of the approximated oracles used the outputs from other libraries/frameworks.
Developers were also found to modify approximated oracles frequently due to code evolution.

\section{Application Scenarios}
\label{sec:applicationsenario}

Machine learning has been widely adopted in different areas.
This section introduces such domain-specific testing approaches in three typical application domains:
autonomous driving, machine translation, and neural language inference.

\subsection{Autonomous Driving}
\label{sec:autodriving}

Testing autonomous vehicles has a comparatively long history. For example, in 2004, Wegener and B\"uhler compared different fitness functions when evaluating the tests of autonomous car parking systems~\cite{wegener2004evaluation}.
Testing autonomous vehicles also has many research opportunities and open questions, as pointed out and discussed by Woehrle et al.~\cite{woehrle2019open}.

\jienew{
More recently, search-based test generation for AV testing has been successfully applied.
Abdessalem et al.~\cite{abdessalem2018testing2,ben2016testing} focused on improving the efficiency and accuracy of search-based testing of advanced driver assistance systems (ADAS) in AVs.
Their algorithms use classification models to improve the efficiency of the search-based test generation for critical scenarios. Search algorithms are further used to refine classification models to improve their accuracy.
Abdessalem et al.~\cite{abdessalem2018testing} also proposed FITEST, a multi-objective search algorithm that searches feature interactions which violate system requirements or lead to failures.
}

Most of the current autonomous vehicle systems that have been put into the market are semi-autonomous vehicles, which require a human driver to serve as a fall-back~\cite{Banerjee2018DSN}, as was the case with the work of Wegener and B\"uhler~\cite{wegener2004evaluation}.
An issue that causes the human driver to take control of the vehicle is called a \emph{disengagement}.

Banerjee et al.~\cite{Banerjee2018DSN} investigated the causes and impacts of 5,328 disengagements from the data of 12 AV manufacturers for 144 vehicles that drove a cumulative 1,116,605 autonomous miles, 42 (0.8\%) of which led to accidents.
They classified the causes of disengagements into 10 types.
64\% of the disengagements were found to be caused by the bugs in the machine learning system, among which the behaviours of image classification (e.g., improper detection of traffic lights, lane markings, holes, and bumps) were the dominant causes accounting for 44\% of all reported disengagements.
The remaining 20\% were due to the bugs in the control and decision framework such as improper motion planning.

Pei et al.~\cite{pei2017deepxplore} used gradient-based differential testing to generate test inputs to detect potential DNN bugs and leveraged neuron coverage as a guideline.
Tian et al.~\cite{tian2018deeptest} proposed to use a set of image transformation to generate tests, which simulate the potential noise that could be present in images obtained from a real-world camera.
Zhang et al.~\cite{deeproad} proposed DeepRoad, a GAN-based approach to generate test images for real-world driving scenes. Their approach is able to support two weather conditions (i.e., snowy and rainy).
The images were generated with the pictures from YouTube videos.
Zhou et al.~\cite{zhou2018deepbillboard} proposed DeepBillboard, which generates real-world adversarial billboards that can trigger potential steering errors of autonomous driving systems. 
It demonstrates the possibility of generating continuous and realistic physical-world tests for practical autonomous-driving systems.

Wicker et al.~\cite{wicker2018feature} used feature-guided Monte Carlo Tree Search to identify elements of an image that are most vulnerable to a self-driving system; adversarial examples.
Jha et al.~\cite{jhaml2019mlbased} accelerated the process of finding `safety-critical' issues via analytically modelling the injection of faults into an AV system as a Bayesian network.
The approach trains the network to identify safety critical faults automatically. 
The evaluation was based on two production-grade AV systems from NVIDIA and Baidu, indicating that the approach can find many situations where faults lead to safety violations. 

Uesato et al.~\cite{uesato2018rigorous} aimed to find catastrophic failures in safety-critical agents like autonomous driving in reinforcement learning. 
They demonstrated the limitations of traditional random testing, then proposed a predictive adversarial example generation approach to predict failures and estimate reliable risks.
The evaluation on TORCS simulator indicates that the proposed approach is both effective and efficient with fewer Monte Carlo runs.

To test whether an algorithm can lead to a problematic model, Dreossi et al.~\cite{dreossi2017systematic} proposed to generate training data as well as test data.
Focusing on Convolutional Neural Networks (CNN), they build a tool to generate natural images and visualise the gathered information to detect blind spots or corner cases under the autonomous driving scenario. 
Although there is currently no evaluation, the tool has been made available\footnote{\url{https://github.com/shromonag/FalsifyNN}}.

Tuncali et al.~\cite{tuncali2018simulation} presented a framework that supports both system-level testing and the testing of those properties of an ML component.
The framework also supports fuzz test input generation and search-based testing using approaches such as Simulated Annealing and Cross-Entropy optimisation.

While many other studies investigated DNN model testing for research purposes,
Zhou et al.~\cite{zhou2019metamorphic} 
combined fuzzing and metamorphic testing to test 
LiDAR, which is an obstacle-perception module of real-life self-driving cars, and detected real-life fatal bugs.

Jha et al. presented AVFI~\cite{jha2018avfi} and Kayotee~\cite{jha2018kayotee}, which are fault injection-based tools to systematically inject faults into autonomous driving systems to assess their safety and reliability.

\jienew{
O'Kelly et al.~\cite{o2018scalable} proposed a `risk-based framework' for AV testing to predict the probability of an accident in a base distribution of traffic behaviour (derived from the public traffic data collected by the US Department of Transportation).
They argued that formally verifying correctness of an AV system is infeasible due to the challenge of formally defining ``correctness'' as well as the white-box requirement. 
Traditional testing AVs in a real environment requires prohibitive amounts of time.
To tackle these problems, they view AV testing as  rare-event simulation problem, then evaluate the accident probability to accelerate AV testing.
}

\subsection{Machine Translation}
\label{sec:machinetranslation}
Machine translation automatically translates text or speech from one language to another.
The BLEU (BiLingual Evaluation Understudy) score~\cite{papineni2002bleu} is a widely-adopted measurement criterion to evaluate machine translation quality.
It assesses the correspondence between a machine's output and that of a human.

Zhou et al.~\cite{Zhou2018metamorphic,pesu2018monte} used self-defined metamorphic relations in their tool `MT4MT' to test the translation consistency of machine translation systems.
The idea is that some changes to the input should not affect the overall structure of the translated output.
Their evaluation showed that Google Translate outperformed Microsoft Translator for long sentences whereas the latter outperformed the former for short and simple sentences. 
They hence suggested that the quality assessment of machine translations should consider multiple dimensions and multiple types of inputs.

Sun et al.~\cite{sun2019automatic} combine mutation testing and metamorphic testing to test and repair the consistency of machine translation systems.
Their approach, TransRepair, enables automatic test input generation, automatic test oracle generation, as well as automatic translation repair. 
They first applied mutation on sentence inputs to find translation inconsistency bugs, then used translations of the mutated sentences to optimise the translation results in a black-box or grey-box manner. 
Evaluation demonstrates that TransRepair fixes 28\% and 19\% bugs on average for Google Translate and Transformer.

Compared with existing model retraining approaches, TransRepair has the following advantages: 
1) more effective than data augmentation; 2) source code in dependant (black box); 
3) computationally cheap (avoids space and time expense of data collection and model retraining); 
4) flexible (enables repair without touching other well-formed translations).

The work of Zheng et al.~\cite{Zheng2018TestingUN,zheng2019TestingUNposter,wang2019TestingUNposter} proposed two algorithms for
detecting two specific types of machine translation violations:
(1) under-translation, where some words/phrases
from the original text are missing in the translation, and (2)
over-translation, where some words/phrases from the original
text are unnecessarily translated multiple times.
The algorithms are based on a statistical analysis of both the original texts and the translations, to check whether there are violations of one-to-one mappings in words/phrases.


\subsection{Natural Language Inference}
\label{sec:NLI}

A Nature Language Inference (NLI) task judges the inference relationship of a pair of natural language sentences.
For example, the sentence `A person is in the room' could be inferred from the sentence `A girl is in the room'.

Several works have tested the robustness of NLI models.
Nie et al.~\cite{nie2018analyzing} generated sentence mutants (called `rule-based adversaries' in the paper) to test whether the existing NLI models have semantic understanding.
Seven state-of-the-art NLI models (with diverse architectures) were all unable to recognise simple semantic differences when the word-level information remains unchanged.

Similarly, Wang et al.~\cite{wang2018if} mutated the inference target pair by simply swapping them.
The heuristic is that 
a good NLI model should report comparable accuracy between the original and swapped test set for contradictory pairs and for neutral pairs,
but lower accuracy in swapped test set for entailment pairs (the hypothesis may or may not be true given a premise).


\section{Analysis of Literature Review}
\label{sec:analysissummary}

This section analyses the research distribution among different testing properties and machine learning categories.
It also summarises the datasets (name, description, size, and usage scenario of each dataset) that have been used in \mlt{}.


\subsection{Timeline}
\label{sec:timeline}

\begin{figure*}[h!]
  \centering
\includegraphics[scale=0.85]{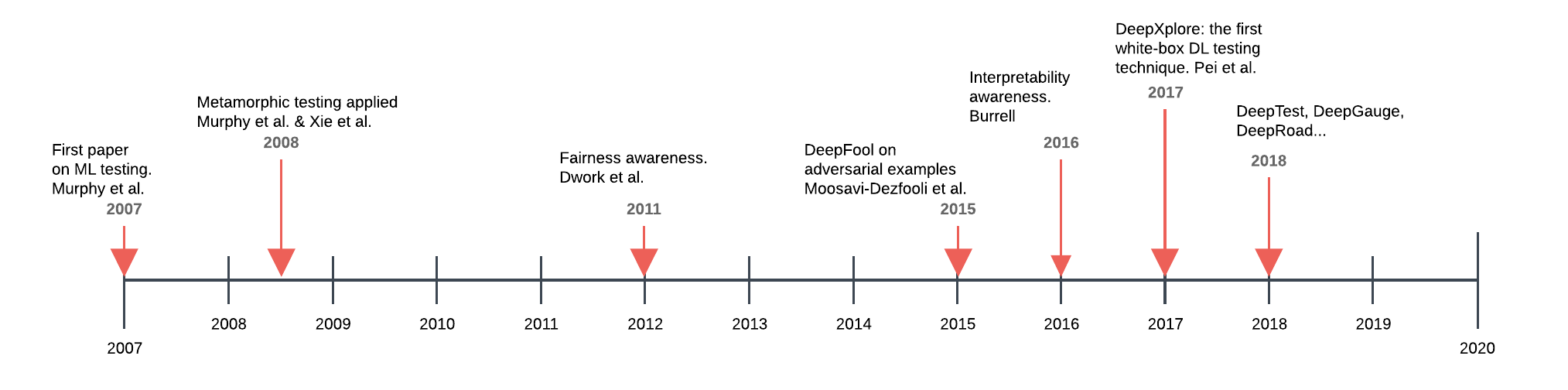}
\caption{Timeline of \mlt{} research}
  \label{fig:timeline}
\end{figure*}

Figure~\ref{fig:timeline} shows several key contributions in the development of \mlt{}.
As early as in 2007, Murphy et al.~\cite{murphy2007approach} mentioned the idea of testing machine learning applications.
They classified machine learning applications as `non-testable' programs considering the difficulty of getting test oracles. 
They primarily consider the detection of implementation bugs, described as ``to ensure that an application using the algorithm correctly implements the specification and fulfils the users' expectations''.
Afterwards, Murphy et al.~\cite{Murphy2008PropertiesOM} discussed the properties of machine learning algorithms that may be adopted as metamorphic relations to detect implementation bugs.

In 2009, Xie et al.~\cite{xie2009application} also applied metamorphic testing on supervised learning applications.

Fairness testing was proposed in 2012 by Dwork et al.~\cite{dwork2012fairness};
the problem of interpretability was proposed in 2016 by Burrell~\cite{burrell2016machine}.

In 2017, Pei et al.~\cite{pei2017deepxplore} published the first white-box testing paper on deep learning systems.
Their work pioneered to propose coverage criteria for DNN.
Enlightened by this paper, a number of machine learning testing techniques have emerged, such as DeepTest~\cite{tian2018deeptest}, DeepGauge~\cite{Ma2018ase}, DeepConcolic~\cite{Sunase2018}, and DeepRoad~\cite{deeproad}.
A number of software testing techniques has been applied to \mlt{},
such as different testing coverage criteria~\cite{tian2018deeptest,sun2018testing,Ma2018ase}, mutation testing~\cite{ma2018deepmutation}, combinatorial testing~\cite{LeideepCT}, metamorphic testing~\cite{chan2018metamorphic}, and fuzz testing~\cite{odena2018tensorfuzz}.

\subsection{Research Distribution among Machine Learning Categories}
\label{sec:type}
This section introduces and compares the research status of each machine learning category. 

\subsubsection{Research Distribution between General Machine Learning and Deep Learning}


To explore research trends in \mlt{}, we classify the collected papers into two categories: those targeting only deep learning and those designed for general machine learning (including deep learning).

Among all \papernum papers, 56 papers (38.9\%) present testing techniques that are specially designed for deep learning alone;
the remaining 88 papers cater for general machine learning. 

\begin{figure}[h!]
\centering
\includegraphics[scale=0.41]{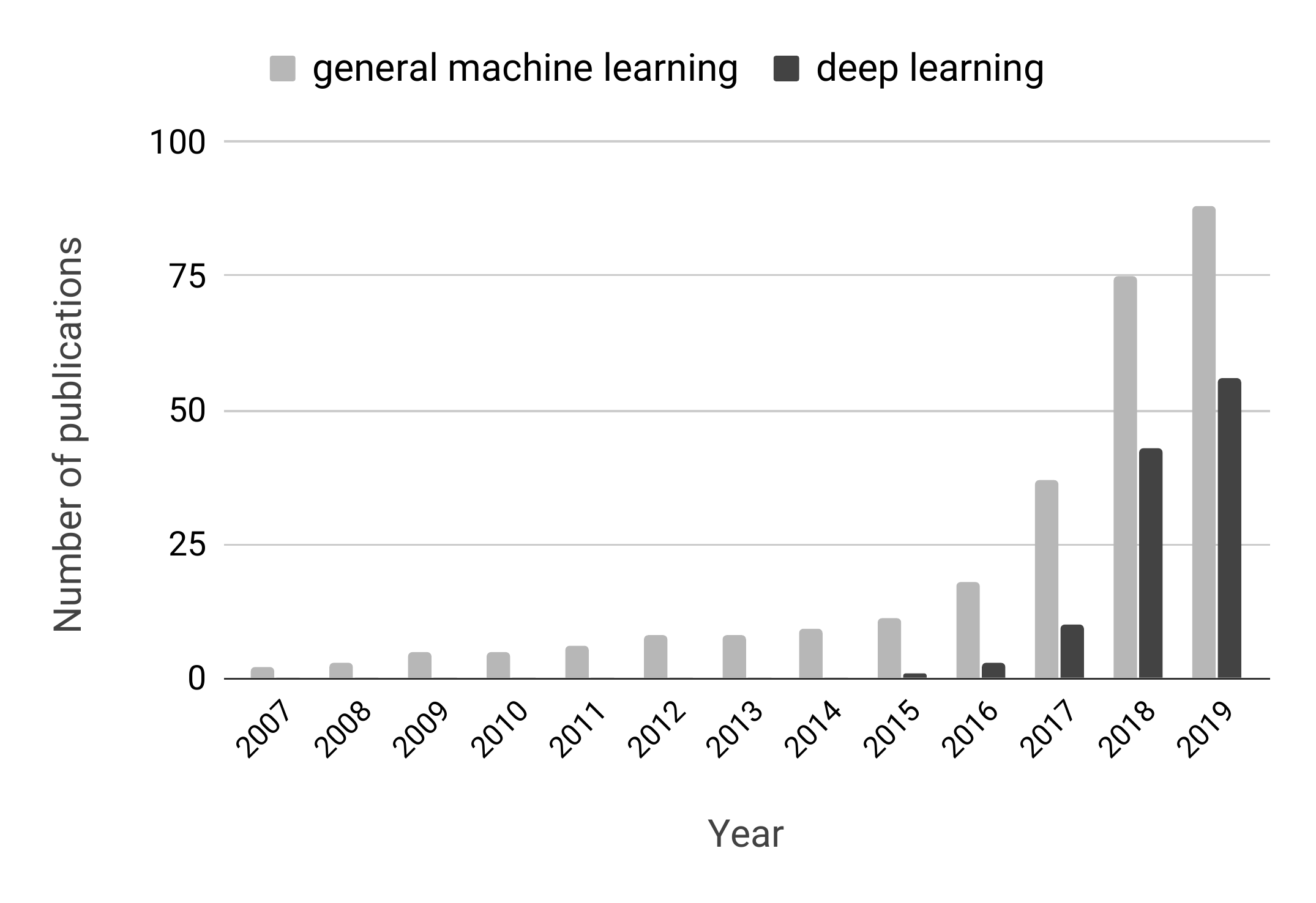}
\caption{Commutative trends in general machine learning and deep learning}
  \label{fig:mltypepublications}
\end{figure}

We further investigated the number of papers in each category for each year, to observe whether there is a trend of moving from testing general machine learning to deep learning.
Figure~\ref{fig:mltypepublications} shows the results.
Before 2017, papers mostly focus on general machine learning; 
after 2018, both general machine learning and deep learning specific testing notably arise.

\subsubsection{Research Distribution among Supervised/Unsupervised/Reinforcement Learning Testing}

We further classified the papers based on the three machine learning categories: 
1) supervised learning testing, 2) unsupervised learning testing, and 3) reinforcement learning testing.
One striking finding is that almost all the work we identified in this survey focused on testing supervised machine learning. 
Among the \papernum related papers, there are currently only three papers testing unsupervised machine learning: 
Murphy et al.~\cite{Murphy2008PropertiesOM} introduced metamorphic relations that work for both supervised and unsupervised learning algorithms.
Ramanathan and Pullum~\cite{Ramanathan2016} proposed  a combination of symbolic and statistical approaches to test the $k$-means clustering algorithm.
Xie et al.~\cite{xie2018mettle} designed metamorphic relations for unsupervised learning.
We were able to find out only one paper that focused on reinforcement learning testing: 
Uesato et al.~\cite{uesato2018rigorous} proposed a predictive adversarial example generation approach to predict failures and estimate reliable risks in reinforcement learning.

Because of this notable imbalance, we sought to understand whether there was also an imbalance of research popularity in the machine learning areas.
To approximate the research popularity of each category, we searched terms `supervised learning', `unsupervised learning', and `reinforcement learning' in Google Scholar and Google.
Table~\ref{tab:researchhit} shows the results of search hits.
The last column shows the number/ratio of papers that touch each machine learning category in ML testing.
For example, 119 out of \papernum papers were observed for supervised learning testing purpose.
The table suggests that testing popularity of different categories is not related to their overall research popularity. 
In particular, reinforcement learning has higher search hits than supervised learning, but we did not observe any related work that conducts direct reinforcement learning testing.

There may be several reasons for this observation.
First, supervised learning is a widely-known learning scenario
associated with classification, regression, and ranking problems~\cite{mohri2012foundations}. 
It is natural that researchers would emphasise the testing of widely-applied, known and familiar techniques at the beginning.
Second, supervised learning usually has labels in the dataset. 
It is thereby easier to judge and analyse test effectiveness.

\begin{table}[t]\small	
	\center
	\caption{\label{tab:researchhit}Search hits and testing distribution of supervised/unsupervised/reinforcement Learning }
	\begin{tabular}{lrrr}
		\toprule
		Category&Scholar hits& Google hits& Testing hits\\ \hline
		Supervised&1,610k&73,500k&119/\papernum \\
		Unsupervised&619k&17,700k&3/\papernum \\ 
		Reinforcement&2,560k&74,800k&1/\papernum \\
		\bottomrule		
	\end{tabular}
\end{table}

Nevertheless, many opportunities clearly remain for research in the widely-studied areas of unsupervised learning and reinforcement learning (we discuss more in Section~\ref{sec:researchdirection}).

\subsubsection{Different Learning Tasks}

ML involves different tasks such as classification, regression, clustering, and dimension reduction (see more in Section~\ref{sec:preliminaries}).
The research focus on different tasks also appears to exhibit imbalance, with a large number of papers focusing on classification.

%
%
%
%
%
%




\subsection{Research Distribution among Different Testing Properties}
\label{sec:distributionproperty}

\begin{figure}[h!]
  \centering
\includegraphics[scale=0.4]{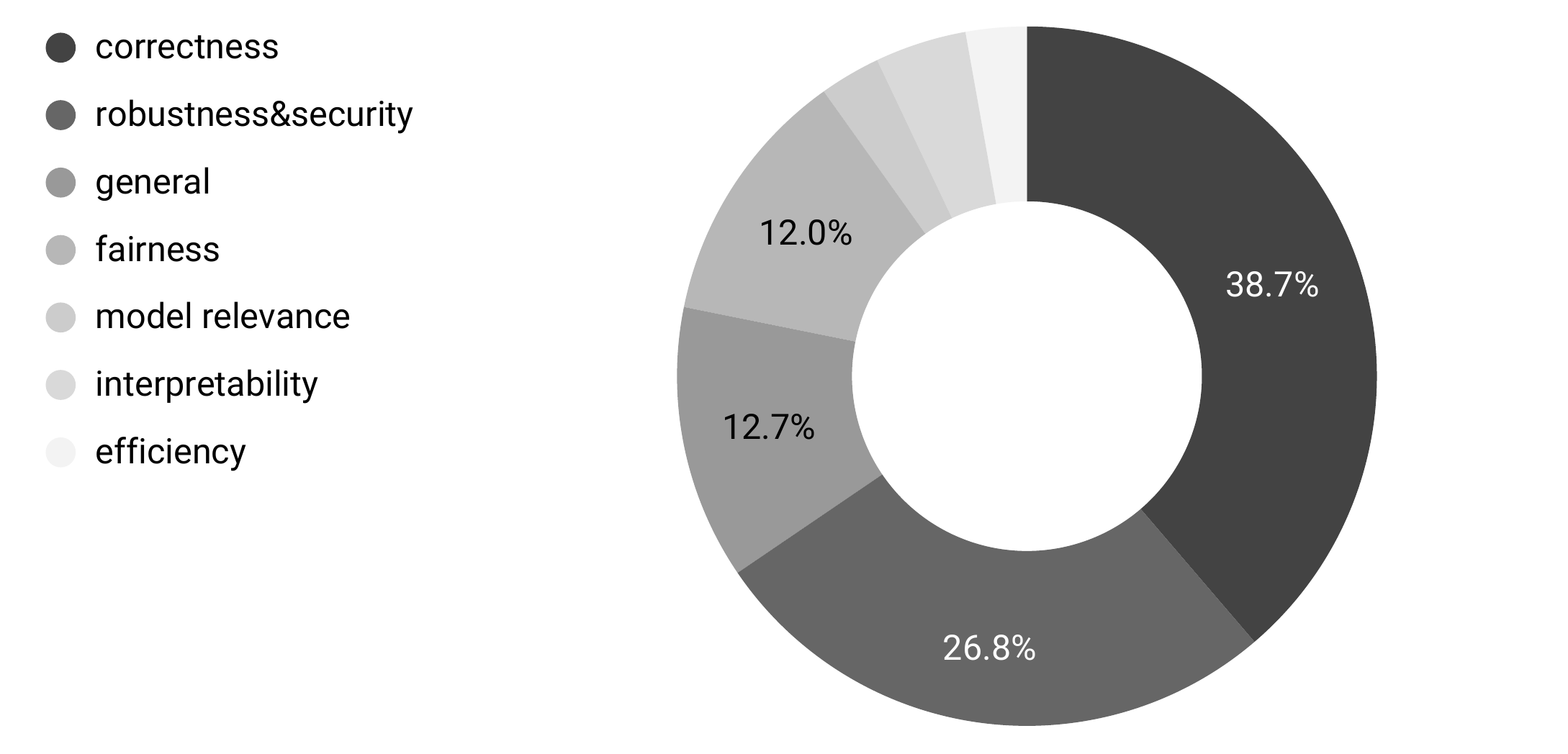}
\caption{Research distribution among different testing properties}
  \label{fig:piechartproperties}
\end{figure}

We counted the number of papers concerning each \mlt{} property.
Figure~\ref{fig:piechartproperties} shows the results.
The properties in the legend are ranked based on the number of papers that are specially focused on testing the associated property (`general' refers to those papers 
discussing or surveying ML testing generally).

From the figure, 
38.7\% of the papers test correctness.
26.8\% of the papers focus on robustness and security problems.
Fairness testing ranks the third among all the properties, with 12.0\% of the papers.

Nevertheless, for model relevance, interpretability testing, efficiency testing, and privacy testing, fewer than 6 papers exist for each category in our paper collection.

\subsection{Datasets Used in ML Testing}
\label{sec:dataset}


\begin{table*}[h!]\small
	\center
	\caption{\label{tab:dataset1}Datasets (1/4): Image Classification}
	\begin{tabular}{p{3.9cm}| p{7.5cm}| p{2cm}| p{2.8cm}}
		\toprule
		\textbf{Dataset}&\textbf{Description}&\textbf{Size}&\textbf{Usage}\\ \midrule 
		MNIST~\cite{mnisthandwrittendigit}&Images of handwritten digits&60,000+10,000&correctness, overfitting, robustness \\ \midrule
		Fashion MNIST~\cite{Xiao2017FashionMNISTAN}&MNIST-like dataset of fashion images&70,000&correctness, overfitting\\ \midrule
		CIFAR-10~\cite{CIFARdataset}&General images with 10 classes&50,000+10,000&correctness, overfitting, robustness\\ \midrule
		ImageNet~\cite{imagenet_cvpr09}&Visual recognition challenge dataset &14,197,122&correctness, robustness\\ \midrule
		IRIS flower~\cite{dataset_irisflower}&The Iris flowers&150&overfitting\\ \midrule
		SVHN~\cite{netzer2011reading}&House numbers&73,257+26,032&correctness,robustness\\ \midrule
		Fruits 360~\cite{fruit360}&Dataset with 65,429 images of 95 fruits&65,429&correctness,robustness\\ \midrule
		Handwritten Letters~\cite{dataset_kaggleletters}&Colour images of Russian letters&1,650&correctness,robustness\\ \midrule
		Balance Scale~\cite{dataset_waveform}&Psychological experimental results&625&overfitting\\ \midrule
		DSRC~\cite{dataset_DSRC}&Wireless communications between vehicles and road side units&10,000&overfitting, robustness\\ \midrule
		Udacity challenge~\cite{udacity}&Udacity Self-Driving Car Challenge images &101,396+5,614&robustness\\ \midrule
		Nexar traffic light challenge~\cite{getnexar}&Dashboard camera& 18,659+500,000&robustness\\ \midrule
		MSCOCO~\cite{dataset_coco}&Object recognition 	&160,000&correctness\\ \midrule
		Autopilot-TensorFlow~\cite{pan2017virtual}& Recorded to test the NVIDIA Dave model&45,568&robustness \\ \midrule
		KITTI~\cite{dataset_KITTI}&Six different scenes
captured by a VW Passat station wagon equipped with four
video cameras	&14,999 &robustness \\  
		\bottomrule		
	\end{tabular}
\end{table*}

Tables~\ref{tab:dataset1} to ~\ref{tab:dataset5} 
show details concerning widely-adopted datasets used in \mlt{} research.
In each table, the first column shows the name and link of each dataset.
The next three columns give a brief description, the size (the ``+'' connects training data and test data if applicable), the testing problem(s), the usage application scenario of each dataset\footnote{These tables do not list datasets adopted in data cleaning evaluation, because such studies usually involve hundreds of data sets~\cite{krishnan2017boostclean}}.

Table~\ref{tab:dataset1} shows the datasets used for image classification tasks.
Datasets can be large (e.g., more than 1.4 million images in ImageNet).
The last six rows show the datasets collected for autonomous driving system testing.
Most image datasets are adopted to test correctness, overfitting, and robustness of ML systems.

Table~\ref{tab:dataset3} shows datasets related to natural language processing. 
The contents are usually text, sentences, or text files, applied to scenarios like robustness and correctness.
\begin{table*}[h!]\small
	\center
	\caption{\label{tab:dataset3}Datasets (2/4): Natural Language Processing}
	\begin{tabular}{p{3.9cm}|  p{7.5cm}| p{2cm}| p{2.9cm}}
		\toprule
		\textbf{Dataset}&\textbf{Description}&\textbf{Size}&\textbf{Usage}\\ \midrule 
			bAbI~\cite{bAbi}&questions and answers for NLP&1000+1000&robustness\\ \midrule
			Tiny Shakespeare~\cite{karpathy2015visualizing}&Samples from actual Shakespeare&100,000 character&correctness\\  \midrule
		Stack Exchange Data Dump~\cite{dataset_stackoverflow}&Stack Overflow questions and answers&365 files&correctness\\ \midrule
		SNLI~\cite{bowman2015large}&Stanford Natural Language Inference Corpus&570,000&robustness\\ \midrule
		MultiNLI~\cite{N18-1101}& Crowd-sourced collection of sentence pairs annotated with textual entailment information&433,000&robustness\\  \midrule
		DMV failure reports~\cite{dataset_DVM}&AV failure reports from 12 manufacturers in California\footnote{The 12 AV manufacturers are: Bosch, Delphi Automotive, Google, Nissan, Mercedes- Benz, Tesla Motors, BMW, GM, Ford, Honda, Uber, and Volkswagen.} &keep  updating&correctness \\
		\bottomrule		
	\end{tabular}
\end{table*}

The datasets used to make decisions are introduced in Table~\ref{tab:dataset3}.
They are usually records with personal information, and thus are widely adopted to test the fairness of the ML models.

\begin{figure}[h!]
  \centering
\includegraphics[scale=0.46]{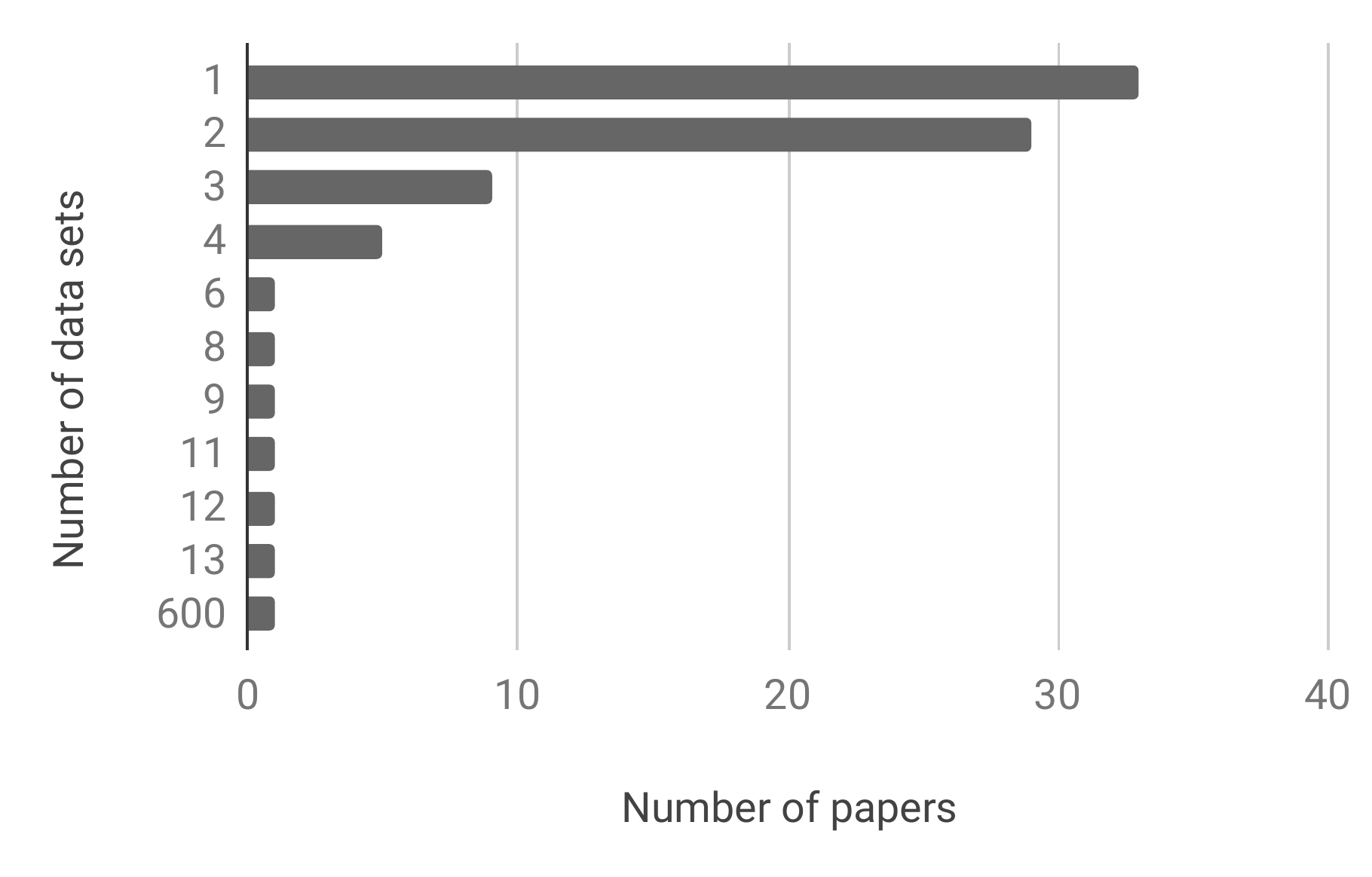}
\caption{Number of papers with different amounts of datasets in experiments}
  \label{fig:datasetnumber}
\end{figure}

We also calculate how many datasets an \mlt{} paper usually uses in its evaluation (for those papers with an evaluation).
Figure~\ref{fig:datasetnumber} shows the results.
Surprisingly, most papers use only one or two datasets in their evaluation;
One reason might be training and testing machine learning models have high costs.
There is one paper with as many as 600 datasets, but that paper used these datasets to evaluate data cleaning techniques, which has relatively low cost~\cite{hynes2017data}.

We also discuss research directions of building dataset and benchmarks  for ML testing in Section~\ref{sec:researchdirection}.

\begin{table*}[h!]\small
	\center
	\caption{\label{tab:dataset4}Datasets (3/4): Records for Decision Making}
	\begin{tabular}{p{3.9cm}|  p{7.5cm}| p{2cm}| p{2.9cm}}
		\toprule
		\textbf{Dataset}&\textbf{Description}&\textbf{Size}&\textbf{Usage}\\ \midrule 
		German Credit~\cite{dataset_germancredit}&Descriptions of customers with good and bad credit risks&1,000&fairness\\ \midrule
		Adult ~\cite{dataset_adult}&Census income& 48,842&fairness\\ \midrule
		Bank Marketing~\cite{moro2014data}&Bank client subscription term deposit data& 45,211&fairness\\ \midrule
		US Executions~\cite{dataset_usexecution}&Records of every execution performed in the United States&1,437&fairness\\ \midrule
		Fraud Detection~\cite{dal2015calibrating}&European Credit cards transactions&284,807 &fairness\\ \midrule
		Berkeley Admissions Data~\cite{bickel1975sex}&Graduate school applications to the six largest departments at University of California, Berkeley in 1973&4,526&fairness\\ \midrule
		Broward County COMPAS~\cite{dataset_compas}&Score to determine whether to release a defendant & 18,610&fairness\\ \midrule
		MovieLens Datasets~\cite{harper2016movielens}&People's preferences for movies& 100k-20m&fairness\\ \midrule
		Zestimate~\cite{dataset_zillow}&data about homes and Zillow's in-house price and predictions&2,990,000&correctness\\  \midrule
		FICO scores~\cite{reserve2007report}&United States credit worthiness&301,536 &fairness\\  \midrule
		Law school success~\cite{dataset_LSAC}&Information concerning law students from 163 law schools in the United States& 21,790&fairness\\
		\bottomrule		
	\end{tabular}
\end{table*}

\begin{table*}[h!]\small
	\center
	\caption{\label{tab:dataset5}Datasets (4/4): Others}
	\begin{tabular}{p{3.9cm}|  p{7.5cm}| p{2cm}| p{2.9cm}}
		\toprule
		\textbf{Dataset}&\textbf{Description}&\textbf{Size}&\textbf{Usage}\\ \midrule 
		VirusTotal~\cite{dataset_VirusTotal}&Malicious PDF files&5,000&robustness \\ \midrule
		Contagio~\cite{dataset_contagio}&Clean and malicious files&28,760&robustness  \\  \midrule
		Drebin~\cite{arp2014drebin}&Applications from different malware families&123,453&robustness\\ \midrule
		Chess~\cite{dataset_chess} &Chess game data: King+Rook versus King+Pawn on a7&3,196&correctness\\ \midrule
		Waveform~\cite{dataset_waveform}&CART book's generated waveform data&5,000&correctness\\

		\bottomrule		
	\end{tabular}
\end{table*}


\subsection{Open-source Tool Support in ML Testing}
\label{sec:tooltechnique}

There are several tools specially designed for ML testing.
Angell et al. presented Themis~\cite{angell2018themis}, an open-source tool for testing group discrimination\footnote{\url{http://fairness.cs.umass.edu/}}.
There is also an ML testing framework for tensorflow, named \emph{mltest}\footnote{\url{https://github.com/Thenerdstation/mltest}},
for writing simple ML unit tests.
Similar to \emph{mltest}, there is a testing framework for writing unit tests for pytorch-based ML systems, named \emph{torchtest}\footnote{\url{https://github.com/suriyadeepan/torchtest}}.
Dolby et al.~\cite{Dolby2018Ariadne} extended WALA to enable static analysis for machine learning code using TensorFlow.

Compared to traditional testing, the existing tool support in \mlt{} is relatively immature.
There remains plenty of space for tool-support improvement for \mlt{}.


\section{Challenges and Opportunities}
\label{sec:researchdirection}

This section discusses the challenges (Section~\ref{sec:challenges}) and research opportunities in ML testing (Section~\ref{sec:researchopport}).

\subsection{Challenges in ML Testing}
\label{sec:challenges}
As this survey reveals, ML testing has experienced rapid recent growth.
Nevertheless, ML testing remains at an early stage in its development, with many challenges and open questions lying ahead.

\noindent\textbf{Challenges in Test Input Generation}.
Although a range of test input generation techniques have been proposed (see more in Section~\ref{sec:input}), test input generation remains  challenging because of the large behaviour space of ML models.

Search-based Software Test generation (SBST)~\cite{mcminn2004search} uses a meta-heuristic optimising search technique, such as a Genetic Algorithm, to automatically generate test inputs.
It is a test generation technique that has been widely used in research (and deployment~\cite{harman2018start}) for traditional software testing paradigms.
As well as generating test inputs for testing functional properties like program correctness, SBST has also been used to explore tensions in algorithmic fairness in requirement analysis.~\cite{afetal:re08,afetal:fairness-jv}.
SBST has been successfully applied in testing autonomous driving systems~\cite{ben2016testing,abdessalem2018testing,abdessalem2018testing2}.
There exist many research opportunities in applying SBST on generating test inputs for testing other ML systems, since there is a clear apparent fit between SBST and ML; SBST adaptively searches for test inputs in large input spaces.

Existing test input generation techniques focus on generating adversarial inputs to test the robustness of an ML system.
However, adversarial examples are often criticised because they do not represent real input data.
Thus, an interesting research direction is to how to generate natural test inputs and how to automatically measure the naturalness of the generated inputs. 

There has been work that tries to generate test inputs to be as natural as possible under the scenario of autonomous driving, such as DeepTest~\cite{tian2018deeptest}, DeepHunter~\cite{Ma2018ase} and DeepRoad~\cite{deeproad}, yet the generated images could still suffer from unnaturalness: sometimes even humans may not recognise the images generated by these tools.
It is both interesting and challenging to explore whether such kinds of test data that are meaningless to humans should be adopted/valid in \mlt{}.




\noindent\textbf{Challenges on Test Assessment Criteria}.
There has been a lot of work exploring how to assess the quality or adequacy of test data (see more in Section~\ref{sec:evaluation}).
However, there is still a lack of systematic evaluation about how different assessment metrics correlate, or how these assessment metrics correlate with tests' fault-revealing ability, a topic that has been widely studied in traditional software testing~\cite{zhang2019pseudo}.
\jienew{The relation between test assessment criteria and test sufficiency remains unclear.
In addition, assessment criteria may provide a way of interpreting and understanding behaviours of ML models, which might be an interesting direction for further exploration.}

\noindent\textbf{Challenges Relating to The Oracle Problem}.
The oracle problem remains a challenge in ML testing. 
Metamorphic relations are effective pseudo oracles but, in most cases, they need to be defined by human ingenuity . 
A remaining challenge is thus to automatically identify and construct reliable test oracles for ML testing.


Murphy et al.~\cite{Murphy2009AST} discussed how flaky tests are likely to arise in metamorphic testing whenever floating point calculations are involved.
Flaky test detection is a challenging problem in traditional software testing~\cite{harman2018start}.
It is perhaps more challenging in ML testing because of the oracle problem.

Even without flaky tests, pseudo oracles may be inaccurate, leading to many false positives.
There is, therefore, a need to explore how to yield more accurate test oracles and how to reduce the false positives among the reported issues.
We could even use ML algorithm $b$ to learn to detect false-positive oracles when testing ML algorithm $a$.


\noindent\textbf{Challenges in Testing Cost Reduction}.
In traditional software testing, the cost problem remains a big problem, yielding many cost reduction techniques such as test selection, test prioritisation, and predicting test execution results. 
In ML testing, the cost problem could be more serious, especially when testing the ML component, because ML component testing usually requires model retraining or repeating of the prediction process. It may also require data generation to explore the enormous mode behaviour space.

A possible research direction for cost reduction is to represent an ML model as an intermediate state to make it easier for testing.

We could also apply traditional cost reduction techniques such as test prioritisation or minimisation to reduce the size of test cases without affecting the test correctness.

More ML solutions are deployed to diverse devices and platforms~(e.g., mobile device, IoT edge device). Due to the resource limitation of a target device, how to effectively test ML model on diverse devices as well as the deployment process would be also a challenge.


\subsection{Research Opportunities in \mlt{}}
\label{sec:researchopport}

There remain many research opportunities in \mlt{}. \jienew{These are not necessarily research challenges, but may greatly benefit machine learning developers and users as well as the whole research community.}

\noindent\textbf{Testing More Application Scenarios}.
Much current research focuses on supervised learning, in particular classification problems. 
More research is needed on problems associated with testing unsupervised and reinforcement learning.

The testing tasks currently tackled in the literature, primarily centre on image classification. 
There remain open exciting testing research opportunities in many other areas, such as speech recognition, natural language processing and agent/game play.

\noindent\textbf{Testing More ML Categories and Tasks}.
We observed pronounced imbalance regarding the coverage of testing techniques for different machine learning categories and tasks, as demonstrated by Table~\ref{tab:researchhit}.
There are both challenges and research opportunities for testing unsupervised  and reinforcement learning systems.

For instance, transfer learning, a topic gaining much recent interest,  focuses on storing knowledge gained while solving one problem and applying it to a different but related problem~\cite{weiss2016survey}.
Transfer learning testing is also important, yet poorly covered in the existing literature.

\noindent\textbf{Testing Other Properties}.
From Figure~\ref{fig:piechartproperties}, we can see that most work tests robustness and correctness, while relatively few papers (less than 3\%) study efficiency, model relevance, or interpretability.

\jie{Model relevance testing is challenging because the distribution of the future data is often unknown, while the capacity of many models is also unknown and hard to measure. It might be interesting to conduct empirical studies on the prevalence of poor model relevance among ML models as well as on the balance between poor model relevance and high security risks.}

\jienew{For testing efficiency, there is a need to test the efficiency at different levels such as the efficiency when switching among different platforms, machine learning frameworks, and hardware devices.}


For testing property interpretability, existing approaches rely primarily on manual assessment, which checks whether humans could understand the logic or predictive results of an ML model. It will be also interesting to investigate the automatic assessment of interpretability and the detection of interpretability violations.

\jienew{There is a lack of consensus regarding the definitions and understanding of fairness and interpretability. There is thus a need for clearer definitions, formalisation, and empirical studies under different contexts.}

There has been a discussion that machine learning testing and traditional software testing may have different requirements in the assurance to be expected for different properties~\cite{nakajima2018quality}. 
Therefore, more work is needed to explore and identify those properties that are most important for machine learning systems, and thus deserve more research and test effort.

\noindent\textbf{Presenting More Testing Benchmarks}
A large number of datasets have been adopted in the existing \mlt{} papers.
As Tables~\ref{tab:dataset1} to ~\ref{tab:dataset5} show, these datasets are usually those adopted for building machine learning systems. 
As far as we know, there are very few benchmarks like CleverHans\footnote{\url{https://github.com/tensorflow/cleverhans}} that are specially designed for \mlt{} research purposes, such as adversarial example construction.

More benchmarks are needed, that are specially designed for \mlt{}.
For example, a repository of machine learning programs with real bugs would present a good benchmark for bug-fixing techniques.
Such an \mlt{} repository, would play a similar (and equally-important) role to that played by data sets such as Defects4J\footnote{\url{https://github.com/rjust/defects4j}} in traditional software testing.

\noindent\textbf{\jie{Covering More Testing Activities}}.
As far we know, requirement analysis for ML systems remains absent in the ML testing literature.
As demonstrated by Finkelstein et al.~\cite{afetal:re08,afetal:fairness-jv}, a good requirements analysis may tackle many non-functional properties such as fairness. 

Existing work is focused on off-line testing. 
Online-testing deserves more research efforts.

According to the work of Amershi et al.~\cite{Amershi2019se4ai}, data testing is especially important.
This topic certainly deserves more research effort.
Additionally, there are also many opportunities for regression testing, bug report analysis, and bug triage in ML testing.

Due to the black-box nature of machine learning algorithms, \mlt{} results are often more difficult for developers to understand, compared to traditional software testing.
Visualisation of testing results might be particularly helpful in \mlt{} to help developers understand the bugs and help with the bug localisation and repair. 

\noindent\textbf{Mutating Investigation in Machine Learning System}.
There have been some studies discussing mutating machine learning code~\cite{Murphy2009AST, cheng2018manifesting}, but no work has explored how to better design mutation operators for machine learning code so that the mutants could better simulate real-world machine learning bugs.
This is another research opportunity.


\section{Conclusion}
\label{sec:conclusion}

We provided a comprehensive overview and analysis of research work on ML testing. The survey presented the definitions and current research status of different ML testing properties, testing components, and testing workflows.
It also summarised the datasets used for experiments and the available open-source testing tools/frameworks,
and analysed the research trends, directions, opportunities, and challenges in ML testing.
We hope this survey will help software engineering and machine learning researchers to become familiar with the current status and open opportunities of and for of \mlt{}.

\section*{Acknowledgement}
\label{sec:acknowledgement}
Before submitting, we sent the paper to those whom we cited, to check our comments for accuracy and omission. This also provided one final stage in the systematic trawling of the literature for relevant work. Many thanks to those members of the community who kindly provided comments and feedback on earlier drafts of this paper.

%
%
%
%
%
%
%
%

    
	\bibliographystyle{unsrt}
	\bibliography{reference}

\begin{thebibliography}{100}

\bibitem{pei2017deepxplore}
Kexin Pei, Yinzhi Cao, Junfeng Yang, and Suman Jana.
\newblock Deepxplore: Automated whitebox testing of deep learning systems.
\newblock In {\em Proceedings of the 26th Symposium on Operating Systems
  Principles}, pages 1--18. ACM, 2017.

\bibitem{chen2015deepdriving}
Chenyi Chen, Ari Seff, Alain Kornhauser, and Jianxiong Xiao.
\newblock Deepdriving: Learning affordance for direct perception in autonomous
  driving.
\newblock In {\em Proceedings of the IEEE International Conference on Computer
  Vision}, pages 2722--2730, 2015.

\bibitem{litjens2017survey}
Geert Litjens, Thijs Kooi, Babak~Ehteshami Bejnordi, Arnaud Arindra~Adiyoso
  Setio, Francesco Ciompi, Mohsen Ghafoorian, Jeroen~Awm Van Der~Laak, Bram
  Van~Ginneken, and Clara~I S{\'a}nchez.
\newblock A survey on deep learning in medical image analysis.
\newblock {\em Medical image analysis}, 42:60--88, 2017.

\bibitem{ammann2016introduction}
Paul Ammann and Jeff Offutt.
\newblock {\em Introduction to software testing}.
\newblock Cambridge University Press, 2016.

\bibitem{galhotra2017fairness}
Sainyam Galhotra, Yuriy Brun, and Alexandra Meliou.
\newblock Fairness testing: testing software for discrimination.
\newblock In {\em Proceedings of the 2017 11th Joint Meeting on Foundations of
  Software Engineering}, pages 498--510. ACM, 2017.

\bibitem{Paculablog2011}
Maciej Pacula.
\newblock {Unit-Testing Statistical Software}.
\newblock
  \url{http://blog.mpacula.com/2011/02/17/unit-testing-statistical-software/},
  2011.

\bibitem{Ramanathan2016}
A.~Ramanathan, L.~L. Pullum, F.~Hussain, D.~Chakrabarty, and S.~K. Jha.
\newblock Integrating symbolic and statistical methods for testing intelligent
  systems: Applications to machine learning and computer vision.
\newblock In {\em 2016 Design, Automation Test in Europe Conference Exhibition
  (DATE)}, pages 786--791, March 2016.

\bibitem{Amershi2019se4ai}
Saleema Amershi, Andrew Begel, Christian Bird, Rob DeLine, Harald Gall, Ece
  Kamar, Nachi Nagappan, Besmira Nushi, and Tom Zimmermann.
\newblock Software engineering for machine learning: A case study.
\newblock In {\em Proc. ICSE (Industry Track)}, pages 291--300, 2019.

\bibitem{barr2015oracle}
Earl~T Barr, Mark Harman, Phil McMinn, Muzammil Shahbaz, and Shin Yoo.
\newblock {The oracle problem in software testing: A survey}.
\newblock {\em IEEE transactions on software engineering}, 41(5):507--525,
  2015.

\bibitem{murphy2007approach}
Christian Murphy, Gail~E Kaiser, and Marta Arias.
\newblock An approach to software testing of machine learning applications.
\newblock In {\em Proc. SEKE}, volume 167, 2007.

\bibitem{Davis1981}
Martin~D. Davis and Elaine~J. Weyuker.
\newblock Pseudo-oracles for non-testable programs.
\newblock In {\em Proceedings of the ACM 81 Conference}, ACM 81, pages
  254--257, 1981.

\bibitem{kaetal:analysis}
Kelly Androutsopoulos, David Clark, Haitao Dan, Mark Harman, and Robert
  Hierons.
\newblock An analysis of the relationship between conditional entropy and
  failed error propagation in software testing.
\newblock In {\em Proc. ICSE}, pages 573--583, Hyderabad, India, June 2014.

\bibitem{voas:testability}
Jeffrey~M. Voas and Keith~W. Miller.
\newblock Software testability: The new verification.
\newblock {\em {IEEE} Software}, 12(3):17--28, May 1995.

\bibitem{clark:squeeziness}
David Clark and Robert~M. Hierons.
\newblock Squeeziness: {A}n information theoretic measure for avoiding fault
  masking.
\newblock {\em Information Processing Letters}, 112(8--9):335 -- 340, 2012.

\bibitem{yjmh:hom-scam-best-paper}
Yue Jia and Mark Harman.
\newblock Constructing subtle faults using higher order mutation testing (best
  paper award winner).
\newblock In {\em 8th International Working Conference on Source Code Analysis
  and Manipulation ({SCAM'08})}, pages 249--258, Beijing, China, 2008. {IEEE}
  Computer Society.

\bibitem{turner1993state}
Christopher~D Turner and David~J Robson.
\newblock The state-based testing of object-oriented programs.
\newblock In {\em 1993 Conference on Software Maintenance}, pages 302--310.
  IEEE, 1993.

\bibitem{harman2001search}
Mark Harman and Bryan~F Jones.
\newblock Search-based software engineering.
\newblock {\em Information and software Technology}, 43(14):833--839, 2001.

\bibitem{harman2010theoretical}
Mark Harman and Phil McMinn.
\newblock A theoretical and empirical study of search-based testing: Local,
  global, and hybrid search.
\newblock {\em IEEE Transactions on Software Engineering}, 36(2):226--247,
  2010.

\bibitem{memon2002gui}
Atif~M Memon.
\newblock Gui testing: Pitfalls and process.
\newblock {\em Computer}, (8):87--88, 2002.

\bibitem{sen2005cute}
Koushik Sen, Darko Marinov, and Gul Agha.
\newblock Cute: a concolic unit testing engine for c.
\newblock In {\em ACM SIGSOFT Software Engineering Notes}, volume~30, pages
  263--272. ACM, 2005.

\bibitem{Hains2018}
G.~Hains, A.~Jakobsson, and Y.~Khmelevsky.
\newblock Towards formal methods and software engineering for deep learning:
  Security, safety and productivity for {DL} systems development.
\newblock In {\em 2018 Annual IEEE International Systems Conference (SysCon)},
  pages 1--5, April 2018.

\bibitem{ma2018secure}
Lei Ma, Felix Juefei-Xu, Minhui Xue, Qiang Hu, Sen Chen, Bo~Li, Yang Liu,
  Jianjun Zhao, Jianxiong Yin, and Simon See.
\newblock Secure deep learning engineering: A software quality assurance
  perspective.
\newblock {\em arXiv preprint arXiv:1810.04538}, 2018.

\bibitem{huang2018safety}
Xiaowei Huang, Daniel Kroening, Marta Kwiatkowska, Wenjie Ruan, Youcheng Sun,
  Emese Thamo, Min Wu, and Xinping Yi.
\newblock Safety and trustworthiness of deep neural networks: A survey.
\newblock {\em arXiv preprint arXiv:1812.08342}, 2018.

\bibitem{masuda2018survey}
Satoshi Masuda, Kohichi Ono, Toshiaki Yasue, and Nobuhiro Hosokawa.
\newblock A survey of software quality for machine learning applications.
\newblock In {\em 2018 IEEE International Conference on Software Testing,
  Verification and Validation Workshops (ICSTW)}, pages 279--284. IEEE, 2018.

\bibitem{ishikawa2018concepts}
Fuyuki Ishikawa.
\newblock Concepts in quality assessment for machine learning-from test data to
  arguments.
\newblock In {\em International Conference on Conceptual Modeling}, pages
  536--544. Springer, 2018.

\bibitem{Braiek2018}
Houssem Braiek and Foutse Khomh.
\newblock On testing machine learning programs.
\newblock {\em arXiv preprint arXiv:1812.02257}, 2018.

\bibitem{shawe2004kernel}
John Shawe-Taylor, Nello Cristianini, et~al.
\newblock {\em Kernel methods for pattern analysis}.
\newblock Cambridge university press, 2004.

\bibitem{mohri2012foundations}
Mehryar Mohri, Afshin Rostamizadeh, and Ameet Talwalkar.
\newblock {\em Foundations of machine learning}.
\newblock MIT press, 2012.

\bibitem{paszke2017automatic}
Adam Paszke, Sam Gross, Soumith Chintala, Gregory Chanan, Edward Yang, Zachary
  DeVito, Zeming Lin, Alban Desmaison, Luca Antiga, and Adam Lerer.
\newblock Automatic differentiation in pytorch.
\newblock 2017.

\bibitem{tensorflow2015whitepaper}
Mart\'{\i}n Abadi, Ashish Agarwal, Paul Barham, Eugene Brevdo, Zhifeng Chen,
  Craig Citro, Greg~S. Corrado, Andy Davis, Jeffrey Dean, Matthieu Devin,
  Sanjay Ghemawat, Ian Goodfellow, Andrew Harp, Geoffrey Irving, Michael Isard,
  Yangqing Jia, Rafal Jozefowicz, Lukasz Kaiser, Manjunath Kudlur, Josh
  Levenberg, Dan Man\'{e}, Rajat Monga, Sherry Moore, Derek Murray, Chris Olah,
  Mike Schuster, Jonathon Shlens, Benoit Steiner, Ilya Sutskever, Kunal Talwar,
  Paul Tucker, Vincent Vanhoucke, Vijay Vasudevan, Fernanda Vi\'{e}gas, Oriol
  Vinyals, Pete Warden, Martin Wattenberg, Martin Wicke, Yuan Yu, and Xiaoqiang
  Zheng.
\newblock {TensorFlow}: Large-scale machine learning on heterogeneous systems,
  2015.
\newblock Software available from tensorflow.org.

\bibitem{scikitlearn}
F.~Pedregosa, G.~Varoquaux, A.~Gramfort, V.~Michel, B.~Thirion, O.~Grisel,
  M.~Blondel, P.~Prettenhofer, R.~Weiss, V.~Dubourg, J.~Vanderplas, A.~Passos,
  D.~Cournapeau, M.~Brucher, M.~Perrot, and E.~Duchesnay.
\newblock {Scikit-learn: Machine Learning in Python }.
\newblock {\em Journal of Machine Learning Research}, 12:2825--2830, 2011.

\bibitem{chollet2015keras}
Fran\c{c}ois Chollet et~al.
\newblock Keras.
\newblock \url{https://keras.io}, 2015.

\bibitem{jia2014caffe}
Yangqing Jia, Evan Shelhamer, Jeff Donahue, Sergey Karayev, Jonathan Long, Ross
  Girshick, Sergio Guadarrama, and Trevor Darrell.
\newblock Caffe: Convolutional architecture for fast feature embedding.
\newblock {\em arXiv preprint arXiv:1408.5093}, 2014.

\bibitem{jordan2015machine}
Michael~I Jordan and Tom~M Mitchell.
\newblock Machine learning: Trends, perspectives, and prospects.
\newblock {\em Science}, 349(6245):255--260, 2015.

\bibitem{safavian1991survey}
S~Rasoul Safavian and David Landgrebe.
\newblock A survey of decision tree classifier methodology.
\newblock {\em IEEE transactions on systems, man, and cybernetics},
  21(3):660--674, 1991.

\bibitem{neter1996applied}
John Neter, Michael~H Kutner, Christopher~J Nachtsheim, and William Wasserman.
\newblock {\em Applied linear statistical models}, volume~4.
\newblock Irwin Chicago, 1996.

\bibitem{mccallum1998comparison}
Andrew McCallum, Kamal Nigam, et~al.
\newblock A comparison of event models for naive bayes text classification.
\newblock In {\em AAAI-98 workshop on learning for text categorization}, volume
  752, pages 41--48. Citeseer, 1998.

\bibitem{lecun2015deep}
Yann LeCun, Yoshua Bengio, and Geoffrey Hinton.
\newblock Deep learning.
\newblock {\em nature}, 521(7553):436, 2015.

\bibitem{kim2014convolutional}
Yoon Kim.
\newblock Convolutional neural networks for sentence classification.
\newblock {\em arXiv preprint arXiv:1408.5882}, 2014.

\bibitem{graves2013speech}
Alex Graves, Abdel-rahman Mohamed, and Geoffrey Hinton.
\newblock Speech recognition with deep recurrent neural networks.
\newblock In {\em 2013 IEEE international conference on acoustics, speech and
  signal processing}, pages 6645--6649. IEEE, 2013.

\bibitem{5399061}
Ieee standard classification for software anomalies.
\newblock {\em IEEE Std 1044-2009 (Revision of IEEE Std 1044-1993)}, pages
  1--23, Jan 2010.

\bibitem{werpachowski2019detecting}
Roman Werpachowski, Andr{\'a}s Gy{\"o}rgy, and Csaba Szepesv{\'a}ri.
\newblock Detecting overfitting via adversarial examples.
\newblock {\em arXiv preprint arXiv:1903.02380}, 2019.

\bibitem{cruz2018enabling}
Juan Cruz-Benito, Andrea V{\'a}zquez-Ingelmo, Jos{\'e}~Carlos
  S{\'a}nchez-Prieto, Roberto Ther{\'o}n, Francisco~Jos{\'e}
  Garc{\'\i}a-Pe{\~n}alvo, and Mart{\'\i}n Mart{\'\i}n-Gonz{\'a}lez.
\newblock Enabling adaptability in web forms based on user characteristics
  detection through a/b testing and machine learning.
\newblock {\em IEEE Access}, 6:2251--2265, 2018.

\bibitem{kaufmann2016complexity}
Emilie Kaufmann, Olivier Capp{\'e}, and Aur{\'e}lien Garivier.
\newblock On the complexity of best-arm identification in multi-armed bandit
  models.
\newblock {\em The Journal of Machine Learning Research}, 17(1):1--42, 2016.

\bibitem{breck2019datavalidation}
Eric Breck, Neoklis Polyzotis, Sudip Roy, Steven Whang, and Martin Zinkevich.
\newblock {Data Validation for Machine Learning}.
\newblock In {\em SysML}, 2019.

\bibitem{cheng2018towards}
Chih-Hong Cheng, Georg N{\"u}hrenberg, Chung-Hao Huang, and Hirotoshi Yasuoka.
\newblock Towards dependability metrics for neural networks.
\newblock {\em arXiv preprint arXiv:1806.02338}, 2018.

\bibitem{alfeld2016data}
Scott Alfeld, Xiaojin Zhu, and Paul Barford.
\newblock Data poisoning attacks against autoregressive models.
\newblock In {\em AAAI}, pages 1452--1458, 2016.

\bibitem{Pham2019cradle}
Weizhen Qi Lin~Tan Viet Hung~Pham, Thibaud~Lutellier.
\newblock Cradle: Cross-backend validation to detect and localize bugs in deep
  learning libraries.
\newblock In {\em Proc. ICSE}, 2019.

\bibitem{chung2012non}
Lawrence Chung, Brian~A Nixon, Eric Yu, and John Mylopoulos.
\newblock {\em Non-functional requirements in software engineering}, volume~5.
\newblock Springer Science \& Business Media, 2012.

\bibitem{afzal2009systematic}
Wasif Afzal, Richard Torkar, and Robert Feldt.
\newblock A systematic review of search-based testing for non-functional system
  properties.
\newblock {\em Information and Software Technology}, 51(6):957--976, 2009.

\bibitem{kirk2014thoughtful}
Matthew Kirk.
\newblock {\em Thoughtful machine learning: A test-driven approach}.
\newblock " O'Reilly Media, Inc.", 2014.

\bibitem{vapnik1994measuring}
Vladimir Vapnik, Esther Levin, and Yann~Le Cun.
\newblock Measuring the vc-dimension of a learning machine.
\newblock {\em Neural computation}, 6(5):851--876, 1994.

\bibitem{rosenberg2007rademacher}
David~S Rosenberg and Peter~L Bartlett.
\newblock The rademacher complexity of co-regularized kernel classes.
\newblock In {\em Artificial Intelligence and Statistics}, pages 396--403,
  2007.

\bibitem{zhang:hal-02139208}
Jie Zhang, Earl~T Barr, Benjamin Guedj, Mark Harman, and John Shawe-Taylor.
\newblock {Perturbed Model Validation: A New Framework to Validate Model
  Relevance}.
\newblock working paper or preprint, May 2019.

\bibitem{159342}
Ieee standard glossary of software engineering terminology.
\newblock {\em IEEE Std 610.12-1990}, pages 1--84, Dec 1990.

\bibitem{shahrokni2013systematic}
Ali Shahrokni and Robert Feldt.
\newblock A systematic review of software robustness.
\newblock {\em Information and Software Technology}, 55(1):1--17, 2013.

\bibitem{katz2017reluplex}
Guy Katz, Clark Barrett, David~L Dill, Kyle Julian, and Mykel~J Kochenderfer.
\newblock Reluplex: An efficient smt solver for verifying deep neural networks.
\newblock In {\em International Conference on Computer Aided Verification},
  pages 97--117. Springer, 2017.

\bibitem{dwork2011differential}
Cynthia Dwork.
\newblock Differential privacy.
\newblock {\em Encyclopedia of Cryptography and Security}, pages 338--340,
  2011.

\bibitem{VoigtGDPR}
Paul Voigt and Axel von~dem Bussche.
\newblock {\em The EU General Data Protection Regulation (GDPR): A Practical
  Guide}.
\newblock Springer Publishing Company, Incorporated, 1st edition, 2017.

\bibitem{californiaCCPA}
wikipedia.
\newblock {California Consumer Privacy Act}.
\newblock \url{https://en.wikipedia.org/wiki/California_Consumer_Privacy_Act}.

\bibitem{8258006}
R.~{Baeza-Yates} and Z.~{Liaghat}.
\newblock Quality-efficiency trade-offs in machine learning for text
  processing.
\newblock In {\em 2017 IEEE International Conference on Big Data (Big Data)},
  pages 897--904, Dec 2017.

\bibitem{corbett2018measure}
Sam Corbett-Davies and Sharad Goel.
\newblock The measure and mismeasure of fairness: A critical review of fair
  machine learning.
\newblock {\em arXiv preprint arXiv:1808.00023}, 2018.

\bibitem{days2004feedback}
Drew~S Days~III.
\newblock Feedback loop: The civil rights act of 1964 and its progeny.
\newblock {\em . Louis ULJ}, 49:981, 2004.

\bibitem{lipton2016mythos}
Zachary~C Lipton.
\newblock The mythos of model interpretability.
\newblock {\em arXiv preprint arXiv:1606.03490}, 2016.

\bibitem{doshi2017towards}
Finale Doshi-Velez and Been Kim.
\newblock Towards a rigorous science of interpretable machine learning.
\newblock {\em arXiv preprint arXiv:1702.08608}, 2017.

\bibitem{sellam2018deepbase}
Thibault Sellam, Kevin Lin, Ian~Yiran Huang, Michelle Yang, Carl Vondrick, and
  Eugene Wu.
\newblock Deepbase: Deep inspection of neural networks.
\newblock In {\em Proc. SIGMOD}, 2019.

\bibitem{biran2017explanation}
Or~Biran and Courtenay Cotton.
\newblock Explanation and justification in machine learning: A survey.
\newblock In {\em IJCAI-17 Workshop on Explainable AI (XAI)}, page~8, 2017.

\bibitem{miller2018explanation}
Tim Miller.
\newblock Explanation in artificial intelligence: Insights from the social
  sciences.
\newblock {\em Artificial Intelligence}, 2018.

\bibitem{goodman2016european}
Bryce Goodman and Seth Flaxman.
\newblock European union regulations on algorithmic decision-making and a"
  right to explanation".
\newblock {\em arXiv preprint arXiv:1606.08813}, 2016.

\bibitem{molnar}
Christoph Molnar.
\newblock {\em Interpretable Machine Learning}.
\newblock https://christophm.github.io/interpretable-ml-book/, 2019.
\newblock \url{https://christophm.github.io/interpretable-ml-book/}.

\bibitem{zhang2014search}
Jie Zhang, Junjie Chen, Dan Hao, Yingfei Xiong, Bing Xie, Lu~Zhang, and Hong
  Mei.
\newblock Search-based inference of polynomial metamorphic relations.
\newblock In {\em Proceedings of the 29th ACM/IEEE international conference on
  Automated software engineering}, pages 701--712. ACM, 2014.

\bibitem{o2018scalable}
Matthew O'Kelly, Aman Sinha, Hongseok Namkoong, Russ Tedrake, and John~C Duchi.
\newblock Scalable end-to-end autonomous vehicle testing via rare-event
  simulation.
\newblock In {\em Advances in Neural Information Processing Systems}, pages
  9827--9838, 2018.

\bibitem{xiang2018verification}
Weiming Xiang, Patrick Musau, Ayana~A Wild, Diego~Manzanas Lopez, Nathaniel
  Hamilton, Xiaodong Yang, Joel Rosenfeld, and Taylor~T Johnson.
\newblock Verification for machine learning, autonomy, and neural networks
  survey.
\newblock {\em arXiv preprint arXiv:1810.01989}, 2018.

\bibitem{wohlin2014guidelines}
Claes Wohlin.
\newblock Guidelines for snowballing in systematic literature studies and a
  replication in software engineering.
\newblock In {\em Proceedings of the 18th international conference on
  evaluation and assessment in software engineering}, page~38. Citeseer, 2014.

\bibitem{albarghouthi2019fairness}
Aws Albarghouthi and Samuel Vinitsky.
\newblock Fairness-aware programming.
\newblock In {\em Proceedings of the Conference on Fairness, Accountability,
  and Transparency}, pages 211--219. ACM, 2019.

\bibitem{tian2018deeptest}
Yuchi Tian, Kexin Pei, Suman Jana, and Baishakhi Ray.
\newblock {DeepTest: Automated testing of deep-neural-network-driven autonomous
  cars}.
\newblock In {\em Proceedings of the 40th International Conference on Software
  Engineering}, pages 303--314. ACM, 2018.

\bibitem{udacity}
udacity.
\newblock Udacity challenge.
\newblock \url{https://github.com/udacity/self-driving-car}.

\bibitem{goodfellow2014generative}
Ian Goodfellow, Jean Pouget-Abadie, Mehdi Mirza, Bing Xu, David Warde-Farley,
  Sherjil Ozair, Aaron Courville, and Yoshua Bengio.
\newblock Generative adversarial nets.
\newblock In {\em Advances in neural information processing systems}, pages
  2672--2680, 2014.

\bibitem{deeproad}
Mengshi Zhang, Yuqun Zhang, Lingming Zhang, Cong Liu, and Sarfraz Khurshid.
\newblock {DeepRoad: GAN-based Metamorphic Testing and Input Validation
  Framework for Autonomous Driving Systems}.
\newblock In {\em Proceedings of the 33rd ACM/IEEE International Conference on
  Automated Software Engineering}, ASE 2018, pages 132--142, 2018.

\bibitem{liu2017unsupervised}
Ming-Yu Liu, Thomas Breuel, and Jan Kautz.
\newblock Unsupervised image-to-image translation networks.
\newblock In {\em Advances in Neural Information Processing Systems}, pages
  700--708, 2017.

\bibitem{zhou2018deepbillboard}
Husheng Zhou, Wei Li, Yuankun Zhu, Yuqun Zhang, Bei Yu, Lingming Zhang, and
  Cong Liu.
\newblock Deepbillboard: Systematic physical-world testing of autonomous
  driving systems.
\newblock {\em arXiv preprint arXiv:1812.10812}, 2018.

\bibitem{du2018deepcruiser}
Xiaoning Du, Xiaofei Xie, Yi~Li, Lei Ma, Jianjun Zhao, and Yang Liu.
\newblock Deepcruiser: Automated guided testing for stateful deep learning
  systems.
\newblock {\em arXiv preprint arXiv:1812.05339}, 2018.

\bibitem{ding2016framework}
Junhua Ding, Dongmei Zhang, and Xin-Hua Hu.
\newblock A framework for ensuring the quality of a big data service.
\newblock In {\em 2016 IEEE International Conference on Services Computing
  (SCC)}, pages 82--89. IEEE, 2016.

\bibitem{rabin2019testing}
Md~Rafiqul~Islam Rabin, Ke~Wang, and Mohammad~Amin Alipour.
\newblock Testing neural program analyzers, 2019.

\bibitem{alon2019code2vec}
Uri Alon, Meital Zilberstein, Omer Levy, and Eran Yahav.
\newblock code2vec: Learning distributed representations of code.
\newblock {\em Proceedings of the ACM on Programming Languages}, 3(POPL):40,
  2019.

\bibitem{sun2019automatic}
Zeyu Sun, Jie~M. Zhang, Mark Harman, Mike Papadakis, and Lu~Zhang.
\newblock Automatic testing and improvement of machine translation.
\newblock In {\em Proc. ICSE (to appear)}, 2020.

\bibitem{mcminn2004search}
Phil McMinn.
\newblock Search-based software test data generation: a survey.
\newblock {\em Software testing, Verification and reliability}, 14(2):105--156,
  2004.

\bibitem{lakhotia2007multi}
Kiran Lakhotia, Mark Harman, and Phil McMinn.
\newblock A multi-objective approach to search-based test data generation.
\newblock In {\em Proceedings of the 9th annual conference on Genetic and
  evolutionary computation}, pages 1098--1105. ACM, 2007.

\bibitem{odena2018tensorfuzz}
Augustus Odena and Ian Goodfellow.
\newblock {TensorFuzz: Debugging Neural Networks with Coverage-Guided Fuzzing}.
\newblock {\em arXiv preprint arXiv:1807.10875}, 2018.

\bibitem{guo2018dlfuzz}
Jianmin Guo, Yu~Jiang, Yue Zhao, Quan Chen, and Jiaguang Sun.
\newblock Dlfuzz: differential fuzzing testing of deep learning systems.
\newblock In {\em Proc. FSE}, pages 739--743. ACM, 2018.

\bibitem{xie2018coverage}
Xiaofei Xie, Lei Ma, Felix Juefei-Xu, Hongxu Chen, Minhui Xue, Bo~Li, Yang Liu,
  Jianjun Zhao, Jianxiong Yin, and Simon See.
\newblock Coverage-guided fuzzing for deep neural networks.
\newblock {\em arXiv preprint arXiv:1809.01266}, 2018.

\bibitem{Ma2018ase}
Lei Ma, Felix Juefei-Xu, Fuyuan Zhang, Jiyuan Sun, Minhui Xue, Bo~Li, Chunyang
  Chen, Ting Su, Li~Li, Yang Liu, Jianjun Zhao, and Yadong Wang.
\newblock {DeepGauge: Multi-granularity Testing Criteria for Deep Learning
  Systems}.
\newblock In {\em Proceedings of the 33rd ACM/IEEE International Conference on
  Automated Software Engineering}, ASE 2018, pages 120--131, 2018.

\bibitem{wicker2018feature}
Matthew Wicker, Xiaowei Huang, and Marta Kwiatkowska.
\newblock Feature-guided black-box safety testing of deep neural networks.
\newblock In {\em International Conference on Tools and Algorithms for the
  Construction and Analysis of Systems}, pages 408--426, 2018.

\bibitem{uesato2018rigorous}
Jonathan Uesato, Ananya Kumar, Csaba Szepesvari, Tom Erez, Avraham Ruderman,
  Keith Anderson, Nicolas Heess, Pushmeet Kohli, et~al.
\newblock Rigorous agent evaluation: An adversarial approach to uncover
  catastrophic failures.
\newblock In {\em International Conference on Learning Representations}, 2019.

\bibitem{zhou2019metamorphic}
Zhi~Quan Zhou and Liqun Sun.
\newblock Metamorphic testing of driverless cars.
\newblock {\em Communications of the ACM}, 62(3):61--67, 2019.

\bibitem{jhaml2019mlbased}
Saurabh Jha, Subho~S Banerjee, Timothy Tsai, Siva~KS Hari, Michael~B Sullivan,
  Zbigniew~T Kalbarczyk, Stephen~W Keckler, and Ravishankar~K Iyer.
\newblock Ml-based fault injection for autonomous vehicles.
\newblock In {\em Proc. DSN}, 2019.

\bibitem{udeshi2019grammar}
Sakshi Udeshi and Sudipta Chattopadhyay.
\newblock Grammar based directed testing of machine learning systems.
\newblock {\em arXiv preprint arXiv:1902.10027}, 2019.

\bibitem{nie2018analyzing}
Yixin Nie, Yicheng Wang, and Mohit Bansal.
\newblock Analyzing compositionality-sensitivity of nli models.
\newblock {\em arXiv preprint arXiv:1811.07033}, 2018.

\bibitem{wang2018if}
Haohan Wang, Da~Sun, and Eric~P Xing.
\newblock What if we simply swap the two text fragments? a straightforward yet
  effective way to test the robustness of methods to confounding signals in
  nature language inference tasks.
\newblock {\em arXiv preprint arXiv:1809.02719}, 2018.

\bibitem{chan2018metamorphic}
Alvin Chan, Lei Ma, Felix Juefei-Xu, Xiaofei Xie, Yang Liu, and Yew~Soon Ong.
\newblock Metamorphic relation based adversarial attacks on differentiable
  neural computer.
\newblock {\em arXiv preprint arXiv:1809.02444}, 2018.

\bibitem{udeshi2018automated}
Sakshi Udeshi, Pryanshu Arora, and Sudipta Chattopadhyay.
\newblock Automated directed fairness testing.
\newblock In {\em Proceedings of the 33rd ACM/IEEE International Conference on
  Automated Software Engineering}, pages 98--108. ACM, 2018.

\bibitem{tuncali2018simulation}
Cumhur~Erkan Tuncali, Georgios Fainekos, Hisahiro Ito, and James Kapinski.
\newblock Simulation-based adversarial test generation for autonomous vehicles
  with machine learning components.
\newblock In {\em IEEE Intelligent Vehicles Symposium {(IV)}}, 2018.

\bibitem{hartman2005software}
Alan Hartman.
\newblock Software and hardware testing using combinatorial covering suites.
\newblock In {\em Graph theory, combinatorics and algorithms}, pages 237--266.
  Springer, 2005.

\bibitem{kirkpatrick1983optimization}
Scott Kirkpatrick, C~Daniel Gelatt, and Mario~P Vecchi.
\newblock Optimization by simulated annealing.
\newblock {\em science}, 220(4598):671--680, 1983.

\bibitem{king1976symbolic}
James~C King.
\newblock Symbolic execution and program testing.
\newblock {\em Communications of the ACM}, 19(7):385--394, 1976.

\bibitem{zhang2010test}
Lingming Zhang, Tao Xie, Lu~Zhang, Nikolai Tillmann, Jonathan De~Halleux, and
  Hong Mei.
\newblock Test generation via dynamic symbolic execution for mutation testing.
\newblock In {\em Software Maintenance (ICSM), 2010 IEEE International
  Conference on}, pages 1--10. IEEE, 2010.

\bibitem{chen2013state}
Ting Chen, Xiao-song Zhang, Shi-ze Guo, Hong-yuan Li, and Yue Wu.
\newblock State of the art: Dynamic symbolic execution for automated test
  generation.
\newblock {\em Future Generation Computer Systems}, 29(7):1758--1773, 2013.

\bibitem{gopinath2018symbolic}
Divya Gopinath, Kaiyuan Wang, Mengshi Zhang, Corina~S Pasareanu, and Sarfraz
  Khurshid.
\newblock Symbolic execution for deep neural networks.
\newblock {\em arXiv preprint arXiv:1807.10439}, 2018.

\bibitem{agarwal2018automated}
Aniya Agarwal, Pranay Lohia, Seema Nagar, Kuntal Dey, and Diptikalyan Saha.
\newblock Automated test generation to detect individual discrimination in ai
  models.
\newblock {\em arXiv preprint arXiv:1809.03260}, 2018.

\bibitem{ribeiro2016should}
Marco~Tulio Ribeiro, Sameer Singh, and Carlos Guestrin.
\newblock Why should i trust you?: Explaining the predictions of any
  classifier.
\newblock In {\em Proceedings of the 22nd ACM SIGKDD international conference
  on knowledge discovery and data mining}, pages 1135--1144. ACM, 2016.

\bibitem{Sunase2018}
Youcheng Sun, Min Wu, Wenjie Ruan, Xiaowei Huang, Marta Kwiatkowska, and Daniel
  Kroening.
\newblock Concolic testing for deep neural networks.
\newblock In {\em Proceedings of the 33rd ACM/IEEE International Conference on
  Automated Software Engineering}, ASE 2018, pages 109--119, 2018.

\bibitem{Murphy2007PRT}
Christian Murphy, Gail Kaiser, and Marta Arias.
\newblock {Parameterizing Random Test Data According to Equivalence Classes}.
\newblock In {\em Proc. ASE}, RT '07, pages 38--41, 2007.

\bibitem{nakajima2016dataset}
Shin Nakajima and Hai~Ngoc Bui.
\newblock Dataset coverage for testing machine learning computer programs.
\newblock In {\em 2016 23rd Asia-Pacific Software Engineering Conference
  (APSEC)}, pages 297--304. IEEE, 2016.

\bibitem{chen1998metamorphic}
Tsong~Y Chen, Shing~C Cheung, and Shiu~Ming Yiu.
\newblock Metamorphic testing: a new approach for generating next test cases.
\newblock Technical report, Technical Report HKUST-CS98-01, Department of
  Computer Science, Hong Kong University of Science and Technology, Hong Kong,
  1998.

\bibitem{Murphy2008PropertiesOM}
Christian Murphy, Gail~E. Kaiser, Lifeng Hu, and Leon Wu.
\newblock Properties of machine learning applications for use in metamorphic
  testing.
\newblock In {\em SEKE}, 2008.

\bibitem{Ding2017}
Junhua Ding, Xiaojun Kang, and Xin-Hua Hu.
\newblock Validating a deep learning framework by metamorphic testing.
\newblock In {\em Proceedings of the 2Nd International Workshop on Metamorphic
  Testing}, MET '17, pages 28--34, 2017.

\bibitem{murphy2009UJR}
Christian Murphy, Kuang Shen, and Gail Kaiser.
\newblock {Using {JML} Runtime Assertion Checking to Automate Metamorphic
  Testing in Applications Without Test Oracles}.
\newblock In {\em Proc. ICST}, pages 436--445. IEEE Computer Society, 2009.

\bibitem{xie2009application}
Xiaoyuan Xie, Joshua Ho, Christian Murphy, Gail Kaiser, Baowen Xu, and
  Tsong~Yueh Chen.
\newblock Application of metamorphic testing to supervised classifiers.
\newblock In {\em Quality Software, 2009. QSIC'09. 9th International Conference
  on}, pages 135--144. IEEE, 2009.

\bibitem{hall2009weka}
Mark Hall, Eibe Frank, Geoffrey Holmes, Bernhard Pfahringer, Peter Reutemann,
  and Ian~H Witten.
\newblock The weka data mining software: an update.
\newblock {\em ACM SIGKDD explorations newsletter}, 11(1):10--18, 2009.

\bibitem{nakajima2017generalized}
Shin Nakajima.
\newblock Generalized oracle for testing machine learning computer programs.
\newblock In {\em International Conference on Software Engineering and Formal
  Methods}, pages 174--179. Springer, 2017.

\bibitem{Dwarakanathissta2018}
Anurag Dwarakanath, Manish Ahuja, Samarth Sikand, Raghotham~M. Rao, R.~P.
  Jagadeesh~Chandra Bose, Neville Dubash, and Sanjay Podder.
\newblock Identifying implementation bugs in machine learning based image
  classifiers using metamorphic testing.
\newblock In {\em Proceedings of the 27th ACM SIGSOFT International Symposium
  on Software Testing and Analysis}, pages 118--128, 2018.

\bibitem{arnab2019balancedata}
Arnab Sharma and Heike Wehrheim.
\newblock Testing machine learning algorithms for balanced data usage.
\newblock In {\em Proc. ICST}, pages 125--135, 2019.

\bibitem{al2017validation}
Sadam Al-Azani and Jameleddine Hassine.
\newblock Validation of machine learning classifiers using metamorphic testing
  and feature selection techniques.
\newblock In {\em International Workshop on Multi-disciplinary Trends in
  Artificial Intelligence}, pages 77--91. Springer, 2017.

\bibitem{ramanagopal2018failing}
Manikandasriram~Srinivasan Ramanagopal, Cyrus Anderson, Ram Vasudevan, and
  Matthew Johnson-Roberson.
\newblock Failing to learn: Autonomously identifying perception failures for
  self-driving cars.
\newblock {\em IEEE Robotics and Automation Letters}, 3(4):3860--3867, 2018.

\bibitem{xie2018mettle}
Xiaoyuan Xie, Zhiyi Zhang, Tsong~Yueh Chen, Yang Liu, Pak-Lok Poon, and Baowen
  Xu.
\newblock Mettle: A metamorphic testing approach to validating unsupervised
  machine learning methods, 2018.

\bibitem{nakajima2018dataset}
Shin Nakajima.
\newblock Dataset diversity for metamorphic testing of machine learning
  software.
\newblock In {\em International Workshop on Structured Object-Oriented Formal
  Language and Method}, pages 21--38. Springer, 2018.

\bibitem{kim2018guiding}
Jinhan Kim, Robert Feldt, and Shin Yoo.
\newblock Guiding deep learning system testing using surprise adequacy.
\newblock {\em arXiv preprint arXiv:1808.08444}, 2018.

\bibitem{Murphy2009AST}
Christian Murphy, Kuang Shen, and Gail Kaiser.
\newblock {Automatic System Testing of Programs Without Test Oracles}.
\newblock In {\em 18th International Symposium on Software Testing and
  Analysis}, ISSTA '09, pages 189--200, 2009.

\bibitem{Zhou2018metamorphic}
L.~Sun and Z.~Q. Zhou.
\newblock {Metamorphic Testing for Machine Translations: MT4MT}.
\newblock In {\em 2018 25th Australasian Software Engineering Conference
  (ASWEC)}, pages 96--100, Nov 2018.

\bibitem{pesu2018monte}
Daniel Pesu, Zhi~Quan Zhou, Jingfeng Zhen, and Dave Towey.
\newblock A monte carlo method for metamorphic testing of machine translation
  services.
\newblock In {\em Proceedings of the 3rd International Workshop on Metamorphic
  Testing}, pages 38--45. ACM, 2018.

\bibitem{mckeeman1998differential}
William~M McKeeman.
\newblock Differential testing for software.
\newblock {\em Digital Technical Journal}, 10(1):100--107, 1998.

\bibitem{le2014compiler}
Vu~Le, Mehrdad Afshari, and Zhendong Su.
\newblock Compiler validation via equivalence modulo inputs.
\newblock In {\em ACM SIGPLAN Notices}, volume~49, pages 216--226. ACM, 2014.

\bibitem{nejadgholistudy}
Mahdi Nejadgholi and Jinqiu Yang.
\newblock A study of oracle approximations in testing deep learning libraries.
\newblock In {\em Proc. ASE}, pages 785--796, 2019.

\bibitem{avizienis1995methodology}
Algirdas Avizienis.
\newblock The methodology of n-version programming.
\newblock {\em Software fault tolerance}, 3:23--46, 1995.

\bibitem{srisakaokul2018multiple}
Siwakorn Srisakaokul, Zhengkai Wu, Angello Astorga, Oreoluwa Alebiosu, and Tao
  Xie.
\newblock Multiple-implementation testing of supervised learning software.
\newblock In {\em Proc. AAAI-18 Workshop on Engineering Dependable and Secure
  Machine Learning Systems (EDSMLS)}, 2018.

\bibitem{qin2018syneva}
Yi~Qin, Huiyan Wang, Chang Xu, Xiaoxing Ma, and Jian Lu.
\newblock Syneva: Evaluating ml programs by mirror program synthesis.
\newblock In {\em 2018 IEEE International Conference on Software Quality,
  Reliability and Security (QRS)}, pages 171--182. IEEE, 2018.

\bibitem{tjeng2017evaluating}
Vincent Tjeng, Kai Xiao, and Russ Tedrake.
\newblock Evaluating robustness of neural networks with mixed integer
  programming.
\newblock {\em arXiv preprint arXiv:1711.07356}, 2017.

\bibitem{dwork2012fairness}
Cynthia Dwork, Moritz Hardt, Toniann Pitassi, Omer Reingold, and Richard Zemel.
\newblock Fairness through awareness.
\newblock In {\em Proceedings of the 3rd innovations in theoretical computer
  science conference}, pages 214--226. ACM, 2012.

\bibitem{hardt2016equality}
Moritz Hardt, Eric Price, Nati Srebro, et~al.
\newblock Equality of opportunity in supervised learning.
\newblock In {\em Advances in neural information processing systems}, pages
  3315--3323, 2016.

\bibitem{zliobaite2017fairness}
Indre Zliobaite.
\newblock Fairness-aware machine learning: a perspective.
\newblock {\em arXiv preprint arXiv:1708.00754}, 2017.

\bibitem{herman2017promise}
Bernease Herman.
\newblock The promise and peril of human evaluation for model interpretability.
\newblock {\em arXiv preprint arXiv:1711.07414}, 2017.

\bibitem{kang2018model}
Daniel Kang, Deepti Raghavan, Peter Bailis, and Matei Zaharia.
\newblock Model assertions for debugging machine learning.
\newblock In {\em NeurIPS MLSys Workshop}, 2018.

\bibitem{li2015lifelong}
Lianghao Li and Qiang Yang.
\newblock Lifelong machine learning test.
\newblock In {\em Proceedings of the Workshop on “Beyond the Turing Test”
  of AAAI Conference on Artificial Intelligence}, 2015.

\bibitem{zhang2016PMT}
Jie Zhang, Ziyi Wang, Lingming Zhang, Dan Hao, Lei Zang, Shiyang Cheng, and
  Lu~Zhang.
\newblock Predictive mutation testing.
\newblock In {\em Proceedings of the 25th International Symposium on Software
  Testing and Analaysis}, ISSTA 2016, pages 342--353, 2016.

\bibitem{sun2018testing}
Youcheng Sun, Xiaowei Huang, and Daniel Kroening.
\newblock Testing deep neural networks.
\newblock {\em arXiv preprint arXiv:1803.04792}, 2018.

\bibitem{dupuy2000empirical}
Arnaud Dupuy and Nancy Leveson.
\newblock An empirical evaluation of the mc/dc coverage criterion on the hete-2
  satellite software.
\newblock In {\em Digital Avionics Systems Conference, 2000. Proceedings. DASC.
  The 19th}, volume~1, pages 1B6--1. IEEE, 2000.

\bibitem{sekhon2019improved}
Jasmine Sekhon and Cody Fleming.
\newblock Towards improved testing for deep learning.
\newblock In {\em Proc. ICSE(NIER track)}, pages 85--88, 2019.

\bibitem{ma2018combinatorial}
Lei Ma, Fuyuan Zhang, Minhui Xue, Bo~Li, Yang Liu, Jianjun Zhao, and Yadong
  Wang.
\newblock Combinatorial testing for deep learning systems.
\newblock {\em arXiv preprint arXiv:1806.07723}, 2018.

\bibitem{LeideepCT}
Lei Ma, Felix Juefei-Xu, Minhui Xue, Bo~Li, Li~Li, Yang Liu, and Jianjun Zhao.
\newblock Deepct: Tomographic combinatorial testing for deep learning systems.
\newblock In {\em Proc. SANER}, pages 614--618, 02 2019.

\bibitem{li2019structural}
Zenan Li, Xiaoxing Ma, Chang Xu, and Chun Cao.
\newblock Structural coverage criteria for neural networks could be misleading.
\newblock In {\em Proc. ICSE(NIER track)}, pages 89--92, 2019.

\bibitem{jia2011analysis}
Yue Jia and Mark Harman.
\newblock An analysis and survey of the development of mutation testing.
\newblock {\em IEEE transactions on software engineering}, 37(5):649--678,
  2011.

\bibitem{ma2018deepmutation}
Lei Ma, Fuyuan Zhang, Jiyuan Sun, Minhui Xue, Bo~Li, Felix Juefei-Xu, Chao Xie,
  Li~Li, Yang Liu, Jianjun Zhao, and Yadong Wang.
\newblock {DeepMutation: Mutation Testing of Deep Learning Systems}, 2018.

\bibitem{shen2018munn}
Weijun Shen, Jun Wan, and Zhenyu Chen.
\newblock {MuNN: Mutation Analysis of Neural Networks}.
\newblock In {\em 2018 IEEE International Conference on Software Quality,
  Reliability and Security Companion (QRS-C)}, pages 108--115. IEEE, 2018.

\bibitem{breck2017ml}
Eric Breck, Shanqing Cai, Eric Nielsen, Michael Salib, and D~Sculley.
\newblock The ml test score: A rubric for ml production readiness and technical
  debt reduction.
\newblock In {\em Big Data (Big Data), 2017 IEEE International Conference on},
  pages 1123--1132. IEEE, 2017.

\bibitem{byun2019input}
Taejoon Byun, Vaibhav Sharma, Abhishek Vijayakumar, Sanjai Rayadurgam, and
  Darren Cofer.
\newblock Input prioritization for testing neural networks.
\newblock {\em arXiv preprint arXiv:1901.03768}, 2019.

\bibitem{zhang2019noise}
Long Zhang, Xuechao Sun, Yong Li, Zhenyu Zhang, and Yang Feng.
\newblock A noise-sensitivity-analysis-based test prioritization technique for
  deep neural networks.
\newblock {\em arXiv preprint arXiv:1901.00054}, 2019.

\bibitem{li2019boosting}
Zenan Li, Xiaoxing Ma, Chang Xu, Chun Cao, Jingwei Xu, and Jian Lu.
\newblock {Boosting Operational DNN Testing Efficiency through Conditioning}.
\newblock In {\em Proc. FSE}, page to appear, 2019.

\bibitem{ma2019test}
Wei Ma, Mike Papadakis, Anestis Tsakmalis, Maxime Cordy, and Yves~Le Traon.
\newblock Test selection for deep learning systems, 2019.

\bibitem{thung2012empirical}
Ferdian Thung, Shaowei Wang, David Lo, and Lingxiao Jiang.
\newblock An empirical study of bugs in machine learning systems.
\newblock In {\em 2012 IEEE 23rd International Symposium on Software
  Reliability Engineering (ISSRE)}, pages 271--280. IEEE, 2012.

\bibitem{Zhangyuhaotensor}
Yuhao Zhang, Yifan Chen, Shing-Chi Cheung, Yingfei Xiong, and Lu~Zhang.
\newblock An empirical study on tensorflow program bugs.
\newblock In {\em Proceedings of the 27th ACM SIGSOFT International Symposium
  on Software Testing and Analysis}, pages 129--140, 2018.

\bibitem{Banerjee2018DSN}
S.~S. Banerjee, S.~Jha, J.~Cyriac, Z.~T. Kalbarczyk, and R.~K. Iyer.
\newblock Hands off the wheel in autonomous vehicles?: A systems perspective on
  over a million miles of field data.
\newblock In {\em 2018 48th Annual IEEE/IFIP International Conference on
  Dependable Systems and Networks (DSN)}, pages 586--597, June 2018.

\bibitem{ma2018mode}
Shiqing Ma, Yingqi Liu, Wen-Chuan Lee, Xiangyu Zhang, and Ananth Grama.
\newblock {MODE: Automated Neural Network Model Debugging via State
  Differential Analysis and Input Selection}.
\newblock pages 175--186, 2018.

\bibitem{dutta2019storm}
Saikat Dutta, Wenxian Zhang, Zixin Huang, and Sasa Misailovic.
\newblock Storm: Program reduction for testing and debugging probabilistic
  programming systems.
\newblock 2019.

\bibitem{cai2016tensorflow}
Shanqing Cai, Eric Breck, Eric Nielsen, Michael Salib, and D~Sculley.
\newblock Tensorflow debugger: Debugging dataflow graphs for machine learning.
\newblock In {\em Proceedings of the Reliable Machine Learning in the Wild-NIPS
  2016 Workshop}, 2016.

\bibitem{vartak2018mistique}
Manasi Vartak, Joana~M F~da Trindade, Samuel Madden, and Matei Zaharia.
\newblock Mistique: A system to store and query model intermediates for model
  diagnosis.
\newblock In {\em Proceedings of the 2018 International Conference on
  Management of Data}, pages 1285--1300. ACM, 2018.

\bibitem{krishnan2017palm}
Sanjay Krishnan and Eugene Wu.
\newblock Palm: Machine learning explanations for iterative debugging.
\newblock In {\em Proceedings of the 2nd Workshop on Human-In-the-Loop Data
  Analytics}, page~4. ACM, 2017.

\bibitem{nushi2017human}
Besmira Nushi, Ece Kamar, Eric Horvitz, and Donald Kossmann.
\newblock On human intellect and machine failures: Troubleshooting integrative
  machine learning systems.
\newblock In {\em AAAI}, pages 1017--1025, 2017.

\bibitem{albarghouthi2017repairing}
Aws Albarghouthi, Loris D’Antoni, and Samuel Drews.
\newblock Repairing decision-making programs under uncertainty.
\newblock In {\em International Conference on Computer Aided Verification},
  pages 181--200. Springer, 2017.

\bibitem{Yang2018TelemadeAT}
Wei Yang and Tao Xie.
\newblock Telemade: A testing framework for learning-based malware detection
  systems.
\newblock In {\em AAAI Workshops}, 2018.

\bibitem{dreossi2017systematic}
Tommaso Dreossi, Shromona Ghosh, Alberto Sangiovanni-Vincentelli, and Sanjit~A
  Seshia.
\newblock Systematic testing of convolutional neural networks for autonomous
  driving.
\newblock {\em arXiv preprint arXiv:1708.03309}, 2017.

\bibitem{tramer2017fairtest}
Florian Tramer, Vaggelis Atlidakis, Roxana Geambasu, Daniel Hsu, Jean-Pierre
  Hubaux, Mathias Humbert, Ari Juels, and Huang Lin.
\newblock Fairtest: Discovering unwarranted associations in data-driven
  applications.
\newblock In {\em 2017 IEEE European Symposium on Security and Privacy
  (EuroS\&P)}, pages 401--416. IEEE, 2017.

\bibitem{nishi2018test}
Yasuharu Nishi, Satoshi Masuda, Hideto Ogawa, and Keiji Uetsuki.
\newblock A test architecture for machine learning product.
\newblock In {\em 2018 IEEE International Conference on Software Testing,
  Verification and Validation Workshops (ICSTW)}, pages 273--278. IEEE, 2018.

\bibitem{thomas2019preventing}
Philip~S Thomas, Bruno~Castro da~Silva, Andrew~G Barto, Stephen Giguere, Yuriy
  Brun, and Emma Brunskill.
\newblock Preventing undesirable behavior of intelligent machines.
\newblock {\em Science}, 366(6468):999--1004, 2019.

\bibitem{kohavi1995study}
Ron Kohavi et~al.
\newblock A study of cross-validation and bootstrap for accuracy estimation and
  model selection.
\newblock In {\em Ijcai}, volume~14, pages 1137--1145. Montreal, Canada, 1995.

\bibitem{efron1994introduction}
Bradley Efron and Robert~J Tibshirani.
\newblock {\em An introduction to the bootstrap}.
\newblock CRC press, 1994.

\bibitem{japkowicz2006question}
Nathalie Japkowicz.
\newblock Why question machine learning evaluation methods.
\newblock In {\em AAAI Workshop on Evaluation Methods for Machine Learning},
  pages 6--11, 2006.

\bibitem{chen2012classifier}
Weijie Chen, Brandon~D Gallas, and Waleed~A Yousef.
\newblock Classifier variability: accounting for training and testing.
\newblock {\em Pattern Recognition}, 45(7):2661--2671, 2012.

\bibitem{chen2013assessment}
Weijie Chen, Frank~W Samuelson, Brandon~D Gallas, Le~Kang, Berkman Sahiner, and
  Nicholas Petrick.
\newblock On the assessment of the added value of new predictive biomarkers.
\newblock {\em BMC medical research methodology}, 13(1):98, 2013.

\bibitem{hynes2017data}
Nick Hynes, D~Sculley, and Michael Terry.
\newblock The data linter: Lightweight, automated sanity checking for ml data
  sets.
\newblock 2017.

\bibitem{krishnan2017boostclean}
Sanjay Krishnan, Michael~J Franklin, Ken Goldberg, and Eugene Wu.
\newblock Boostclean: Automated error detection and repair for machine
  learning.
\newblock {\em arXiv preprint arXiv:1711.01299}, 2017.

\bibitem{schelter2018automating}
Sebastian Schelter, Dustin Lange, Philipp Schmidt, Meltem Celikel, Felix
  Biessmann, and Andreas Grafberger.
\newblock Automating large-scale data quality verification.
\newblock {\em Proceedings of the VLDB Endowment}, 11(12):1781--1794, 2018.

\bibitem{baylor2017tfx}
Denis Baylor, Eric Breck, Heng-Tze Cheng, Noah Fiedel, Chuan~Yu Foo, Zakaria
  Haque, Salem Haykal, Mustafa Ispir, Vihan Jain, Levent Koc, et~al.
\newblock Tfx: A tensorflow-based production-scale machine learning platform.
\newblock In {\em Proceedings of the 23rd ACM SIGKDD International Conference
  on Knowledge Discovery and Data Mining}, pages 1387--1395. ACM, 2017.

\bibitem{hawkins2004problem}
Douglas~M Hawkins.
\newblock The problem of overfitting.
\newblock {\em Journal of chemical information and computer sciences},
  44(1):1--12, 2004.

\bibitem{chan1999classifier}
Heang-Ping Chan, Berkman Sahiner, Robert~F Wagner, and Nicholas Petrick.
\newblock Classifier design for computer-aided diagnosis: Effects of finite
  sample size on the mean performance of classical and neural network
  classifiers.
\newblock {\em Medical physics}, 26(12):2654--2668, 1999.

\bibitem{sahiner2000feature}
Berkman Sahiner, Heang-Ping Chan, Nicholas Petrick, Robert~F Wagner, and
  Lubomir Hadjiiski.
\newblock Feature selection and classifier performance in computer-aided
  diagnosis: The effect of finite sample size.
\newblock {\em Medical physics}, 27(7):1509--1522, 2000.

\bibitem{fukunaga1989effects}
Keinosuke Fukunaga and Raymond~R. Hayes.
\newblock Effects of sample size in classifier design.
\newblock {\em IEEE Transactions on Pattern Analysis \& Machine Intelligence},
  (8):873--885, 1989.

\bibitem{gossmann2018test}
Alexej Gossmann, Aria Pezeshk, and Berkman Sahiner.
\newblock Test data reuse for evaluation of adaptive machine learning
  algorithms: over-fitting to a fixed'test'dataset and a potential solution.
\newblock In {\em Medical Imaging 2018: Image Perception, Observer Performance,
  and Technology Assessment}, volume 10577, page 105770K. International Society
  for Optics and Photonics, 2018.

\bibitem{moosavi2016deepfool}
Seyed-Mohsen Moosavi-Dezfooli, Alhussein Fawzi, and Pascal Frossard.
\newblock Deepfool: a simple and accurate method to fool deep neural networks.
\newblock In {\em Proceedings of the IEEE Conference on Computer Vision and
  Pattern Recognition}, pages 2574--2582, 2016.

\bibitem{bastani2016measuring}
Osbert Bastani, Yani Ioannou, Leonidas Lampropoulos, Dimitrios Vytiniotis,
  Aditya Nori, and Antonio Criminisi.
\newblock Measuring neural net robustness with constraints.
\newblock In {\em Advances in neural information processing systems}, pages
  2613--2621, 2016.

\bibitem{carlini2017towards}
Nicholas Carlini and David Wagner.
\newblock Towards evaluating the robustness of neural networks.
\newblock In {\em 2017 IEEE Symposium on Security and Privacy (SP)}, pages
  39--57. IEEE, 2017.

\bibitem{ruan2018global}
Wenjie Ruan, Min Wu, Youcheng Sun, Xiaowei Huang, Daniel Kroening, and Marta
  Kwiatkowska.
\newblock Global robustness evaluation of deep neural networks with provable
  guarantees for l0 norm.
\newblock {\em arXiv preprint arXiv:1804.05805}, 2018.

\bibitem{gopinath2018deepsafe}
Divya Gopinath, Guy Katz, Corina~S P{\u{a}}s{\u{a}}reanu, and Clark Barrett.
\newblock Deepsafe: A data-driven approach for assessing robustness of neural
  networks.
\newblock In {\em International Symposium on Automated Technology for
  Verification and Analysis}, pages 3--19. Springer, 2018.

\bibitem{Mangal2019robustness}
Ravi Mangal, Aditya Nori, and Alessandro Orso.
\newblock Robustness of neural networks: A probabilistic and practical
  perspective.
\newblock In {\em Proc. ICSE(NIER track)}, pages 93--96, 2019.

\bibitem{Banerjee2019towards}
Subho~S Banerjee, James Cyriac, Saurabh Jha, Zbigniew~T Kalbarczyk, and
  Ravishankar~K Iyer.
\newblock Towards a bayesian approach for assessing faulttolerance of deep
  neural networks.
\newblock In {\em Proc. DSN (extended abstract)}, 2019.

\bibitem{papernot2016distillation}
Nicolas Papernot, Patrick McDaniel, Xi~Wu, Somesh Jha, and Ananthram Swami.
\newblock Distillation as a defense to adversarial perturbations against deep
  neural networks.
\newblock In {\em 2016 IEEE Symposium on Security and Privacy (SP)}, pages
  582--597. IEEE, 2016.

\bibitem{papernot2016cleverhans}
Nicolas Papernot, Ian Goodfellow, Ryan Sheatsley, Reuben Feinman, and Patrick
  McDaniel.
\newblock cleverhans v1.0.0: an adversarial machine learning library.
\newblock {\em arXiv preprint arXiv:1610.00768}, 2016.

\bibitem{papernot2018cleverhans}
Nicolas Papernot, Fartash Faghri, Nicholas Carlini, Ian Goodfellow, Reuben
  Feinman, Alexey Kurakin, Cihang Xie, Yash Sharma, Tom Brown, Aurko Roy,
  Alexander Matyasko, Vahid Behzadan, Karen Hambardzumyan, Zhishuai Zhang,
  Yi-Lin Juang, Zhi Li, Ryan Sheatsley, Abhibhav Garg, Jonathan Uesato, Willi
  Gierke, Yinpeng Dong, David Berthelot, Paul Hendricks, Jonas Rauber, and
  Rujun Long.
\newblock Technical report on the cleverhans v2.1.0 adversarial examples
  library.
\newblock {\em arXiv preprint arXiv:1610.00768}, 2018.

\bibitem{jha2018avfi}
Saurabh Jha, Subho~S Banerjee, James Cyriac, Zbigniew~T Kalbarczyk, and
  Ravishankar~K Iyer.
\newblock {AVFI: Fault injection for autonomous vehicles}.
\newblock In {\em 2018 48th Annual IEEE/IFIP International Conference on
  Dependable Systems and Networks Workshops (DSN-W)}, pages 55--56. IEEE, 2018.

\bibitem{jha2018kayotee}
Saurabh Jha, Timothy Tsai, Siva Hari, Michael Sullivan, Zbigniew Kalbarczyk,
  Stephen~W Keckler, and Ravishankar~K Iyer.
\newblock Kayotee: A fault injection-based system to assess the safety and
  reliability of autonomous vehicles to faults and errors.
\newblock In {\em 3rd IEEE International Workshop on Automotive Reliability \&
  Test}, 2018.

\bibitem{spieker2019towards}
Helge Spieker and Arnaud Gotlieb.
\newblock Towards testing of deep learning systems with training set reduction.
\newblock {\em arXiv preprint arXiv:1901.04169}, 2019.

\bibitem{barocas2016big}
Solon Barocas and Andrew~D Selbst.
\newblock Big data's disparate impact.
\newblock {\em Cal. L. Rev.}, 104:671, 2016.

\bibitem{gajane2017formalizing}
Pratik Gajane and Mykola Pechenizkiy.
\newblock On formalizing fairness in prediction with machine learning.
\newblock {\em arXiv preprint arXiv:1710.03184}, 2017.

\bibitem{verma2018fairness}
Sahil Verma and Julia Rubin.
\newblock Fairness definitions explained.
\newblock In {\em International Workshop on Software Fairness}, 2018.

\bibitem{saxena2018fairness}
Nripsuta Saxena, Karen Huang, Evan DeFilippis, Goran Radanovic, David Parkes,
  and Yang Liu.
\newblock How do fairness definitions fare? examining public attitudes towards
  algorithmic definitions of fairness.
\newblock {\em arXiv preprint arXiv:1811.03654}, 2018.

\bibitem{afetal:re08}
Anthony Finkelstein, Mark Harman, Afshin Mansouri, Jian Ren, and Yuanyuan
  Zhang.
\newblock Fairness analysis in requirements assignments.
\newblock In {\em $16^{th}$ {IEEE} International Requirements Engineering
  Conference}, pages 115--124, Los Alamitos, California, USA, September 2008.

\bibitem{kusner2017counterfactual}
Matt~J Kusner, Joshua Loftus, Chris Russell, and Ricardo Silva.
\newblock Counterfactual fairness.
\newblock In {\em Advances in Neural Information Processing Systems}, pages
  4066--4076, 2017.

\bibitem{grgic2016case}
Nina Grgic-Hlaca, Muhammad~Bilal Zafar, Krishna~P Gummadi, and Adrian Weller.
\newblock The case for process fairness in learning: Feature selection for fair
  decision making.
\newblock In {\em NIPS Symposium on Machine Learning and the Law, Barcelona,
  Spain}, volume~8, 2016.

\bibitem{zafar2015fairness}
Muhammad~Bilal Zafar, Isabel Valera, Manuel~Gomez Rodriguez, and Krishna~P
  Gummadi.
\newblock Fairness constraints: Mechanisms for fair classification.
\newblock {\em arXiv preprint arXiv:1507.05259}, 2015.

\bibitem{Metevier19neurips}
Blossom Metevier, Stephen Giguere, Sarah Brockman, Ari Kobren, Yuriy Brun, Emma
  Brunskill, and Philip Thomas.
\newblock {Offline Contextual Bandits with High Probability Fairness
  Guarantees}.
\newblock In {\em Proceedings of the 33rd Annual Conference on Neural
  Information Processing Systems (NeurIPS)}, December 2019.

\bibitem{li2010contextual}
Lihong Li, Wei Chu, John Langford, and Robert~E Schapire.
\newblock A contextual-bandit approach to personalized news article
  recommendation.
\newblock In {\em Proceedings of the 19th international conference on World
  wide web}, pages 661--670. ACM, 2010.

\bibitem{massart2007concentration}
Pascal Massart.
\newblock Concentration inequalities and model selection.
\newblock 2007.

\bibitem{agarwal2018reductions}
Alekh Agarwal, Alina Beygelzimer, Miroslav Dud{\'\i}k, John Langford, and Hanna
  Wallach.
\newblock A reductions approach to fair classification.
\newblock {\em arXiv preprint arXiv:1803.02453}, 2018.

\bibitem{angell2018themis}
Rico Angell, Brittany Johnson, Yuriy Brun, and Alexandra Meliou.
\newblock Themis: Automatically testing software for discrimination.
\newblock In {\em Proceedings of the 2018 26th ACM Joint Meeting on European
  Software Engineering Conference and Symposium on the Foundations of Software
  Engineering}, pages 871--875. ACM, 2018.

\bibitem{Brittany2018causaltesting}
Brittany Johnson, Yuriy Brun, and Alexandra Meliou.
\newblock Causal testing: Finding defects' root causes.
\newblock {\em CoRR}, abs/1809.06991, 2018.

\bibitem{Friedler2019interpretability}
Sorelle~A. Friedler, Chitradeep~Dutta Roy, Carlos Scheidegger, and Dylan Slack.
\newblock {Assessing the Local Interpretability of Machine Learning Models}.
\newblock {\em CoRR}, abs/1902.03501, 2019.

\bibitem{zhou2018metamorphicrelation}
Zhi~Quan Zhou, Liqun Sun, Tsong~Yueh Chen, and Dave Towey.
\newblock Metamorphic relations for enhancing system understanding and use.
\newblock {\em IEEE Transactions on Software Engineering}, 2018.

\bibitem{chen2018calibration}
Weijie Chen, Berkman Sahiner, Frank Samuelson, Aria Pezeshk, and Nicholas
  Petrick.
\newblock Calibration of medical diagnostic classifier scores to the
  probability of disease.
\newblock {\em Statistical methods in medical research}, 27(5):1394--1409,
  2018.

\bibitem{ding2018detecting}
Zeyu Ding, Yuxin Wang, Guanhong Wang, Danfeng Zhang, and Daniel Kifer.
\newblock Detecting violations of differential privacy.
\newblock In {\em Proceedings of the 2018 ACM SIGSAC Conference on Computer and
  Communications Security}, pages 475--489. ACM, 2018.

\bibitem{bichsel2018dp}
Benjamin Bichsel, Timon Gehr, Dana Drachsler-Cohen, Petar Tsankov, and Martin
  Vechev.
\newblock Dp-finder: Finding differential privacy violations by sampling and
  optimization.
\newblock In {\em Proceedings of the 2018 ACM SIGSAC Conference on Computer and
  Communications Security}, pages 508--524. ACM, 2018.

\bibitem{polyzotis2017data}
Neoklis Polyzotis, Sudip Roy, Steven~Euijong Whang, and Martin Zinkevich.
\newblock Data management challenges in production machine learning.
\newblock In {\em Proceedings of the 2017 ACM International Conference on
  Management of Data}, pages 1723--1726. ACM, 2017.

\bibitem{metzen2017detecting}
Jan~Hendrik Metzen, Tim Genewein, Volker Fischer, and Bastian Bischoff.
\newblock On detecting adversarial perturbations.
\newblock {\em arXiv preprint arXiv:1702.04267}, 2017.

\bibitem{wang2018detecting}
Jingyi Wang, Guoliang Dong, Jun Sun, Xinyu Wang, and Peixin Zhang.
\newblock Adversarial sample detection for deep neural network through model
  mutation testing.
\newblock In {\em Proc. ICSE}, pages 1245--1256, 2019.

\bibitem{jingyiwang2018detecting}
Jingyi Wang, Jun Sun, Peixin Zhang, and Xinyu Wang.
\newblock Detecting adversarial samples for deep neural networks through
  mutation testing.
\newblock {\em CoRR}, abs/1805.05010, 2018.

\bibitem{Carlini2017}
Nicholas Carlini and David Wagner.
\newblock Adversarial examples are not easily detected: Bypassing ten detection
  methods.
\newblock In {\em Proceedings of the 10th ACM Workshop on Artificial
  Intelligence and Security}, AISec '17, pages 3--14. ACM, 2017.

\bibitem{krishnan2016activeclean}
Sanjay Krishnan, Jiannan Wang, Eugene Wu, Michael~J Franklin, and Ken Goldberg.
\newblock Activeclean: interactive data cleaning for statistical modeling.
\newblock {\em Proceedings of the VLDB Endowment}, 9(12):948--959, 2016.

\bibitem{krishnan2016activeclean2}
Sanjay Krishnan, Michael~J Franklin, Ken Goldberg, Jiannan Wang, and Eugene Wu.
\newblock Activeclean: An interactive data cleaning framework for modern
  machine learning.
\newblock In {\em Proceedings of the 2016 International Conference on
  Management of Data}, pages 2117--2120. ACM, 2016.

\bibitem{liu2008isolation}
Fei~Tony Liu, Kai~Ming Ting, and Zhi-Hua Zhou.
\newblock Isolation forest.
\newblock In {\em 2008 Eighth IEEE International Conference on Data Mining},
  pages 413--422. IEEE, 2008.

\bibitem{krishnan2019alphaclean}
Sanjay Krishnan and Eugene Wu.
\newblock Alphaclean: automatic generation of data cleaning pipelines.
\newblock {\em arXiv preprint arXiv:1904.11827}, 2019.

\bibitem{rahm2000data}
Erhard Rahm and Hong~Hai Do.
\newblock Data cleaning: Problems and current approaches.
\newblock {\em IEEE Data Eng. Bull.}, 23(4):3--13, 2000.

\bibitem{mcclure2017tensorflow}
Nick McClure.
\newblock {\em TensorFlow machine learning cookbook}.
\newblock Packt Publishing Ltd, 2017.

\bibitem{schaul2014unit}
Tom Schaul, Ioannis Antonoglou, and David Silver.
\newblock Unit tests for stochastic optimization.
\newblock In {\em ICLR 2014}, 2014.

\bibitem{sunempirical2017}
X.~Sun, T.~Zhou, G.~Li, J.~Hu, H.~Yang, and B.~Li.
\newblock An empirical study on real bugs for machine learning programs.
\newblock In {\em 2017 24th Asia-Pacific Software Engineering Conference
  (APSEC)}, pages 348--357, 2017.

\bibitem{Guo2018study}
An~Orchestrated Empirical~Study on~Deep Learning~Frameworks and Lei Ma Qiang Hu
  Ruitao Feng Li Li Yang Liu Jianjun Zhao Xiaohong~Li Platforms Qianyu~Guo,
  Xiaofei~Xie.
\newblock An orchestrated empirical study on deep learning frameworks and
  platforms.
\newblock {\em arXiv preprint arXiv:1811.05187}, 2018.

\bibitem{Karpov2018}
Yu.~L. Karpov, L.~E. Karpov, and Yu.~G. Smetanin.
\newblock Adaptation of general concepts of software testing to neural
  networks.
\newblock {\em Programming and Computer Software}, 44(5):324--334, Sep 2018.

\bibitem{Fu2017fse}
Wei Fu and Tim Menzies.
\newblock Easy over hard: A case study on deep learning.
\newblock In {\em Proc. FSE}, ESEC/FSE 2017, pages 49--60. ACM, 2017.

\bibitem{Liu2018}
Zhongxin Liu, Xin Xia, Ahmed~E. Hassan, David Lo, Zhenchang Xing, and Xinyu
  Wang.
\newblock Neural-machine-translation-based commit message generation: How far
  are we?
\newblock In {\em Proceedings of the 33rd ACM/IEEE International Conference on
  Automated Software Engineering}, ASE 2018, pages 373--384. ACM, 2018.

\bibitem{Dolby2018Ariadne}
Julian Dolby, Avraham Shinnar, Allison Allain, and Jenna Reinen.
\newblock Ariadne: Analysis for machine learning programs.
\newblock In {\em Proceedings of the 2Nd ACM SIGPLAN International Workshop on
  Machine Learning and Programming Languages}, MAPL 2018, pages 1--10, 2018.

\bibitem{xiao2018security}
Qixue Xiao, Kang Li, Deyue Zhang, and Weilin Xu.
\newblock Security risks in deep learning implementations.
\newblock In {\em 2018 IEEE Security and Privacy Workshops (SPW)}, pages
  123--128. IEEE, 2018.

\bibitem{chase2017unitest}
Chase Roberts.
\newblock How to unit test machine learning code, 2017.

\bibitem{cheng2018manifesting}
Dawei Cheng, Chun Cao, Chang Xu, and Xiaoxing Ma.
\newblock Manifesting bugs in machine learning code: An explorative study with
  mutation testing.
\newblock In {\em 2018 IEEE International Conference on Software Quality,
  Reliability and Security (QRS)}, pages 313--324. IEEE, 2018.

\bibitem{Xie2011JSS}
Xiaoyuan Xie, Joshua W.~K. Ho, Christian Murphy, Gail Kaiser, Baowen Xu, and
  Tsong~Yueh Chen.
\newblock Testing and validating machine learning classifiers by metamorphic
  testing.
\newblock {\em J. Syst. Softw.}, 84(4):544--558, April 2011.

\bibitem{ma2005mujava}
Yu-Seung Ma, Jeff Offutt, and Yong~Rae Kwon.
\newblock Mujava: an automated class mutation system.
\newblock {\em Software Testing, Verification and Reliability}, 15(2):97--133,
  2005.

\bibitem{wegener2004evaluation}
Joachim Wegener and Oliver B{\"u}hler.
\newblock Evaluation of different fitness functions for the evolutionary
  testing of an autonomous parking system.
\newblock In {\em Genetic and Evolutionary Computation Conference}, pages
  1400--1412. Springer, 2004.

\bibitem{woehrle2019open}
Matthias Woehrle, Christoph Gladisch, and Christian Heinzemann.
\newblock Open questions in testing of learned computer vision functions for
  automated driving.
\newblock In {\em International Conference on Computer Safety, Reliability, and
  Security}, pages 333--345. Springer, 2019.

\bibitem{abdessalem2018testing2}
Raja~Ben Abdessalem, Shiva Nejati, Lionel~C Briand, and Thomas Stifter.
\newblock Testing vision-based control systems using learnable evolutionary
  algorithms.
\newblock In {\em Proc. ICSE}, pages 1016--1026. IEEE, 2018.

\bibitem{ben2016testing}
Raja Ben~Abdessalem, Shiva Nejati, Lionel~C Briand, and Thomas Stifter.
\newblock Testing advanced driver assistance systems using multi-objective
  search and neural networks.
\newblock In {\em Proceedings of the 31st IEEE/ACM International Conference on
  Automated Software Engineering}, pages 63--74. ACM, 2016.

\bibitem{abdessalem2018testing}
Raja~Ben Abdessalem, Annibale Panichella, Shiva Nejati, Lionel~C Briand, and
  Thomas Stifter.
\newblock Testing autonomous cars for feature interaction failures using
  many-objective search.
\newblock In {\em Proceedings of the 33rd ACM/IEEE International Conference on
  Automated Software Engineering}, pages 143--154. ACM, 2018.

\bibitem{papineni2002bleu}
Kishore Papineni, Salim Roukos, Todd Ward, and Wei-Jing Zhu.
\newblock Bleu: a method for automatic evaluation of machine translation.
\newblock In {\em Proceedings of the 40th annual meeting on association for
  computational linguistics}, pages 311--318. Association for Computational
  Linguistics, 2002.

\bibitem{Zheng2018TestingUN}
Wujie Zheng, Wenyu Wang, Dian Liu, Changrong Zhang, Qinsong Zeng, Yuetang Deng,
  Wei Yang, Pinjia He, and Tao Xie.
\newblock Testing untestable neural machine translation: An industrial case.
\newblock 2018.

\bibitem{zheng2019TestingUNposter}
Wujie Zheng, Wenyu Wang, Dian Liu, Changrong Zhang, Qinsong Zeng, Yuetang Deng,
  Wei Yang, Pinjia He, and Tao Xie.
\newblock Testing untestable neural machine translation: An industrial case.
\newblock In {\em ICSE (poster track)}, 2019.

\bibitem{wang2019TestingUNposter}
Wenyu Wang, Wujie Zheng, Dian Liu, Changrong Zhang, Qinsong Zeng, Yuetang Deng,
  Wei Yang, Pinjia He, and Tao Xie.
\newblock Detecting failures of neural machine translation in the absence of
  reference translations.
\newblock In {\em Proc. DSN (industry track)}, 2019.

\bibitem{burrell2016machine}
Jenna Burrell.
\newblock How the machine ‘thinks’: Understanding opacity in machine
  learning algorithms.
\newblock {\em Big Data \& Society}, 3(1):2053951715622512, 2016.

\bibitem{mnisthandwrittendigit}
Yann LeCun and Corinna Cortes.
\newblock {MNIST} handwritten digit database.
\newblock 2010.

\bibitem{Xiao2017FashionMNISTAN}
Han Xiao, Kashif Rasul, and Roland Vollgraf.
\newblock Fashion-mnist: a novel image dataset for benchmarking machine
  learning algorithms.
\newblock {\em CoRR}, abs/1708.07747, 2017.

\bibitem{CIFARdataset}
Alex Krizhevsky, Vinod Nair, and Geoffrey Hinton.
\newblock Cifar-10 (canadian institute for advanced research).

\bibitem{imagenet_cvpr09}
J.~Deng, W.~Dong, R.~Socher, L.-J. Li, K.~Li, and L.~Fei-Fei.
\newblock {ImageNet: A Large-Scale Hierarchical Image Database}.
\newblock In {\em CVPR09}, 2009.

\bibitem{dataset_irisflower}
R.A. Fisher.
\newblock {Iris Data Set }.
\newblock \url{http://archive.ics.uci.edu/ml/datasets/iris}.

\bibitem{netzer2011reading}
Yuval Netzer, Tao Wang, Adam Coates, Alessandro Bissacco, Bo~Wu, and Andrew~Y
  Ng.
\newblock Reading digits in natural images with unsupervised feature learning.
\newblock 2011.

\bibitem{fruit360}
Horea Mureșan and Mihai Oltean.
\newblock Fruit recognition from images using deep learning.
\newblock {\em Acta Universitatis Sapientiae, Informatica}, 10:26--42, 06 2018.

\bibitem{dataset_kaggleletters}
Olga Belitskaya.
\newblock {Handwritten Letters}.
\newblock \url{https://www.kaggle.com/olgabelitskaya/handwritten-letters}.

\bibitem{dataset_waveform}
{Balance Scale Data Set }.
\newblock \url{http://archive.ics.uci.edu/ml/datasets/balance+scale}.

\bibitem{dataset_DSRC}
Sharaf Malebary.
\newblock {DSRC Vehicle Communications Data Set }.
\newblock
  \url{http://archive.ics.uci.edu/ml/datasets/DSRC+Vehicle+Communications}.

\bibitem{getnexar}
Nexar.
\newblock the nexar dataset.
\newblock \url{https://www.getnexar.com/challenge-1/}.

\bibitem{dataset_coco}
Tsung-Yi Lin, Michael Maire, Serge Belongie, James Hays, Pietro Perona, Deva
  Ramanan, Piotr Doll{\'a}r, and C.~Lawrence Zitnick.
\newblock Microsoft coco: Common objects in context.
\newblock In David Fleet, Tomas Pajdla, Bernt Schiele, and Tinne Tuytelaars,
  editors, {\em Computer Vision -- ECCV 2014}, pages 740--755, Cham, 2014.
  Springer International Publishing.

\bibitem{pan2017virtual}
Xinlei Pan, Yurong You, Ziyan Wang, and Cewu Lu.
\newblock Virtual to real reinforcement learning for autonomous driving.
\newblock {\em arXiv preprint arXiv:1704.03952}, 2017.

\bibitem{dataset_KITTI}
A~Geiger, P~Lenz, C~Stiller, and R~Urtasun.
\newblock Vision meets robotics: The kitti dataset.
\newblock {\em The International Journal of Robotics Research},
  32(11):1231--1237, 2013.

\bibitem{bAbi}
Facebook research.
\newblock the babi dataset.
\newblock \url{https://research.fb.com/downloads/babi/}.

\bibitem{karpathy2015visualizing}
Andrej Karpathy, Justin Johnson, and Li~Fei-Fei.
\newblock Visualizing and understanding recurrent networks.
\newblock {\em arXiv preprint arXiv:1506.02078}, 2015.

\bibitem{dataset_stackoverflow}
{ Stack Exchange Data Dump }.
\newblock \url{https://archive.org/details/stackexchange}.

\bibitem{bowman2015large}
Samuel~R Bowman, Gabor Angeli, Christopher Potts, and Christopher~D Manning.
\newblock A large annotated corpus for learning natural language inference.
\newblock {\em arXiv preprint arXiv:1508.05326}, 2015.

\bibitem{N18-1101}
Adina Williams, Nikita Nangia, and Samuel Bowman.
\newblock A broad-coverage challenge corpus for sentence understanding through
  inference.
\newblock In {\em Proceedings of the 2018 Conference of the North American
  Chapter of the Association for Computational Linguistics: Human Language
  Technologies, Volume 1 (Long Papers)}, pages 1112--1122. Association for
  Computational Linguistics, 2018.

\bibitem{dataset_DVM}
Califormia dmv failure reports.
\newblock
  \url{https://www.dmv.ca.gov/portal/dmv/detail/vr/autonomous/autonomousveh_ol316}.

\bibitem{dataset_germancredit}
Dr.~Hans Hofmann.
\newblock Statlog (german credit data) data set.
\newblock
  \url{http://archive.ics.uci.edu/ml/datasets/statlog+(german+credit+data)}.

\bibitem{dataset_adult}
Ron Kohav.
\newblock Adult data set.
\newblock \url{http://archive.ics.uci.edu/ml/datasets/adult}.

\bibitem{moro2014data}
S{\'e}rgio Moro, Paulo Cortez, and Paulo Rita.
\newblock A data-driven approach to predict the success of bank telemarketing.
\newblock {\em Decision Support Systems}, 62:22--31, 2014.

\bibitem{dataset_usexecution}
Executions in the united states.
\newblock \url{https://deathpenaltyinfo.org/views-executions}.

\bibitem{dal2015calibrating}
Andrea Dal~Pozzolo, Olivier Caelen, Reid~A Johnson, and Gianluca Bontempi.
\newblock Calibrating probability with undersampling for unbalanced
  classification.
\newblock In {\em Computational Intelligence, 2015 IEEE Symposium Series on},
  pages 159--166. IEEE, 2015.

\bibitem{bickel1975sex}
Peter~J Bickel, Eugene~A Hammel, and J~William O'Connell.
\newblock {Sex bias in graduate admissions: Data from Berkeley}.
\newblock {\em Science}, 187(4175):398--404, 1975.

\bibitem{dataset_compas}
propublica.
\newblock data for the propublica story `machine bias'.
\newblock \url{https://github.com/propublica/compas-analysis/}.

\bibitem{harper2016movielens}
F~Maxwell Harper and Joseph~A Konstan.
\newblock The movielens datasets: History and context.
\newblock {\em Acm transactions on interactive intelligent systems (tiis)},
  5(4):19, 2016.

\bibitem{dataset_zillow}
Zillow.
\newblock {Zillow Prize: Zillow’s Home Value Prediction (Zestimate)}.
\newblock \url{https://www.kaggle.com/c/zillow-prize-1/overview}.

\bibitem{reserve2007report}
US~Federal Reserve.
\newblock Report to the congress on credit scoring and its effects on the
  availability and affordability of credit.
\newblock {\em Board of Governors of the Federal Reserve System}, 2007.

\bibitem{dataset_LSAC}
Law School~Admission Council.
\newblock Lsac national longitudinal bar passage study (nlbps).
\newblock
  \url{http://academic.udayton.edu/race/03justice/legaled/Legaled04.htm}.

\bibitem{dataset_VirusTotal}
VirusTotal.
\newblock Virustotal.
\newblock \url{https://www.virustotal.com/#/home/search}.

\bibitem{dataset_contagio}
contagio malware dump.
\newblock
  \url{http://contagiodump.blogspot.com/2013/03/16800-clean-and-11960-malicious-files.html}.

\bibitem{arp2014drebin}
Daniel Arp, Michael Spreitzenbarth, Malte Hubner, Hugo Gascon, Konrad Rieck,
  and CERT Siemens.
\newblock Drebin: Effective and explainable detection of android malware in
  your pocket.
\newblock In {\em Ndss}, volume~14, pages 23--26, 2014.

\bibitem{dataset_chess}
Alen Shapiro.
\newblock {UCI} chess (king-rook vs. king-pawn) data set.
\newblock
  \url{https://archive.ics.uci.edu/ml/datasets/Chess+%28King-Rook+vs.+King-Pawn%29},
  1989.

\bibitem{harman2018start}
Mark Harman and Peter O'Hearn.
\newblock From start-ups to scale-ups: opportunities and open problems for
  static and dynamic program analysis.
\newblock In {\em 2018 IEEE 18th International Working Conference on Source
  Code Analysis and Manipulation (SCAM)}, pages 1--23. IEEE, 2018.

\bibitem{afetal:fairness-jv}
Anthony Finkelstein, Mark Harman, Afshin Mansouri, Jian Ren, and Yuanyuan
  Zhang.
\newblock A search based approach to fairness analysis in requirements
  assignments to aid negotiation, mediation and decision making.
\newblock {\em Requirements Engineering}, 14(4):231--245, 2009.

\bibitem{zhang2019pseudo}
Jie Zhang, Lingming Zhang, Dan Hao, Meng Wang, and Lu~Zhang.
\newblock Do pseudo test suites lead to inflated correlation in measuring test
  effectiveness?
\newblock In {\em Proc. ICST}, pages 252--263, 2019.

\bibitem{weiss2016survey}
Karl Weiss, Taghi~M Khoshgoftaar, and DingDing Wang.
\newblock A survey of transfer learning.
\newblock {\em Journal of Big data}, 3(1):9, 2016.

\bibitem{nakajima2018quality}
Shin NAKAJIMA.
\newblock Quality assurance of machine learning software.
\newblock In {\em 2018 IEEE 7th Global Conference on Consumer Electronics
  (GCCE)}, pages 601--604. IEEE, 2018.

\end{thebibliography}

\end{document}